\documentclass[lettersize,journal]{IEEEtran}
\usepackage{subfigure} 
\usepackage[permil]{overpic}
\usepackage{booktabs}
\usepackage[pagebackref=true,breaklinks=true,colorlinks,bookmarks=false]{hyperref} 

\usepackage{amsmath,amsfonts}
\usepackage{algorithmic}
\usepackage{array}
\usepackage{textcomp}
\usepackage{stfloats}
\usepackage{url}
\usepackage{verbatim}
\usepackage{graphicx}
\usepackage{multirow}
\def\BibTeX{{\rm B\kern-.05em{\sc i\kern-.025em b}\kern-.08em
    T\kern-.1667em\lower.7ex\hbox{E}\kern-.125emX}}
\usepackage{balance}

\newcommand{\cuthalfcaptionup}{\vspace*{-5pt}}

\newcommand{\rebuttal}[1]{\textcolor{black}{#1}}

\def\eg{\textit{e.g.}}

\def\etal{\textit{et al.}}

\begin{document}
\title{S-NeRF++: Autonomous Driving Simulation via Neural Reconstruction and Generation}
\author{ Yurui Chen \quad Junge Zhang \quad Ziyang Xie \quad Wenye Li \quad Feihu Zhang \quad Jiachen Lu  \quad Li Zhang 

\thanks{
Li Zhang is the corresponding author (lizhangfd@fudan.edu.cn).
Yurui Chen, Junge Zhang, Ziyang Xie, Wenye Li, Jiachen Lu and Li Zhang are with the School of Data Science, Fudan University.
Feihu Zhang is with University of Oxford.
 }
}

\markboth{Journal of \LaTeX\ Class Files,~Vol.~18, No.~9, September~2020}%
{How to Use the IEEEtran \LaTeX \ Templates}

\maketitle

\begin{abstract} 
Autonomous driving simulation system plays a crucial role in enhancing self-driving data and simulating complex and rare traffic scenarios, ensuring navigation safety. However, traditional simulation systems, which often heavily rely on manual modeling and 2D image editing, struggled with scaling to extensive scenes and generating realistic simulation data.
In this study, we present S-NeRF++, an innovative autonomous driving simulation system based on neural reconstruction. Trained on widely-used self-driving datasets such as nuScenes and Waymo, S-NeRF++ can generate a large number of realistic street scenes and foreground objects with high rendering quality as well as offering considerable flexibility in manipulation and simulation. 
Specifically, S-NeRF++ is an enhanced neural radiance field for synthesizing large-scale scenes and moving vehicles, with improved scene parameterization and camera pose learning. The system effectively utilizes noisy and sparse LiDAR data to refine training and address depth outliers, ensuring high-quality reconstruction and novel-view rendering. It also provides a diverse foreground asset bank by reconstructing and generating different foreground vehicles to support comprehensive scenario creation.
Moreover, we have developed an advanced foreground-background fusion pipeline that skillfully integrates illumination and shadow effects, further enhancing the realism of our simulations. 
With the high-quality simulated data provided by our S-NeRF++, we found the perception methods enjoy performance boosts on several autonomous driving downstream tasks, further demonstrating our proposed simulator's effectiveness.
\end{abstract}


\section{Introduction}
\IEEEPARstart{T}{he} recent surge in street view data acquisition by autonomous vehicles has opened new avenues in the field of autonomous driving simulations. This data can be effectively employed to construct simulation systems that are not only capable of generating new datasets but also proficient in synthesizing corner case scenarios. Historically, this area has seen the introduction of various innovative methods, reflecting a significant advancement in autonomous driving technology.
\cite{CARLA, Airsim} leverage virtual 3D modeling and rendering engines to establish comprehensive simulation systems. Another notable technique, as proposed by \cite{AADS}, involves the innovative use of image warping to generate simulations with novel perspectives. \cite{geosim} focuses on reconstructing vehicle meshes from self-driving datasets and integrating them into videos based on depth information. \cite{drivegan} takes advantage of generative adversarial networks (GANs) to enhance the realism of edited videos, aligning them more closely with given datasets. Despite their merits, these methods either demand substantial computing resources or exhibit limitations in terms of simulation flexibility.

The recent emergence of neural radiance fields (NeRFs)~\cite{nerf}, a kind of 3D reconstruction algorithm which has shown impressive performance on photo-realistic novel view rendering, gives us a new way of thinking about construction simulation systems. Original NeRF is usually designed for object-centric scenes and requires camera views to be heavily overlapped.
However, urban driving data are often collected in the unbounded outdoor scenes (\eg\  nuScenes~\cite{nuscenes2019} and Waymo \cite{Sun_2020_CVPR} datasets). The camera placements of such data acquisition systems are usually in panoramic settings without object-centric camera views. This presents a challenge for NeRF in achieving high-quality reconstructions in these contexts.

We introduce a novel NeRF design, S-NeRF++, as the core of our driving simulation system for synthesizing both expansive background scenes and dynamic foreground vehicles. To address the challenge NeRF faces in accurately reconstructing large scenes, we leverage dense depth supervision to guide the NeRF training. The per-pixel dense depth maps are completed from the sparse LiDAR points using an off-the-shelf depth completion model. To obtain more reliable depth supervision, we introduce to use learnable confidence map to evaluate the depth quality thus ensuring more robust and stable optimization. 

To construct an effective simulation system, it's essential to include manipulable foreground objects for dynamic and realistic scenario generation. 
\cite{unisim, mars} solve this problem by decoupling the foreground reconstruction from training logs, enabling manipulation of foreground vehicles. However, their approach is limited to reusing vehicles within the same scene and does not consider the natural composition of foreground objects and background scenes, which leads to unrealistic generation. To address this issue, we have developed a sophisticated foreground-background fusion algorithm. This solution enhances the realism of the generated virtual data and ensures that the lighting of the inserted foreground objects seamlessly adapts to various background environmental conditions, such as night and cloudy settings. Furthermore, our sampling strategy facilitates the automated placement of foreground objects, streamlining the process of integrating these elements into diverse scenarios without manual intervention. This approach significantly enhances the efficiency and flexibility of the simulation, allowing for rapid and diverse scene generation.

We proposed to use advanced generative algorithms to further enhance the variety of our foreground data. In order to ensure a comprehensive and diverse range of scenarios, we leverage a category-specific, fine-tuned latent diffusion model (LDM) to generate various foreground assets from uncolored meshes, including but not limited to people, bicycles, and motorcycles, which are not present in the training dataset.
Beyond just utilizing generative algorithms, we also broaden our foreground asset library by aggregating and reconstructing a variety of objects from multiple datasets. This comprehensive approach not only diversifies the types of foreground objects available but also enhances the overall richness and authenticity of our asset library.

\textcolor{black}{
\textbf{Our preliminary work S-NeRF~\cite{snerf} has been published in ICLR 2023.} We have extended our conference version as follows: 
\textbf{(i) Superior reconstruction quality.} 
S-NeRF++ delivers substantially improved reconstruction quality, particularly in its ability to jointly reconstruct dynamic vehicles and static backgrounds.
\textbf{(ii) Foreground asset bank and enrichment.} 
Unlike S-NeRF, S-NeRF++ introduces a novel method for generating a foreground asset bank. This generation method not only diversifies the simulated data but also directly enhances the quality and realism of simulations. The ability to generate and manage a rich variety of foreground elements is a significant advancement that extends the utility of the simulation system beyond what S-NeRF could achieve.
\textbf{(iii) Automated object insertion pipeline.} 
The object insertion demonstrated in S-NeRF lacks the automated, scalable pipeline that S-NeRF++ implements. S-NeRF++ supports efficient, automated data production with varied labels, optimized for downstream tasks. Furthermore, S-NeRF++ includes an object insertion refinement module that S-NeRF did not, enabling higher-quality compositions through techniques like shadowing, relighting, and BEV-aware pose sampling. These enhancements are critical for producing realistic and diverse simulation scenarios.
\textbf{(iv) Proven effectiveness in downstream tasks.} 
Most importantly, S-NeRF++ has demonstrated its effectiveness in enhancing downstream tasks, proving the practical value of the advancements we have made.
}
\section{Related Work}

\subsection{Data-driven self-driving simulation system}
Data-driven self-driving simulation system aims to simulate the driving environment or novel data for the self-driving tasks from the given sensor data collected by a vehicle driving in the urban. Different from the simulation platforms~\cite{CARLA, Airsim} based on virtual 3D modeling and rendering engine. Data-driven simulation has less computation demand and less sim-to-real gap, making it a more efficient and realistic approach for training and testing autonomous driving systems. \cite{AADS} use warping images to get novel view simulation. \cite{geosim} reconstructs meshes of vehicles from the self-driving dataset and inserts the reconstructed cars into given videos through the depth, while it can not synthesize the novel view of the urban scene leading to the lack of diversity. \cite{lidarsim} simulates the LiDAR data by aggregating the LiDAR scans, building meshes and performing raycasting to generate point clouds. \cite{drivegan} uses GANs to edit videos more reasonably and similar to the given dataset. \cite{unisim, mars} utilize NeRF-based methods decoupled the foreground from training logs to possess the ability to render scenes when manipulating foreground vehicles to some extent.

However, these methods frequently struggle to generate novel views of urban scenes, suffer from a lack of diversity in the generated data, and encounter difficulties in accurately rendering foreground vehicles when manipulating them within the scenes. These issues can hinder the effectiveness and realism of the simulation system.

\subsection{3D reconstruction and novel view synthesis}
3D reconstruction of objects or scenes through images has always been a popular but difficult research topic.
Traditional reconstruction and novel view rendering~\cite{agarwal2011building} often rely on structure-from-motion (SfM), multi-view stereo and graphic rendering~\cite{losasso2004geometry}. Recently, from differentiable rasterization to volume rendering, learning-based approaches~\cite{ DeepVoxels, DISN, Engelmann_2021_CVPR} have been widely used in 3D scene and object reconstruction. They often use a deep neural network to extract the feature of the sensor inputs and learn various representations, such as voxels~\cite{kar2017learning,DeepVoxels},  patches~\cite{AtlasNet} and meshes~\cite{Pixel2Mesh,geosim}, to reconstruct the 3D geometry of the scene.

Recently, Neural Radiance Fields (NeRF)~\cite{nerf} has been widely used for 3D reconstruction. To further speed up training, various types of NeRFs~\cite{yu2021plenoctrees,Rebain_2021_CVPR} have been proposed. Explicit grid occupancy is used in~\cite{DVGO, NSVF} for more efficient query by raytracing, hash grid-based method like~\cite{Instant-NGP} balance speed and storage.  

\paragraph{Unbounded street-view NeRF}
Many NeRFs have been proposed to address the challenges of large-scale outdoor scenes.
NeRF in the wild~\cite{nerfW}  applies appearance and transient embeddings to solve the lighting changes and transient occlusions. 
Using Mip-NeRF as a base block, Block-NeRF~\cite{blocknerf} employs a block-combination strategy along with pose refinement,  appearance, and exposure embedding on large-scale scenes. Mip-NeRF 360~\cite{mipnerf360} improves the Mip-NeRF for unbounded scenes by contracting the whole space into a bounded area to get a more representative position encoding. Zip-NeRF~\cite{zipnerf} utilizes hash-grid encoding and contracting space with multi-sampling in conical frustum to achieve better synthesis quality and quicker rendering speed.
\textcolor{black}{For static urban scene reconstruction, methods such as StreetSurf~\cite{streetsurf} and FEGR~\cite{Neuralfieldsmeetexplicit} optimize the signed distance field to better extract mesh. Additionally,~\cite{deformableneuralmesh} and~\cite{LightningNeRF} use point clouds to initialize the geometry and efficiently represent urban scenes. \cite{neo360} investigates the sparse view reconstruction.~\cite{neurad, alignmif} explore miss alignment across different sensors. UniSim~\cite{unisim} uses the feature space to share commonalities between foreground moving actors. DrivingGaussian~\cite{drivinggaussian} and StreetGaussian~\cite{streetgaussian} use different gaussian groups with bounding boxes to represent the background and foreground.}
For scenes with moving objects like vehicles on the road, some methods~\cite{PNF, NSG} try to model moving objects decoupling from the static background by their 3D box.
\textcolor{black}{Recent advancements by SUDS~\cite{suds} and EmerNeRF~\cite{emernerf} have fused 3D and 4D representations to model dynamic urban scenes jointly. While these methods effectively distinguishes between dynamic and static components, they lack the capability to further manipulate the dynamic actors.}
\paragraph{3D object reconstruction}
It is often difficult to obtain a high-quality surface of the object with NeRFs expressed in volume density. Signed Distance Field (SDF) based method~\cite{wang2021neus} who learn the absolute distance from a point to the object surface can reconstruct the object's surface more precisely, and allow us to get the mesh of the objects by MarchingCube~\cite{marching_cubes}. Other method~\cite{geosim} use a learned deformable mesh by differentiable rasterization to make the construction.
More prior knowledge is used to model non-rigid objects like human avatars like SMPL~\cite{loper2015smpl}. \cite{neuralbody} uses NeRF with coordinate mapping from canonical space to the deformable space to reconstruct the avatar from multi-camera videos. \cite{animate-nerf} allow the constructed human NeRF rendered in different poses without finetuning. \cite{selfrecon} extracts the textured mesh by the neural rendering method.

\subsection{3D object generation and insertion}
3D object generation aims to get 3D representation by text-prompt or noises which does not need the ground truth training data like reconstruction methods. 

2D image generation has been well developed. Generative adversarial networks (GANs)~\cite{pix2pix, stylegan} and a variety of
variational autoencoders (VAEs)~\cite{vqvae} have synthesized high-quality images partly because of the massive 2D training data. Recently, latent diffusion models (LDM)~\cite{SD,DALLE2} show strong generative ability for the images from different conditions.

The key to 3D generation is how to make the representation consistent with the view poses. It is hard to generate the 3D models directly for the lack of 3D ground truth data. Many works generate 3d models by lifting the 2D image generation models' knowledge to 3D representation like NeRF through an optimization process. \cite{eg3d, urbangiraffe} use the frozen discriminator in the pre-trained GAN to make the renderings' distribution close to the GAN. \cite{dreamfusion, magic3d} apply score distillation sampling (SDS) who use LDMs to realize text-to-3D. \cite{Texture} texture for a given mesh through inpainting the bare mesh patch-by-patch.

\textcolor{black}{
Recently, several methods have been proposed for inserting vehicles into urban scenes to achieve better consistency. 3D-Aug~\cite{3daug} utilizes the technology of NeRF to elevate object insertion from image copying to 3D rendering.~\cite{lift3d, discoscene, neurallightfield, lightsim} employ a discriminator and a differentiable approach to optimize for more consistent shadows and relighting.}

\section{Simulation system deployment}
\begin{figure*}[h]
\centering
 \includegraphics[width=0.95\linewidth]{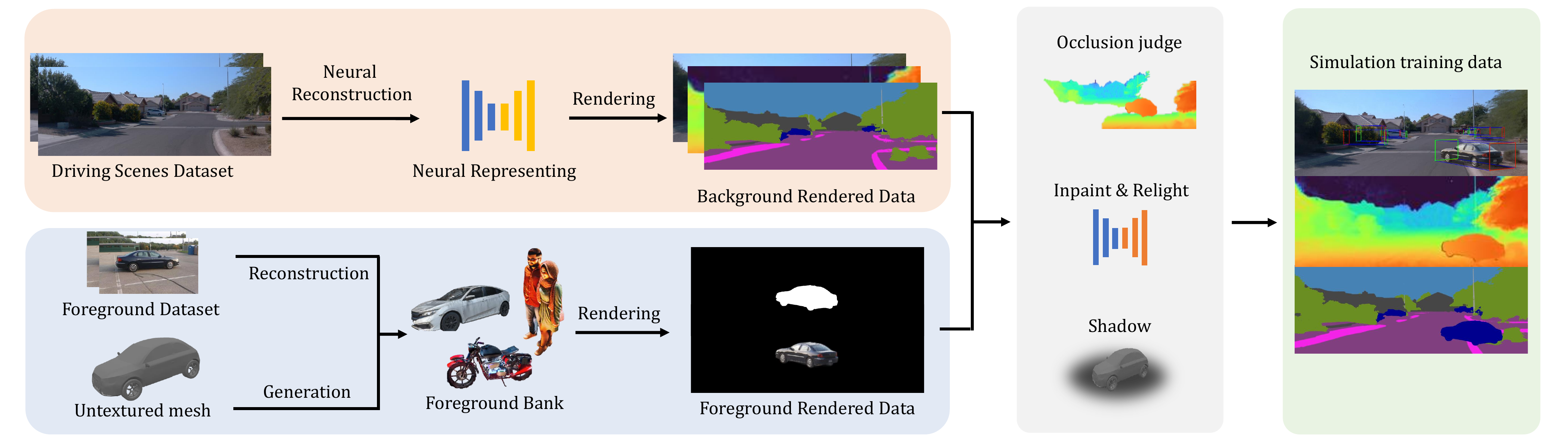}
 \cuthalfcaptionup
 \caption{\color{black} The whole pipeline of our simulation system. We construct our background representation and foreground bank. Given any pose of the camera and object, we render for these two branches separately. Then we judge the occlusion relationship between the foreground and background by depth, process the edge and illumination of the inserted object, render the shadow, and finally obtain the simulation data with labels.}
 \label{fig:pipeline}
\end{figure*}

Our goal is to create a self-driving simulation system that can produce photo-realistic street views. This system will also allow for the manipulation of foreground objects, enabling object insertion, movement, or alteration, by utilizing a comprehensive autonomous driving dataset. This dataset encompasses a wide array of data, including images, precise camera positions, and synchronized LiDAR point clouds. These elements are crucial for accurately reconstructing urban street scenes.

In this section, we first introduce how we construct our simulation asset bank by reconstructing and generating diverse background scenes and foreground objects using a common self-driving dataset. We further propose an automatic foreground object insertion strategy to simplify the manipulation of the foreground objects as well as ensure geometric consistency when rendering novel views. Additionally, we employ a learning-based image refinement module to enhance lighting consistency and overall realism of the generated scenes. This approach is designed to minimize the simulation-to-reality gap, facilitating easier application in real-world conditions.
Our comprehensive strategy enables the creation of a versatile and adaptable simulated environment, crucial for autonomous driving research and capable of accommodating a wide range of environmental conditions.

\subsection{Preliminary}
Neural radiance field (NeRF) represents a scene as a continuous radiance field and learns a mapping function $f: (\mathbf{x}, \mathbf{
\theta}) \rightarrow (\mathbf{c}, \sigma)$. It takes the 3D position $\rm x_i\in \mathbb{R}^3$ and the viewing direction $\theta_i$ as input and outputs the corresponding color $c_i$ with its differential density $\sigma_i$.
The mapping function is realized by two successive multi-layer perceptrons (MLPs).

NeRF uses volume rendering to render image pixels. For each 3D point in the space, its color can be rendered through the camera ray $\mathbf{r}(t) = \mathbf{o} + t\mathbf{d}$ with $N$ stratified sampled bins between the near and far bounds of the distance. The output color is rendered as:
\begin{align}
\setlength{\abovecaptionskip}{4pt}
\setlength{\belowcaptionskip}{4pt}
    \hat{\mathbf I}(\textrm{r}) = \sum_{i=1}^NT_i(1-e^{-\sigma_i\delta_i}) \mathbf c_i, \quad
    T_i = \exp\left( -\sum_{j=1}^{i-1}\sigma_j\delta_j \right) \label{eq:volume_rendering}
\end{align}
where $\mathbf{o}$ is the origin of the ray, $T_i$ is the accumulated transmittance along the ray, $\mathbf c_i$ and $\sigma_i$ are the corresponding color and density at the sampled point $t_i$. $\delta_j = t_{j+1}-t_j$ refers to the distance between the adjacent point samples. 

\subsection{Data preparation}

Reconstruction tasks using existing autonomous driving datasets can be challenging due to their complex nature, as these datasets are typically not tailored for reconstruction purposes.  In this section, we address this challenge by introducing an advanced pipeline designed to reconstruct street views more effectively using the available data. This pipeline incorporates a decoupled camera pose preprocessing step for reconstructing the foreground vehicles and background scenes, a confidence-based dense depth supervision mechanism aimed at enhancing the overall geometric accuracy, and semantic distillation techniques to integrate semantic information into the reconstruction process for more effective manipulation of the scene.

\subsubsection{Camera pose processing}\label{sec:came_pose}
In previous NeRF implementations, structure-from-motion (SfM~\cite{sfm}) techniques often struggle to accurately reconstruct camera poses for self-driving data. This difficulty arises due to limited overlap in camera views. Additionally, the presence of large numbers of moving objects within the scene further complicates the feature-matching process, reducing the success rate of SfM.

\textcolor{black}{For the background reconstruction, we use the camera parameters achieved by sensor-fusion SLAM and IMU of the self-driving cars~\cite{nuscenes2019,Sun_2020_CVPR} and further reduce the inconsistency between multi-cameras with a learning-based pose refinement network. 
We follow Wang \etal ~\cite{wang2021nerfmm} to implicitly learn a pair of refinement offsets $\Delta P = ({\Delta} {R}, {\Delta} T)$ for each original camera pose $P = [R, T]$, where $R\in SO(3)$ and $T \in \mathbb{R}^3$. The pose refine network can help us to ameliorate the error introduced in the SLAM algorithm, thereby enhancing the robustness of our system. 
For moving foreground vehicles, their 3D tracking boxes tend to be sparse in the time series and often lack accuracy. To address this, we first interpolate the box poses at each camera capture timestamp. Similarly, we apply a learnable translation offset $\delta T \in \mathbb{R}^3$ and orientation refinement offset $\delta \theta \in [0, 2\pi]$ for the vehicle's track at each timestamp.}

\subsubsection{Depth} 
To improve the reconstruction quality of the scene, we proposed to use dense depth supervision to guide the model training. We first project the LiDAR point clouds onto the camera's image plane to acquire sparse depth information and then propagate these sparse depths to a dense depth map to supervise the NeRF training. We further create a confidence map to manage and mitigate depth outliers, which serves as a critical component in ensuring the accuracy and reliability of our depth data. Our depth confidence is defined as a learnable combination of the \textbf{perception confidence} and the \textbf{geometry confidence}. During the training, the confidence maps are jointly optimized with the color rendering for each input view. 

\begin{figure*}[t]
\setlength{\abovecaptionskip}{5pt}
\setlength{\belowcaptionskip}{-11pt}
 \centering
\subfigure[Noisy sparse points]{
\includegraphics[width=0.22\linewidth]{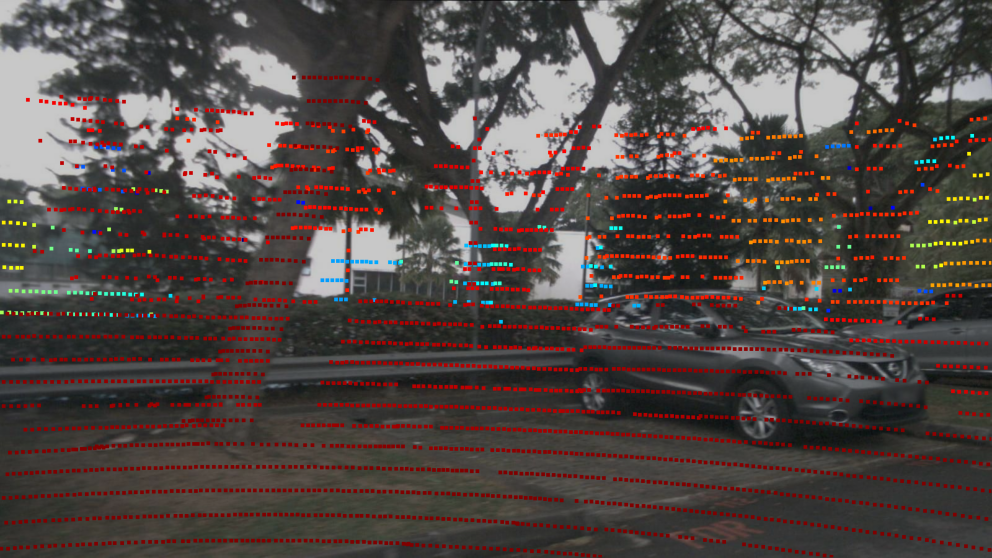}
\label{subfig:sparse_depth}
}
\hspace{-3mm}
\subfigure[Our depth supervision]{
\includegraphics[width=0.22\linewidth]{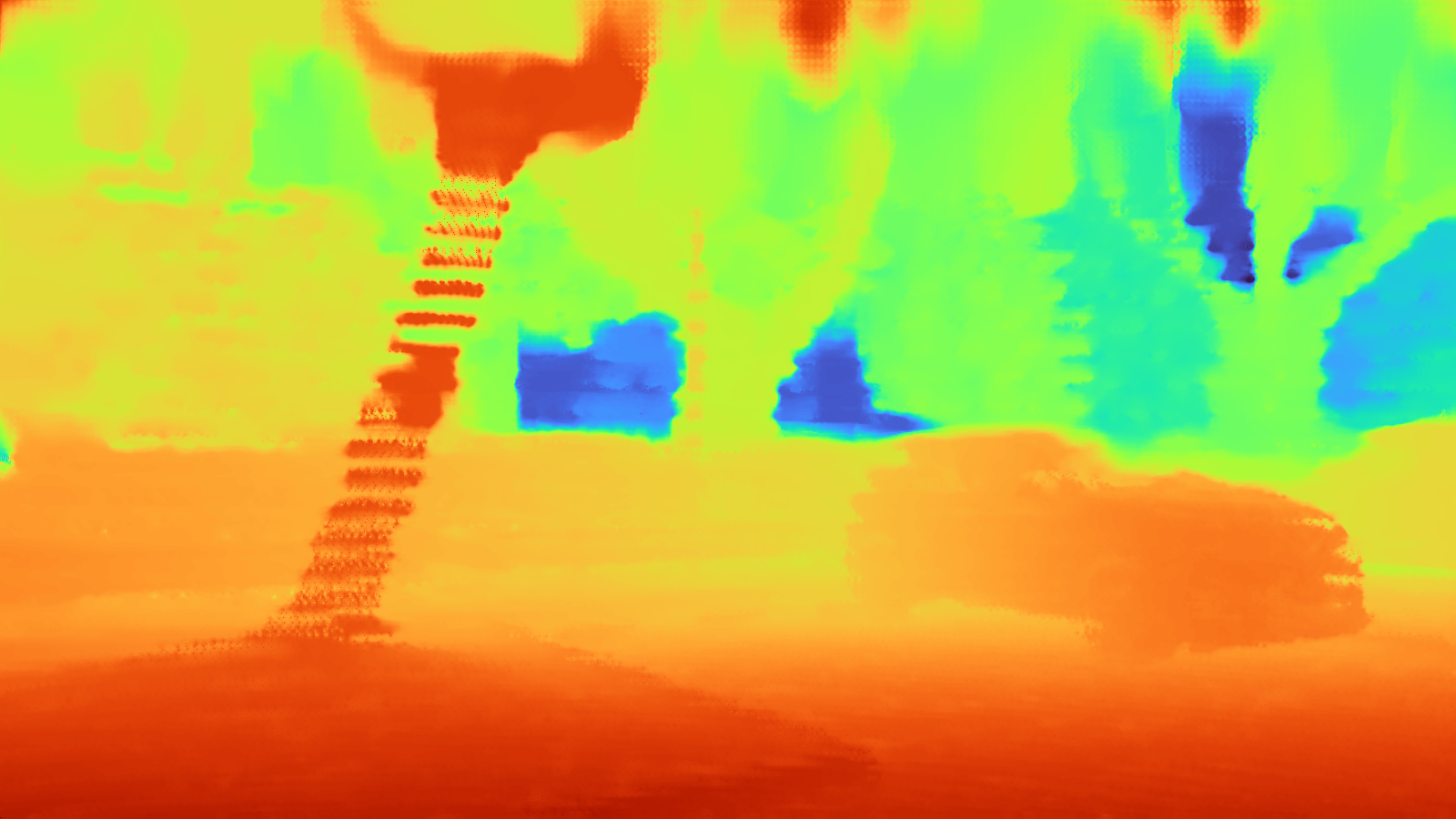}
\label{subfig:dense_depth}
}
\hspace{-3mm}
\subfigure[Learned confidence]{
\includegraphics[width=0.22\linewidth]{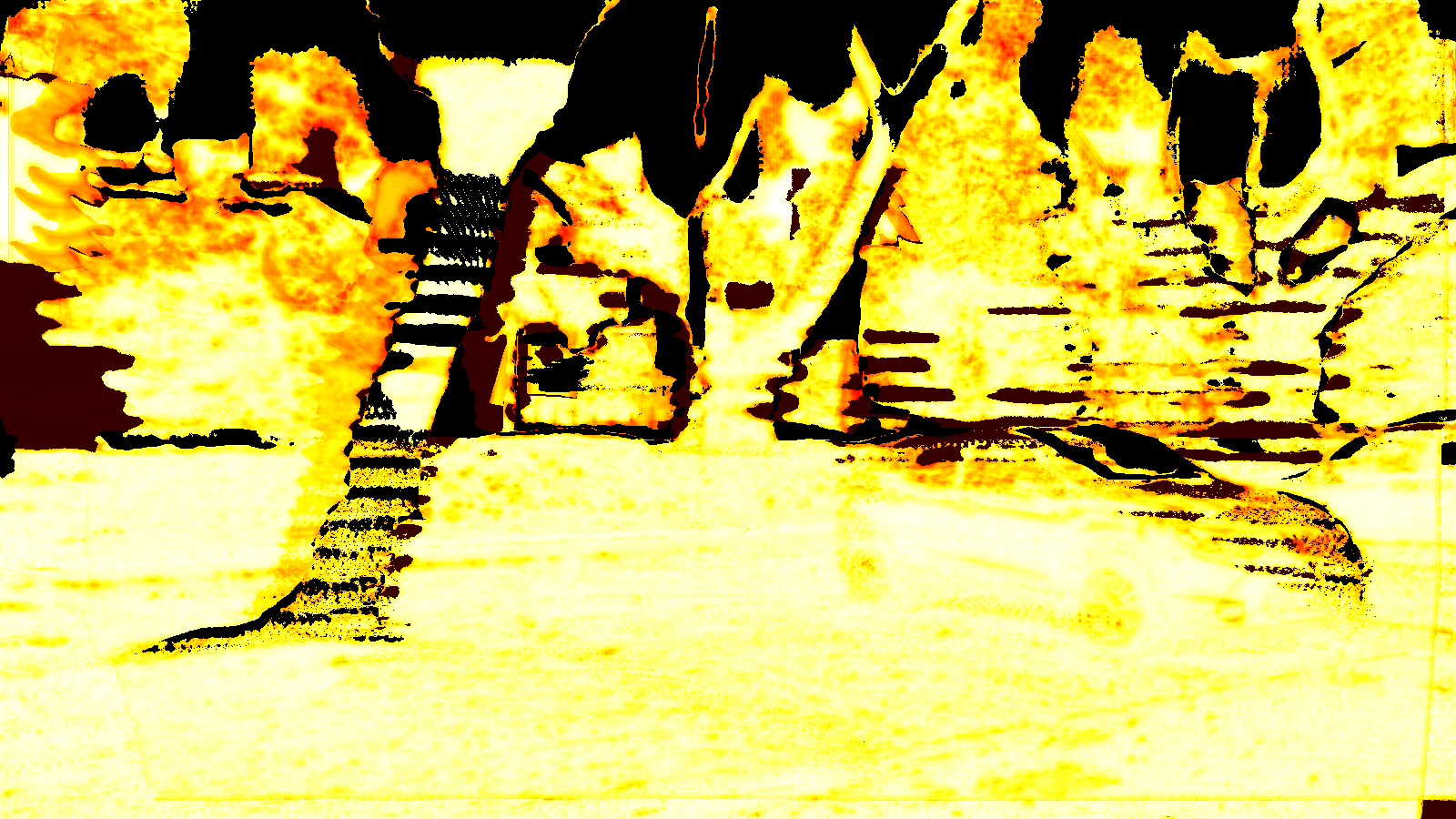}
\label{subfig:confidence}
}
\hspace{-3mm}
\subfigure[Our depth rendering]{
\includegraphics[width=0.22\linewidth]{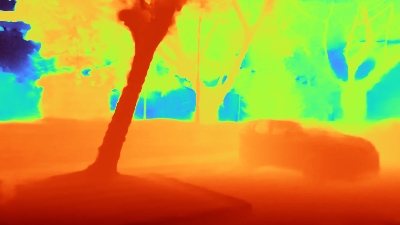}
\label{subfig:our_depth}
}
 \cuthalfcaptionup
 \caption{Our depth supervision with confidence maps}
 \label{fig:overview}
\end{figure*}

We use NLSPN~\cite{park2020non}  to propagate the sparse depth to surrounding pixels. Although NLSPN performs well with 64-channel LiDAR data (\eg\  KITTI~\cite{Geiger2012CVPR, Geiger2013IJRR} dataset), it doesn't generate good results with nuScenes' 32-channel LiDAR data, which are too sparse for the depth completion network. As a result, we accumulate neighbor LiDAR frames to get much denser depths for NLSPN. These accumulated LiDAR data, however, contain a great number of outliers due to the moving objects, ill poses and occlusions, which give wrong depth supervision. To address this problem, we design a robust confidence measurement that can be jointly optimized while training our NeRF as illustrated in Figure~\ref{fig:overview}. 

Our depth confidence is defined as a learnable combination of the \textbf{perception confidence} and the \textbf{geometry confidence}. During the training, the confidence maps are jointly optimized with the color rendering for each input view.

\paragraph*{Perception confidence} \label{sec:perception confidence}
To measure the accuracy of the depths and locate the outliers, we first use a warping operation $\psi$ to reproject pixels $\mathbf{X} = (x,y,d)$ from the source images $I_s$ to the target image $I_t$. Let $P_s$, $P_t$ be the source and the target camera parameters, $d\in \mathcal{D}_s$ as the depth from the source view. The warping operation can be represented as:

\begin{align}
    \mathbf{X}_t = \psi(\psi^{-1}(\mathbf{X}_s, P_s),~P_t)
\label{eq:reproject}
\end{align}
$\psi$ represents the warping function that maps 3D points to the camera plane and $\psi^{-1}$ refers to the inverse operation from 2D to 3D points. 
Since the warping process relies on depth maps $\mathcal{D}_s$, the depth outliers can be located by comparing the source image and the inverse warping one.  
We introduce the RGB, SSIM~\cite{ssim}  and the pre-trained VGG feature~\cite{vgg} similarities to measure the perception confidence in the pixel, patch structure, and feature levels:
\begin{equation*}
    \mathcal{C}_\text{rgb} = 1 - |\mathcal{I}_s - \mathcal{\hat{I}}_s|,\quad
\mathcal{C}_\text{ssim} = \text{SSIM}(\mathcal{I}_s, \mathcal{\hat{I}}_s)),
\end{equation*}
\begin{equation}
    \mathcal{C}_\text{vgg} = 1 - ||\mathcal{F}_s-\hat{\mathcal{F}_s}||.
\end{equation}
Where $\mathcal{\hat{I}}_s = \mathcal{I}_t(\mathbf{X_t})$ is the warped RGB image, and the 
$\hat{\mathcal{F}_s}=\mathcal{F}_t(\mathbf{X_t})$ refers to the feature reprojection. 
The receptive fields of these confidence maps increase from pixels and local patches to non-local regions. This gradual expansion helps in building strong and reliable measurements of confidence.

\paragraph*{Geometry confidence} \label{sec:geometry confidence}
We further impose a geometry constraint to measure the geometry consistency of the depths and flows across different views. Given a pixel $\mathbf{X}_s = (x_s, y_s, d_s)$ on the depth image $\mathcal{D}_s$ we project it to a set of target views using \eqref{eq:reproject}.
The coordinates of the projected pixel $\mathbf{X}_t = (x_t, y_t, d_t)$ are then used to measure the geometry consistency. For the projected depth $d_t$, we compute its consistency with the original target view's depth $\hat{d}_t=D_t(x_t, y_t)$:

\begin{align}
     \mathcal{C}_{depth} =  \gamma(|d_t - \hat{d}_t)|/d_s),\ 
     \gamma(x) = \left
    \{
    \begin{array}{cc}
    0,       &
    \text{if}~~x \geq \tau,\\
    1 - {x}/{\tau}, & \text{otherwise.}
    \end{array}
    \right.
    \label{eq:depth consistency}
\end{align}
For the flow consistency, we apply off-the-shelf optical flow prediction model~\cite{Zhang2021SepFlow} to compute the pixel's motions from the source image to the adjacent target views $f_{s\rightarrow t}$. This approach could further reduce inconsistency by aligning the pixel movements across different frames, ensuring a more coherent and seamless transition of depth. The flow consistency is then formulated as:

\begin{align}
     \mathcal{C}_{flow} =  \frac{\gamma\|\Delta_{x,y} - f_{s\rightarrow t}(x_s, y_s) \|}{\|\Delta_{x,y}\|},
     \Delta_{x,y} = (x_t - x_s, y_t-y_s).
     \label{eq:flow consistency}
\end{align}
Where $\tau$ is a threshold in $\gamma$ to identify the outliers through the depth and flow consistencies.

\begin{figure*}[h]
\centering
 \includegraphics[width=1\linewidth]{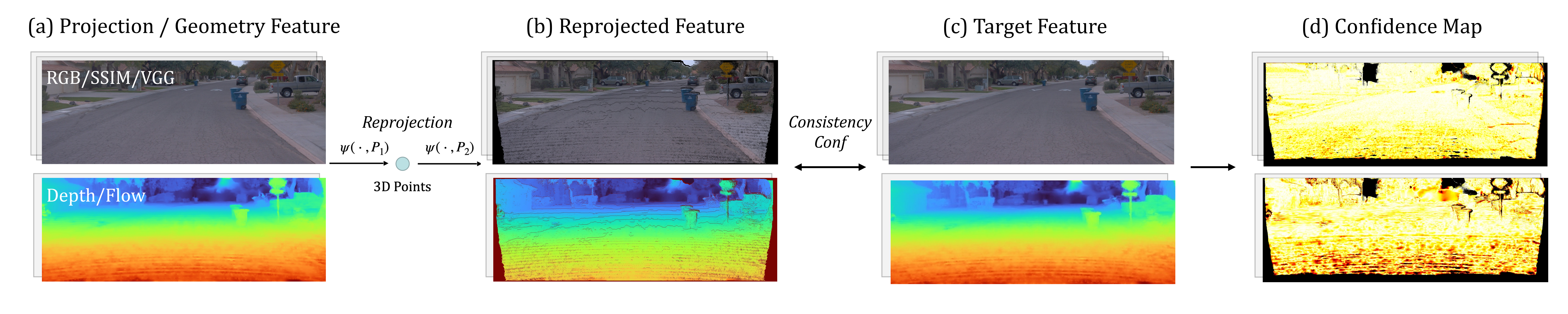}
 \cuthalfcaptionup
 \hspace{-3mm}
 \caption{Illustration of the confidence computation process.}
 \label{fig:confidence_process}
\end{figure*}
\begin{figure*}[h]
\centering
 \setlength{\abovecaptionskip}{6pt}
 \setlength{\belowcaptionskip}{-5pt}
 \includegraphics[width=1\linewidth]{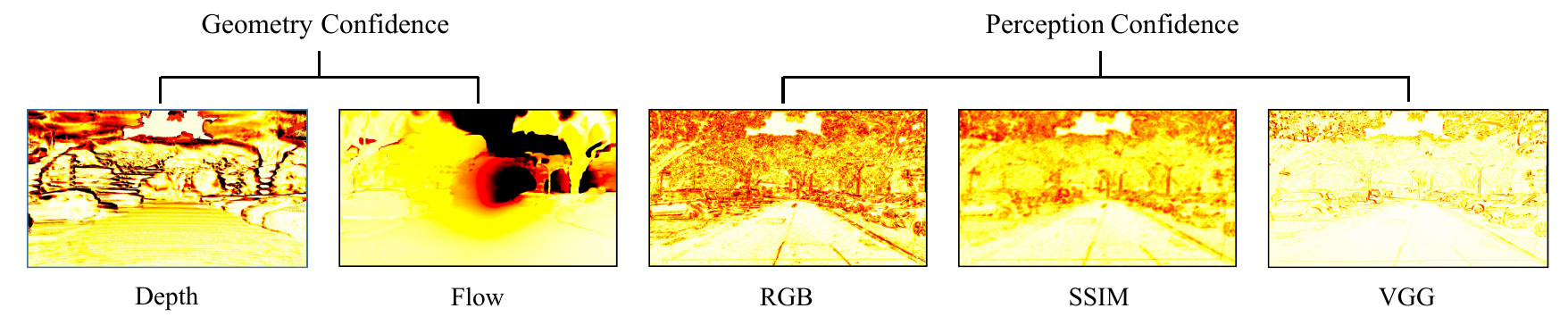}
 \cuthalfcaptionup
 \caption{Visualization of each confidence component. Brighter regions indicate higher confidence. Geometry confidences (flow and depth) represents the geometry consistency. The perception confidence measure the photometric, local structure and feature consistency.}
 \label{fig:confidence_maps}
 \vspace{-2mm}
\end{figure*}
\paragraph*{Computation of confidence} Figure \ref{fig:confidence_process} shows the process of how we generate the confidence maps. We compute the depth confidence by reprojection the depth maps to other views and measure the confidence level based on the reprojection and geometry consistency. 
Specifically, given the LiDAR positions of a set of consecutive frames $[P_{t-1}, P_t, P_{t+1}]$ at different time $[t-1, t, t+1]$, we project the 3D LiDAR points at time $t-1$ and $t+1$ to $t$ by $\Delta P$ between $P_t$ and $P_{t-1}/P_{t+1}$. This is similar to the mapping function $\psi$ in Eq. \eqref{eq:reproject}.  Most outliers of the moving objects and other regions can be removed by the consistency check using the optical flow and the depth projection (Eq.\eqref{eq:depth consistency} and \eqref{eq:flow consistency}). The rest outliers can also be handled by the proposed confidence-guided learning.
In Figure \ref{fig:confidence_maps}, we visualize different confidence components. Depth and optical flow confidence maps focus on geometry consistency between adjacent frames, while RGB, SSIM and VGG confidence maps compute the photometric consistency from pixel, structure, and feature level.
\paragraph*{Learnable confidence combination}
To compute the final confidence map, we assign learnable weights $\omega$ for each confidence metric  and jointly optimize them during the training.  The final confidence map can be learned as $\Hat{\mathcal{C}} = \sum_{i} \omega_i \mathcal{C}_i\text{, where } \sum_i\omega_i = 1$.
The $i \in \{rgb, ssim, vgg, depth, flow\}$ represents the optional confidence metrics. The learnable weights $\omega$ adapt the model to automatically focus on correct confidence. One thing to note is that we predict a sky map and set the depth of the sky at a sufficiently large value.

\subsubsection{Semantic labels}
The semantic labels are crucial to the scene understanding in autonomous driving environments, serving for the subsequent decision-making and control. Here, we first consider the most difficult cases (\eg~nuScenes scenes) where there is no annotated label from the collected driving data (images and LiDAR points). Given unlabeled real images collected from multiple cameras of the self-driving cars, 
we train a SegFormer~\cite{segformer} model, on the mixture of other datasets including Cityscapes ~\cite{cordts2016cityscapes}, Mapillary~\cite{neuhold2017mapillary}, BDD~\cite{yu2020bdd100k}, IDD~\cite{varma2019idd} to compute weak labels that serve as inputs to the NeRF reconstruction model.  To achieve better cross-dataset generalization  of the SegFormer and avoid conflicts in the label definition, we utilize the learning settings and label merging strategy in the ~\cite{lambert2020mseg}. 

Considering that the generated weak labels may have many outliers, to avoid the influence of the outliers and generate more accurate 3D point labels, we take full use of  multi-view geometric and video spacial temporal consistency in our NeRF reconstruction. 

In some other cases, when there is a small number of labeled images or LiDAR frames, we can also leverage the existing ground-truth labels for more robust label generation. For example, in the nuScenes dataset, a small part of the LiDAR frames (about 1/10) was labeled with semantic annotations.  We take the sparse 3D point labels along with the weak 2D image labels to learn more accurate semantic radiance fields.

\subsection{Representation of street scenes}\label{sec:representation}

Recent advances in neural rendering techniques have enabled efficient and high-quality scene reconstruction through versatile representation.
\textcolor{black}{Zip-NeRF~\cite{zipnerf} ingeniously integrates hash-grid and frustum-encoding methods to enhance scene reconstruction techniques. It employs a multi-scale hash grid as the feature backbone and adjusts the weights of the feature in different hash-grid scales based on the point's cone radius $\dot{r}t$ along the ray  which is named $downweighting$. Additionally, it $multisamples$ and aggregates the features within the truncated cone around the sample point to obtain direction-aware and scale-aware features, which are then passed to the MLP decoder.}

\begin{figure*}[h]
\centering
 \includegraphics[width=0.95\linewidth]{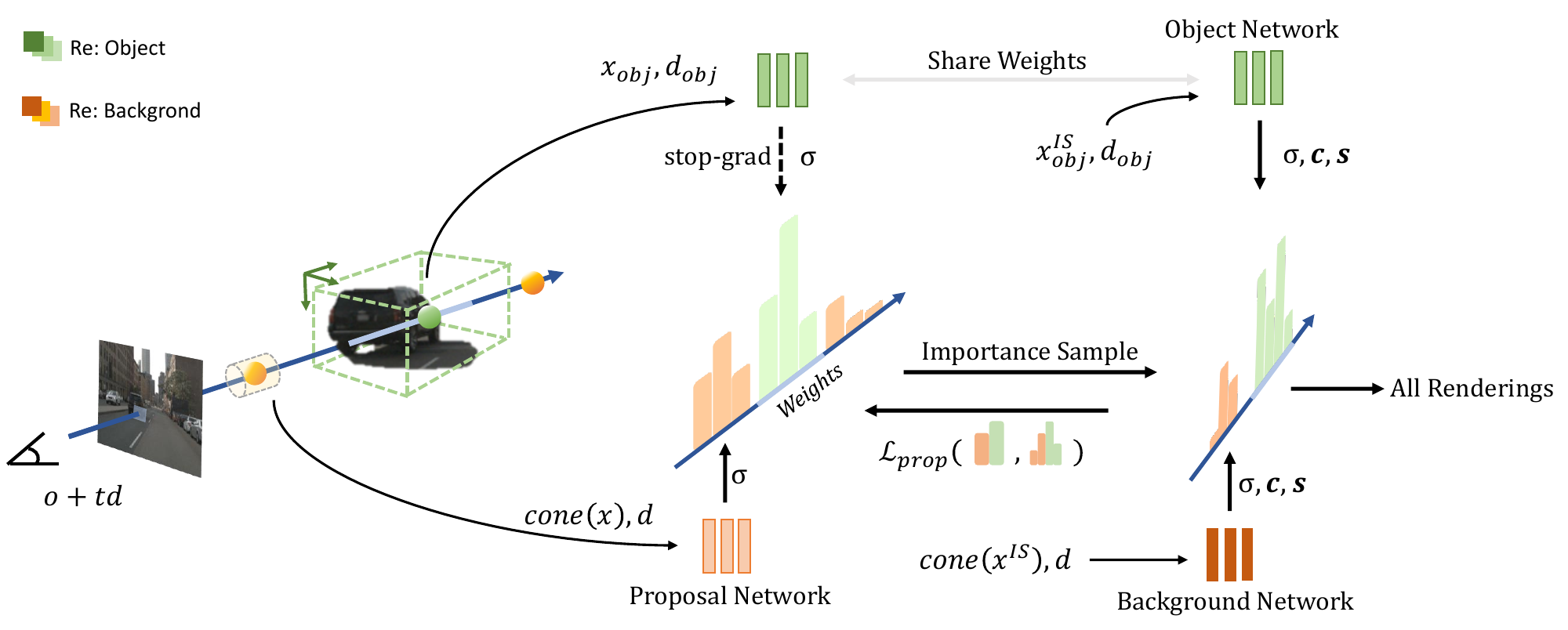}
 \cuthalfcaptionup
 \caption{
 \textcolor{black}{Cascade sampling strategy to reconstruct the background and moving objects together. The object network whose weights are shared through all sampling stages to decode volume rendering attributes when the point is in the corresponding object bounding box while background/proposal network to decode the point not in any box. The rendering weights in the proposal stage will guide the sampling points in the next stage and will be only supervised by the rendering weights in the next stage by $\mathcal{L}_{prop}$. Notice that the gradient will not pass to the object network in the proposal stage.}
 }
 \label{fig:mv_bg_recon}
\end{figure*}

However, models designed for static scenes are less effective in outdoor environments with dynamic elements like cars, trucks, buses, and pedestrians. These moving objects can hinder the learning process of scene representation. Inspired by~\cite{NSG,PNF}, we model the dynamic objects as individual instances and integrate the learning process of foreground and background during the scene reconstruction. This approach allows for the comprehensive reconstruction of scenes, incorporating both moving and static elements.

\textcolor{black}{We then use an instance-specific object network to query the features and decode the color, semantic logits, and density for the point in a certain object box. If the point is not within any object box, we obtain these attributes through the background network.}
\textcolor{black}{We describe $i^{th}$ object as a series of time-dependent 3D bounding box with 6DoF parameters $[\textbf{R}_{i, t}, \textbf{T}_{i,t}]$ in the background world and a size ${\textbf{dim}}_i = [w_i, h_i, l_i]^{\top}$. We track all the objects in a training scene. We can easily check if one ray is intersected with a certain object box at time $t$, and for each sampled point on the ray, we decode its color and density by two branches. If the point is in $i^{th}$ object box, we transform the point position $\textbf{x}$, ray direction $\textbf{d}$ and ray cone radius $r$ at depth $t$ to the uniformed object box coordinate system from the background world coordinate system by 
\begin{equation}
\label{eq:scene_param}
    \textbf{x}_o = \textbf{R}_{i, t}^{\top}(\textbf{x}-\textbf{T}_{i,t})\cdot \frac{1}{{\textbf{dim}}_i}, \quad \textbf{d}_o = \textbf{R}_{i, t}^{\top}\textbf{d}, \quad r_o = \frac{r}{\overline{{\textbf{dim}}_i}}
\end{equation}
}
\textcolor{black}{Following~\cite{zipnerf}, we use a cascade sampling strategy to reconstruct the background and moving objects together. For each sample point on the ray casting from $o$ with direction $d$, we use an object network whose weights are shared through all sampling stages to decode volume rendering attributes when the point is in the corresponding object bounding box while background/proposal network to decode the point not in any box. Notice that if the sampling point is not in the final stage, the network only outputs density $\sigma$ and the gradient will not pass to the object network. Rendering weights computed by the proposal stage will be only supervised by the rendering weights in the next stage by $\mathcal{L}_{prop}$. Each network is composed of a multi-scale hash grid and an MLP decoder. For better anti-aliasing rendering, we follow~\cite{zipnerf} and aggregate the hash grid features in the cone for the background points before passing them to the MLP while only using downweighting features for the object points. Figure~\ref{fig:mv_bg_recon} illustrates the combination process of moving objects with the static background in our reconstruction method.}

\begin{figure*}[h]
\centering
 \includegraphics[width=0.95\linewidth]{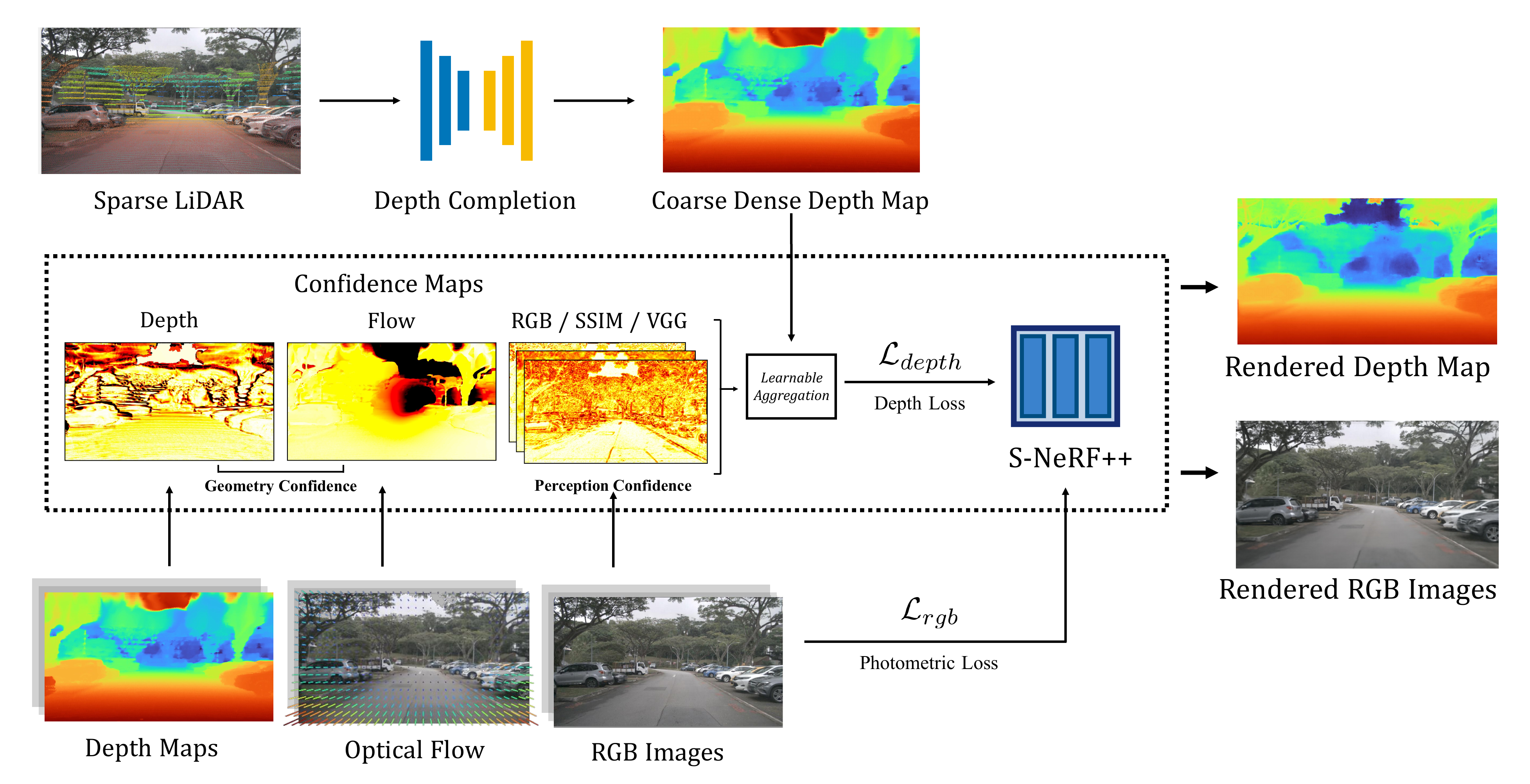}
 \cuthalfcaptionup
 \caption{Overview of our S-NeRF++ reconstruction framework supervised by generated dense depth maps with learnable confidence maps.}
 \label{fig:model_overview}
\end{figure*}
The overview of our reconstruction method is shown in Figure \ref{fig:model_overview}. We first propagate the sparse LiDAR points into a dense depth map and compute the geometry and the perception confidence maps. We use learnable combinations to achieve the final confidence maps  to reduce the influence of depth outliers.

\subsection{Reconstruction learning}
In this part, due to the extreme complication of the driving scenes, we demonstrate all the effective losses including the supervision and regularization for the training process of reconstruction. The overall loss, represented as $\mathcal{L}$, is formulated as follows:
\begin{equation}
    \mathcal{L} = \mathcal{L}_{sup}+\mathcal{L}_{reg}.
\end{equation}

The supervision loss, denoted as $\mathcal{L}_{sup}$, can be expressed as:
\begin{align}
&\mathcal{L}_{sup} = \mathcal{L}_{rgb}+\lambda_{d} \mathcal{L}_{depth}+\lambda_{L}\mathcal{L}_{pc}+\lambda_{s} \mathcal{L}_{sem}. \\
&\mathcal{L}_{rgb} = |\hat{\mathbf{I}}-\mathbf{I}|, \quad \mathcal{L}_{depth} = \mathcal{\hat{C}}|\mathcal{\hat{D}}-\mathcal{D}|, \\
&\mathcal{L}_{pc} = \sum_{\mathbf{r}\in \mathcal{R}_{L}} |\mathcal{\hat{D}}(\textrm{r})-\mathcal{D}_L(\textrm{r})|.
\end{align}
In our loss expression, we simplify the notation by omitting the summation over rays $\sum_{\mathcal{R}}(\cdot)$. Although, in practice, we sample $|\mathcal{R}|$ rays ays in each training step. We employ an RGB L1 loss, denoted as $\mathcal{L}_{rgb}$, and a disparity L1 loss with masking, represented as $\mathcal{L}_{depth}$. In this context, $\mathcal{\hat{C}}$ denotes the weights of the learnable confidence map, $\hat{\mathcal{D}}$ refers to the disparity map rendered by our reconstruction network, $\mathcal{D}$ indicates the inversion of dense depth map propagated from the sparse LiDAR data. 

Furthermore, we use sparse point clouds $\mathcal{R}_L$ as the ground truth depth of a ray $\textrm{r}$ captured by the LiDAR sensor. To enhance geometric accuracy, we include a disparity loss term, $\mathcal{L}_{pc}$, which aligns with the sparse point cloud data. This approach helps in refining the geometry in our reconstructions.

For our reconstruction backbone outputs semantic label logits $\mathbf{s_i}$, we can calculate the semantic label of a ray by following Eq. \ref{eq:volume_rendering}:
\begin{align}
    \hat{\mathbf S}(\textrm{r}) = \sum_{i=1}^NT_i(1-e^{-\sigma_i\delta_i}) \mathrm{Softmax}(\mathbf s_i).\label{eq:volume_rendering_s}
\end{align}
The semantic loss can be calculated as the cross entropy loss of the rendered semantic map $\hat{\mathbf{S}}$ and the supervision semantic label $\mathbf{S}$:
\begin{equation}
    \mathcal{L}_{sem} = \mathrm{CrossEntropy}(\hat{\mathbf{S}}, \mathbf{S})
\end{equation}
It is important to note that the $\sigma_i$ is stopped-gradient with respect to the reconstruction network when we calculate $\mathcal{L}_{sem}$. This approach is adopted to prevent the influence of potentially inaccurate semantic information on the geometric aspects of the model. By doing so, we ensure that the geometry remains unaffected by any errors in the semantic data.

Regularization loss $\mathcal{L}_{reg}$ is consist of distortion loss $\mathcal{L}_{dt}$, anti-aliased interlevel loss $\mathcal{L}_{prop}$ and edge-aware smoothness loss $\mathcal{L}_{smo}$ with weights $\lambda_{dt}$, $\lambda_{prop}$ and $\lambda_{smo}$. We follow~\cite{zipnerf} to set up $\mathcal{L}_{dt}$ and $\mathcal{L}_{prop}$, and $\mathcal{L}_{smo}$ is implemented as
\begin{equation}
    \mathcal{L}_{smo} = |\partial_{x}\Hat{\mathcal{D}}|\ \exp^{-|\partial_{x}I|} + |\partial_{y}\Hat{\mathcal{D}}|\ \exp^{-|\partial_{y}I|}.
\end{equation}
The whole regularization term can be written as:
\begin{equation}
    \mathcal{L}_{reg} = \lambda_{dt} \mathcal{L}_{dt}+
    \lambda_{prop}\mathcal{L}_{prop}+\lambda_{smo}\mathcal{L}_{smo}.
\end{equation}

\subsection{Foreground augmentation}

Foreground object manipulation plays a crucial role in the system of driving scene simulation, which requires an instance-aware representation of 3D objects, considering that the actions of foreground objects are critical for the driving scenario. 
Section \ref{sec:representation} introduces our method to reconstruct moving vehicles through street view data, which will be part of our foreground bank. Reasonable use of various open-source data sets and generative models on the network can greatly increase the types and quantities of foreground objects.
To enrich the foreground assets as much as possible, we apply two strategies: the \textbf{generation-based} simulation and the \textbf{reconstruction-based} simulation. The generation-based approach utilizes a pre-trained diffusion model to guide the generation process, while the reconstruction approach leverages a neural rendering-based method to represent foreground objects and reconstruct meshes.

\paragraph{Generation-based Approach}

Diffusion models are latent variable models of the form $ p_{\theta}(\textbf{x}_0) = \int p_{\theta} (\textbf{x}_{0:T}) d\textbf{x}_{1:T}$. $\textbf{x}_{1:T}$ are the latent variables following the posterior distribution of a Markov chain with Gaussian
transition. The posterior $q(\textbf{x}_{1:T}| \textbf{x}_0) = \prod_{t=1}^{T} q(\textbf{x}_t|\textbf{x}_{t-1})$ is called forward process. The learned distribution  $p_\theta(\textbf{x}_{0:T})=p_{\theta}(\textbf{x}_T)\prod_{t=1}^{T}p_{\theta}(\textbf{x}_{t-1}|\textbf{x}_t)$ is also approximated as a Gaussian transition Markov chain which is known as reverse process. After we maximize the ELBO of $p_{\theta}(\textbf{x}_0)$, we can sample $\textbf{x}_0$ from a normal noise by the reverse process. A \textbf{latent diffusion model} is a diffusion model not directly calculating the distribution 
of $\textbf{x}_0$ but its latent codes $\textbf{z}_0$. It utilizes an autoencoder's encoder to map $\textbf{x}_0$ to the latent space $\textbf{z}_0$, and uses its decoder to decode the latent codes $\textbf{z}_0$ to $\hat{\textbf{x}}_0$. Recently, StableDiffusion~\cite{SD} and other LDM show powerful 2D image generation capabilities. However, it is hard to utilize 2D LDM knowledge for better 3D generation. Dreamfusion~\cite{dreamfusion} raises a method that builds the bridge of 3D and 2D images named \textbf{score distillation sampling}. It aims to minimize the KL devergence of $q(\textbf{z}_t|g(\phi),y,t)$ and $p_\theta(\textbf{z}_t|y,t)$. It has a simplified calculation version of the grad of $\phi$:
\begin{equation}
\label{eq:sds}
\nabla_\phi \mathcal{L}_{\mathrm{SDS}}(\theta, \mathbf{z}\!=\!g(\phi)) \!=\! \mathbb{E}_{t, \epsilon}\! \left[w(t)\! \left(\hat{\epsilon}_\theta \! \left(\mathbf{z}_t ; y, t\right) \!-\! \epsilon\right)\! \frac{\partial \mathbf{z}}{\partial \phi}\right].
\end{equation}
Where  $\hat{\epsilon}_\theta$ is the fixed pre-trained LDM U-net module, $y$ is the condition of LDM. In 3D generation, $g(\phi)$ can be the differentiable rendering process of an object. The LDM and SDS allow us to manipulate the texture of a mesh or directly generate 3D models in different representations. 

Given a 3D mesh with its category, we use input text to control the texture of the mesh represented as an atlas through a pre-computed UV mapping. \cite{Texture} proposed a pipeline of painting texture through a pre-trained depth-conditioned diffusion model and inpainting diffusion model, while the generated textures are always mismatched and lack realism, especially painting textures for cars and bicycles, which are highly concerned objects in autonomous driving. To get high-fidelity 3D objects for our foreground bank, we fine-tuned the diffusion model with category-specific data to learn the data distribution prior better and then guide the generation processing. To generate a texture with the ensured quality we further refine the UV map through the Score Distillation Sampling process.

Due to the inherent flexibility of non-rigid human bodies, we implement skinning techniques for avatar meshes, allowing us to create diverse postures either through a text-guided generative method or pre-programmed motions. This enables us to render humans in various poses using the skinned, textured mesh along with specified pose parameters.

\paragraph{Reconstruction-based Approach}
Unlike the former which utilizes pre-constructed meshes, this approach reconstructs meshes and renders novel images of existing objects via a NeRF-like method. Recently, methods based on signed distance function including NeuS~\cite{wang2021neus} and VolSDF~\cite{yariv2021volume} are preferred for mesh reconstruction and novel rendering as they combine SDF (Sign Distance Function, from which can be extracted mesh easily) and NeRF (which performs high qualities in novel rendering) together. Here, we use NeuS~\cite{wang2021neus} with the accelerating technique of hash encoding presented in Instant-NGP~\cite{Instant-NGP} as the reconstruction method. Given multi-view object-centered images of an object, we utilize COLMAP~\cite{sfm} to extract camera matrices and fit the object's SDF through NeuS~\cite{wang2021neus}. Then we can extract mesh from the reconstructed SDF via the Marching Cubes algorithm~\cite{marching_cubes} and render images based on NeuS~\cite{wang2021neus} render equation. Foreground masks can be derived from NeuS rendering and mesh rasterization. Compared with the generation approach, the reconstruction approach costs more computation but provides a more realistic representation of foreground objects. 

It should be noted that we can render both the RGB and the foreground mask using volume rendering. Alternatively, we can extract their textured meshes and achieve renderings through rasterization, especially since the foreground is reconstructed using an SDF-based method.

For human reconstruction, we follow another neural rendering-based approach {SelRecon}~\cite{selfrecon} which utilizes the prior knowledge of the human body as much as possible to reconstruct the mesh and texture of the avatar from the video with a person in motion.

\subsection{Composition of background and foreground}

\begin{figure*}[h]
\centering
 \includegraphics[width=0.95\linewidth]{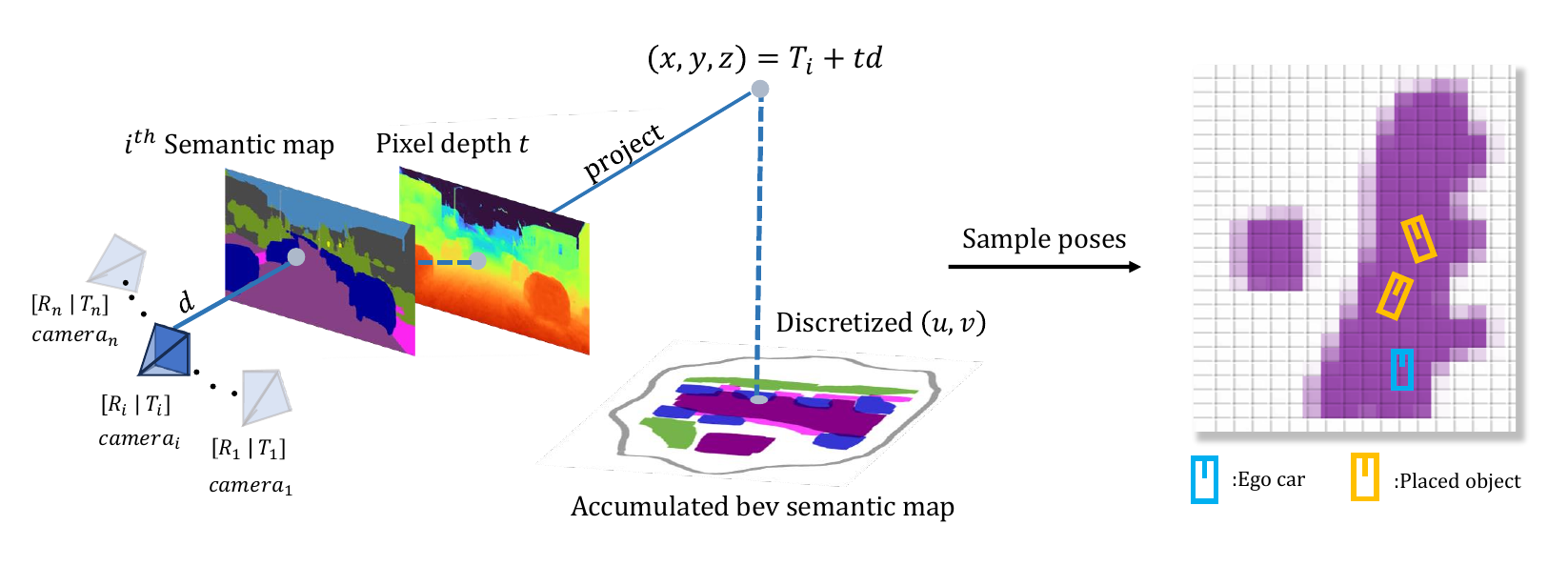}
 \cuthalfcaptionup
 \caption{\color{black} The procedure of sampling foreground object's placement. We project 3D semantic point cloud from rendered semantic and depth maps to explicitly represent the reconstructed scene. Then we vertically project it to 2D ground plans, recording semantic labels and height of each grid. Finally, we randomly sample placement positions in valid area (The purple part in ground plan above represents the placeable area).}
 \label{fig:fg_placement}
\end{figure*}

After reconstructing background scenes and object-center foreground objects, we propose a straightforward method to synthesize simulated data, including RGB images, dense depth maps, semantic labels, and 3D detection annotations. The overview of our simulation system is in Figure \ref{fig:pipeline}. First, we construct a BEV-aware ground plan based on rendered background depth maps and semantic labels to determine the placement of foreground objects. Given the computed render poses, the corresponding foreground RGB figures are rendered with masks and aligned meshes. Foreground objects are then merged in background scenes producing combined RGB figures and annotations. At last, boundary inpainting, illumination handling, and shadow generating are performed to make synthetic figures more realistic.

\subsubsection{Foreground placement based on 2D ground plan}
\textcolor{black}{To place foreground objects in sensible regions, we need a global BEV map to determine suitable placement areas. In cases like nuScenes, where a high-quality BEV map with detailed information on dynamic actors is available, we sample poses from the historical traces of various vehicles, apply perturbations, and verify their validity by checking factors such as lane direction and potential collisions.
}

When a BEV map is not available, we construct a rough BEV map ourselves by leveraging rendered background data to create a BEV-aware ground plan with semantic annotations, and we randomly select placement locations within the available regions. Given rendered dense depth maps $\mathbf{D}_i$, semantic maps $\mathbf{S}_i$ and camera matrices $\mathbf{P}_i$, a 3D semantic point cloud $\mathbf{C}$ can be constructed by inverse projecting 2D 
pixel-level semantic label into 3D space:
\begin{align}
\setlength{\abovecaptionskip}{4pt}
\setlength{\belowcaptionskip}{4pt}
    \mathbf{C}= \bigcup_{i=1}^{K} \pi^{-1}\left(\mathbf{S}_i, \mathbf{D}_i, \mathbf{P}_i\right) = \{ (\mathbf{x}_j, s_j) \}_{j=1}^{N} 
\label{eq:3D point cloud}
\end{align}
where $\pi^{-1}$ means inverse projection, $\mathbf{x}_j \in \mathbb{R}^3$ and $s_j \in \mathcal{S}$ represents the coordinates and semantic label of point $j$. 

Point cloud $\mathbf{C}$ represents reconstructed background scenes with semantic information, but it is not suitable for placement planning because of its huge storage cost (the number of points $N$ can be huge) and redundancy (points are too compact). Therefore, we first randomly drop a large proportion of points and modify the point cloud to a grid-based ground plan. To be more detailed, we set $H \times W$ 2D ground grids and project semantic point cloud $\mathbf{C}$ vertically to determine the semantic label of each grid, providing a semantic ground plan $\mathbf{G}_s$. Similarly, we construct another ground plan $\mathbf{G}_h$ with the same size to record the height of each grid. Each plan can be represented as a $H \times W$ array (i.e. a matrix). To determine placement, we randomly choose grids on available regions in $\textbf{G}_s$ and query corresponding height values in $\textbf{G}_h$. The available regions can be defined by the user. Fig \ref{fig:fg_placement} illustrates the placement sampling procedure.

For some cases, such as nuscenes, a high-quality BEV map with detailed information on dynamic actors is provided, we can easily sample poses from the historical traces of various vehicles, apply perturbations, and verify their validity by checking factors such as lane direction and potential collisions.

\subsubsection{Image synthesis}

After determining the placement, we derive the camera pose to render the foreground object, giving RGB figure $\mathbf{I}$, foreground mask $\mathbf{M}$, with a mesh that has been reconstructed in former parts. The composition of background and foreground can be approached through occlusion-aware pasting, and various annotations are extracted from mesh and mask. Explicitly, we first place mesh under the camera coordinate to align with the figure $\mathbf{I}$, then apply rasterization to compute the depth map of the foreground object $\mathbf{D}_f$. Based on $\mathbf{D}_f$ and background depth map $\mathbf{D}_g$, we can derive occlusion mask $\hat{\mathbf{M}}$ by comparing two depth value of each pixel, i.e. $\hat{\mathbf{M}} = \mathbf{M} \cdot \left( \mathbf{D}_f < \mathbf{D}_g \right)$. 2D synthetic data including RGB figures, semantic labels, and depth maps are composited through pasting corresponding data concerning mask $\hat{\mathbf{M}}$; 3D annotations like bounding boxes can be extracted from the aligned mesh. 

It is noted that $\mathbf{D}_f$ can be also computed through the NeRF-based methods in the render stage. We choose the mesh-based approach to be consistent with the 3D annotations in the detection task.

\subsubsection{Object insertion refinement}

\begin{figure*}[th]
\centering
 \includegraphics[width=0.95\linewidth]{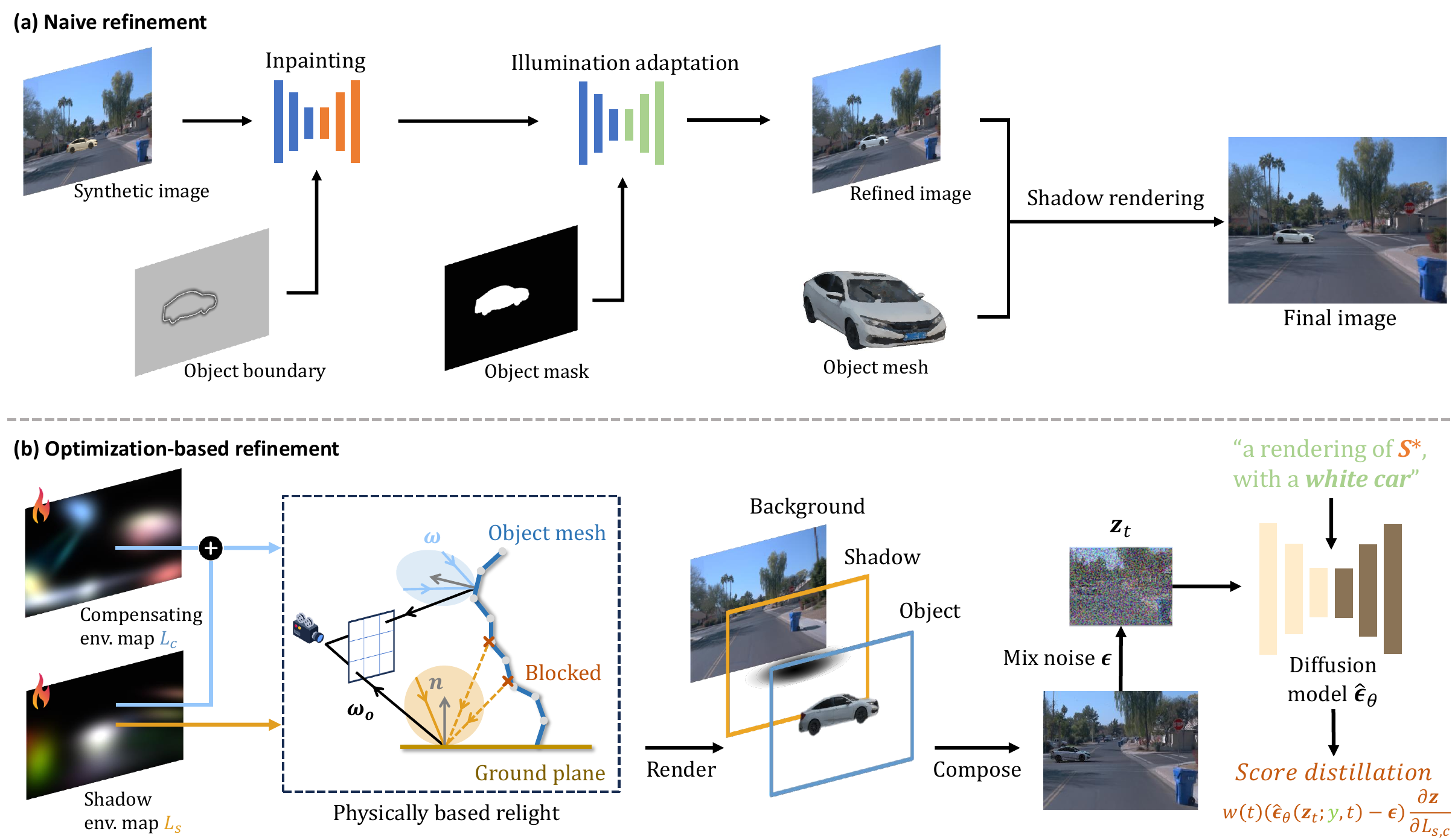}
 \cuthalfcaptionup
 \caption{\rebuttal{
We present two different approaches to refine synthetic data:
(a) A naive neural network-based method that leverages networks to inpaint object edges and adjust the lighting corresponding to the object mask.
(b) An optimization-based refinement approach that utilizes a differentiable PBR pipeline to apply SDS loss on the composed image, optimizing the background ambient lighting to achieve realistic relighting and shadow rendering.}}
 \label{fig:image_refine}
\end{figure*}

Due to the boundary's inconsistency and illumination difference, the inserted foreground objects might look conspicuous and weird in color. To make it more realistic, we follow GeoSim~\cite{geosim} to apply an inpainting-based method to first ``soften'' the boundary of the inserted object and then adapt lighting. We use Lama~\cite{lama} as our inpainting network architecture. First we extract the 
foreground boundary $\mathbf{M}_b$ from foreground object's mask $\hat{\mathbf{M}}$ and mask the inconsistent boundary in original synthetic image $\mathbf{I}$, getting a masked image $\mathbf{I}_m = \mathbf{I} \cdot \left( 1 - \mathbf{M}_b \right)$. Network takes boundary mask $\mathbf{M}_b$ and figure $\mathbf{I}_m$ to inpaint the masked regions in $\mathbf{I}_m$, providing a more natural synthetic image $\mathbf{I}^{\prime}$. For adapting illumination, network takes $\hat{\mathbf{M}}$ and figure ${\mathbf{I}}^\prime$ as inputs, and outputs the light-balanced figure ${\mathbf{I}}^{\prime\prime}$. Note that both of the two stages share the same inpainting network architecture with the only difference in the weight of parameters and the network input.

The inserted object looks floating on the ground as the lack of shadows. Note that we have the aligned object's mesh, so one can use conventional methods or existing applications (e.g. Blender) to render the corresponding shadows. To speed up the simulating, we set the environment light direction and project the mesh's vertices to the ground by light direction to derive a coarse mask of shadows $\mathbf{M}_s$. As the vertices of the mesh are discrete, we perform a morphological closing operation to fill the holes in $\mathbf{M}_s$. At last, a Gaussian blur is performed to make the shadow more realistic. The procedure of refining synthetic images is shown in Figure \ref{fig:image_refine}\textcolor{red}{(a)}.

\subsubsection{ Enhancement of reality by physically based rendering}
\rebuttal{
A single forward pass through a neural network for object insertion refinement offers computational efficiency and often delivers satisfactory results. However, such methods sometimes struggle to accurately model complex lighting conditions and realistic shadow interactions. Recent advancements in latent diffusion model generation techniques have showcased remarkable capabilities in understanding and reproducing scene lighting. Inspired by this progress, we incorporate an additional method leveraging physically based rendering (PBR) principles, wherein we optimize environment maps to achieve more harmonious relighting and realistic shadow rendering.
Following FEGR~\cite{Neuralfieldsmeetexplicit} and DiPIR~\cite{DiPIR}, we adopt the Disney lighting equation for relighting objects. The rendered object color is computed as:
\begin{align}
   \mathbf{I}_{fg}(\mathbf{x}) = 
   \int_{\boldsymbol{\Omega}} f_r(\mathbf{x}, \boldsymbol{\omega}, \boldsymbol{\omega}_{o})L_i(\boldsymbol{\omega})|\boldsymbol{\omega}\cdot \mathbf{n}|  d\boldsymbol{\omega}, 
\end{align}
where 
$\mathbf{x}$ represents the intersection of the camera ray $\boldsymbol{\omega}_o$ with the object surface, while $\mathbf{n}$ denotes the surface normal at $\mathbf{x}$. 
The observed color $\mathbf{I}_{fg}(\mathbf{x})$ is calculated by integrating incoming light $L_i$ in the direction $\omega$ over the hemisphere $\boldsymbol{\Omega}$, where the weight is determined by the bidirectional reflectance distribution function (BRDF) $f_r$ and the angle between the ray and the surface normal $|\boldsymbol{\omega}\cdot \mathbf{n}|$.
}

\rebuttal{
For shadow rendering, we assume the ground is a horizontal plane closely aligned with the object. For each ray emitted by the camera, its intersection point with the ground is first computed. Subsequently, the rays are uniformly sampled over the hemisphere at this intersection point, and their occlusion by the object mesh is determined. Assuming the ground is a perfectly Lambertian diffuse surface, the shadow intensity is calculated as the ratio of the flux of light rays not blocked by the object to the total light intensity over the hemisphere:
\begin{align}
   Intensity(\mathbf{x}) = 
   \frac{
   \int_{\boldsymbol{\Omega}} L_s(\boldsymbol{\omega})|\boldsymbol{\omega}\cdot \mathbf{n}| v(\boldsymbol{\omega}, \mathbf{x}) d\boldsymbol{\omega} }
   {
   \int_{\boldsymbol{\Omega}} L_s(\boldsymbol{\omega})|\boldsymbol{\omega}\cdot \mathbf{n}| d\boldsymbol{\omega} }, 
\end{align}
where $L_s$ represents the shadow environment map function, $\mathbf{x}$ denotes the intersection of the camera ray with the ground, $\mathbf{\omega}$ is the ray direction, and $v$ is a function that queries whether a ray directed toward $\mathbf{x}$ is blocked by the object mesh:
\begin{equation}
v\left(\mathbf{x}, \boldsymbol{\omega}\right)= \begin{cases}0 & \text { if } \boldsymbol{\omega} \text { is blocked by the object} \\ 1 & \text { otherwise.}\end{cases}
\end{equation}
Finally, the composed image $\mathbf{I}_c$ can be written as:
\begin{align}
    \mathbf{I}_c = \mathbf{I}_{bg}\cdot Intensity \cdot (1-\mathbf{M}_{fg})+\mathbf{I}_{fg}\cdot \mathbf{M}_{fg},
\end{align}
where $\mathbf{I}_{bg}$ represents the rendered background to be inserted, $\mathbf{I}_{fg}$ and $\mathbf{M}_{fg}$ denote the rendered RGB image and mask of the inserted object, respectively.
}

\rebuttal{
We apply SDS loss~(Eq.~\ref{eq:sds}) on $\mathbf{I}_{c}$ to optimize the environment maps $L_s$ and $L_i$. In our experiments, we observed that clear and specific prompts for background description can accelerate convergence and improve the effectiveness of the SDS loss. Therefore, we adopt the approach from textual inversion~\cite{gal2022image} and learn a special embedding $S*$ to serve as the textual prompt for the background image. This is then combined with a description of the inserted object to form the complete prompt.
}

\rebuttal{
Due to the high degree of freedom in high-resolution environment maps, which makes them difficult to optimize and susceptible to overfitting,  we use a similar approach in~\cite{DiPIR} and represent the lat long-type environment map as a mixture $n$ Gaussian models $\{\mathcal{G}_i\}_{i=1}^n$. Each Gaussian is defined by learnable parameters: position $\mathbf{p}_i \in \mathbb{R}^2$, color $\mathbf{c}_i\in \mathbb{R}^3$, weight $\alpha_i \in \mathbb{R}$  and scale $\mathbf{s}_i \in \mathbb{R}^2$. A environment map $L\in \mathbb{R}^3\to\mathbb{R}^3$ is expressed as:
\begin{align}
    L(\phi(\mathbf{x})) = \sum_i^n \mathbf{c}_i \alpha_i \exp \left({-\frac{1}{2}\|\frac{\mathbf{x}-\mathbf{p}_i}{\mathbf{s}_{i}}\|_2^2} \right),
\end{align}
where $\mathbf{x} \in \mathbb{R}^2$ represents 2D latitude-longitude coordinates and $\phi \in \mathbb{R}^2\to\mathbb{R}^3$ is a function that transforms these coordinates into the directions.
For simplicity, we set the reconstructed or generated texture of the mesh as the Albedo map. Consequently, the reconstructed mesh inherently incorporates information from the dataset environment map $L_d$, which is unknown. Therefore, we do not directly optimize $L_i$. Instead, we define $L_i=L_s+L_c$, where $L_c$ is a compensating environment map optimized alongside $L_s$. We aim for $L_c+L_d$ to approximate the observed color proportional to the true albedo, while $L_s$ is optimized to simulate the lighting conditions of the inserted background for relighting and shadowing. Consequently, the final learnable parameters are $\{\mathcal{G}_i^{s}\}_{i=1}^n\cup \{\mathcal{G}_i^{c}\}_{i=1}^n$. The optimization pipeline is shown in Figure \ref{fig:image_refine}\textcolor{red}{(b)}.
}
\section{Experiments}
In this section, we will provide a detailed explanation of how neural reconstruction is applied to existing autonomous driving datasets, including specific training details.  We then introduce our foreground asset library and describe the simulation process on our autonomous driving platform, concluding with the validation of our simulation data through training on downstream tasks.

\subsection{Implementation details}
\subsubsection{Dataset}
We reconstruct the urban scenes on two datasets, \textbf{nuScenes}~\cite{nuscenes2019} and \textbf{Waymo Open Dataset}~\cite{Sun_2020_CVPR}.
\textbf{nuScenes} dataset contains 1000 scenes collected from different city blocks. Each scene consists of around 1000 images in six views that cover the 360$^\circ$ field of view (captured by the front, right-front, right-back, back, left-back, and left-front cameras). 32-channel LiDAR data are collected at 20 Hz. Human annotations are given in key LiDAR frames (one frame is labeled in every 10 frames). \textbf{Waymo Open Dataset} includes data from 1,000 driving segments. Each driving segment contains 20 seconds of continuous driving. Each segment has the data captured by five high-resolution 64-beam lidars and five front-and-side-facing cameras. All lidar frames and images are with vehicles, pedestrians, cyclists, and signage carefully labeled.
\subsubsection{Street scenes reconstruction settings}
 Due to the limitation of storage and NeRF's expression ability, we divided each scene sequence into several scene fragments for reconstruction. More specifically, each scene segment contains about 50 successive video frames, each of which has its corresponding 360-degree LiDAR data, multi-view camera data, and foreground object labels. We select 100 scene segments each on Waymo and 60 segments on nuScenes to train NeRFs for the background reconstruction.
 
\textcolor{black}{We set the $\lambda_{sem} = 4e-3, \lambda_{d}=0.3, \lambda_L=0.3, \lambda_{smo}=1e-3, \lambda_{dt}=0.01, \lambda_{prop}=1$. We use a two-stage proposal network, sampling 64 points in the proposal stage and 32 points in the final stage for both Waymo and nuScenes training experiments.}

\subsubsection{Foreground bank settings}
Some of our foreground data is sourced from vehicles reconstructed in nuScenes. This collection is significantly expanded by incorporating foreground objects reconstructed from our collected 360-degree surround, object-centered dataset. Additionally, it is further enriched with assets created using our proposed generation method.

\paragraph{Foreground objects reconstructed from collected datasets}
We collected more than 50 360-degree object videos which contain cars, bicycles, motorcycles, and traffic cones. Most data are collected from the Co3D dataset~\cite{CO3D}, and some are self-shot videos. Then we run COLMAP~\cite{sfm} for sparse point cloud reconstruction and abnormal failure cases. To model the foreground objects separated from the background, we apply a segmentation method to get the foreground mask.

We reconstruct a human avatar from the People-snap-shot dataset. For simplicity, we use the official checkpoint from SelfRcon~\cite{selfrecon}. We then give the meshes T-pose bone binding on the MAXIMO website and save a series of poses to a series of transformed meshes to render different poses of the human avatar.

\paragraph{Generative foreground objects}
We take objects mesh from diverse Internet resources, and most of them lack satisfying textures. Our generative foreground bank includes bicycles, motorcycles, cars, buses, and human avatars.

For each object category, we collect thousands of images of the corresponding category and use a control-net to fine-tune the StableDiffusion~\cite{SD} to get the category-specific text-condition diffusion model. We use TEXTure~\cite{Texture} to get the initial texture of a mesh. Then we refine the texture by SDS in latent space. We use a normal form `a \{object\} in \{color\}' as text insertion of the diffusion model.

For human avatar texturing, we directly use the checkpoint of StableDiffusion to apply SDS to train the texture in latent space. Then we decode the latent texture map by the VAE decoder of the LDM to get a high-resolution RGB texture.

\setlength{\tabcolsep}{10pt}
\renewcommand\arraystretch{1.3}
\begin{table}[t]
\setlength{\abovecaptionskip}{2pt}
\setlength{\belowcaptionskip}{4pt}
\centering
\caption{\color{black} Our method quantitatively outperforms state-of-the-art methods \cite{emernerf, streetsurf}. Experiments are conducted on two static scenes in Waymo. Average PSNR, SSIM and LPIPS are reported.}

\begin{tabular}{lccc}
\midrule
\multicolumn{4}{c}{\textbf{Large-scale static scenes synthesis on Waymo}} \\
Methods & PSNR$\uparrow$ & SSIM$\uparrow$ & LPIPS$\downarrow$ \\

\hline
EmerNeRF ~\cite{emernerf}& 24.14 & 0.787 & 0.415 \\
StreetSurf ~\cite{streetsurf} & 23.08  & 0.806 & 0.396 \\
Zip-NeRF~\cite{zipnerf}  & 25.05& 0.799 & 0.230\\
S-NeRF~\cite{snerf} & 23.60 & 0.743 & 0.422 \\
Our S-NeRF++ & \textbf{25.78} & \textbf{0.810} & \textbf{0.224}\\
\midrule
\end{tabular}
\label{tab:waymo_background_eval}

\end{table}
\begin{table}[th]
\setlength{\abovecaptionskip}{0pt}
\setlength{\belowcaptionskip}{-4pt}
\centering
\caption{\color{black} Our method quantitatively outperforms state-of-the-art methods. Methods are tested on four nuScenes Sequences with dynamic vehicles. Average PSNR and SSIM about full image and dynamic vehicles only are reported.}  
\begin{tabular}{p{1.9cm}p{0.4cm}p{0.6cm}|c}
\midrule
\multicolumn{4}{c}{\textbf{Large-scale dynamic scenes synthesis on NuScenes}} \\
 & \multicolumn{2}{c}{Full Image}& Dynamic Veh.\\
  Methods& PSNR$\uparrow$ & SSIM$\uparrow$ & PSNR$\uparrow$\\

\hline
EmerNeRF ~\cite{emernerf}& 26.23  & 0.734  & 21.91\\
Our S-NeRF++& 
\textbf{27.10} & \textbf{0.827 }& \textbf{24.97}\\

\midrule
\end{tabular}

\label{tab:scenes}

\end{table}

\subsubsection{Figure refinement module implementation}
In this section, we further explain the details of our figure refinement network. We modified the network architecture of LaMa~\cite{lama} and finetuned it using the training datasets from Waymo and nuScenes. To construct the inpainting training dataset. We first generate an initial boundary mask based on the mask of moving vehicles. Then for each training step, we randomly expand the boundary for better overall performance. We then transfer the inpainting dataset to our relighting dataset by using different color-jitter to the mask of the vehicle and rest background. This adaptation is designed to enable our relighting network to learn and adapt to changes in light conditions, ensuring that the recolored vehicles harmonize with the background scene.

\begin{figure}[ht!]
\centering
\hspace{-3mm}
\subfigure{
\begin{minipage}[t]{0.315\linewidth}
\centering
\includegraphics[width=1\linewidth]{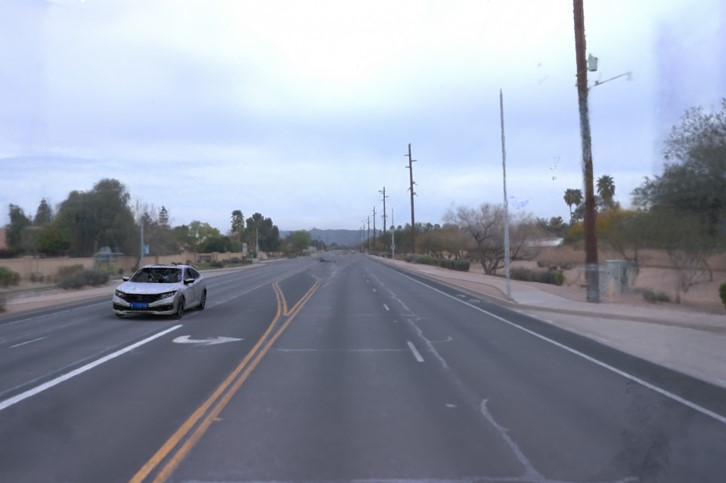}
\end{minipage}%
}%
\hspace{-2mm}
\subfigure{
\begin{minipage}[t]{0.315\linewidth}
\centering
\includegraphics[width=1\linewidth]{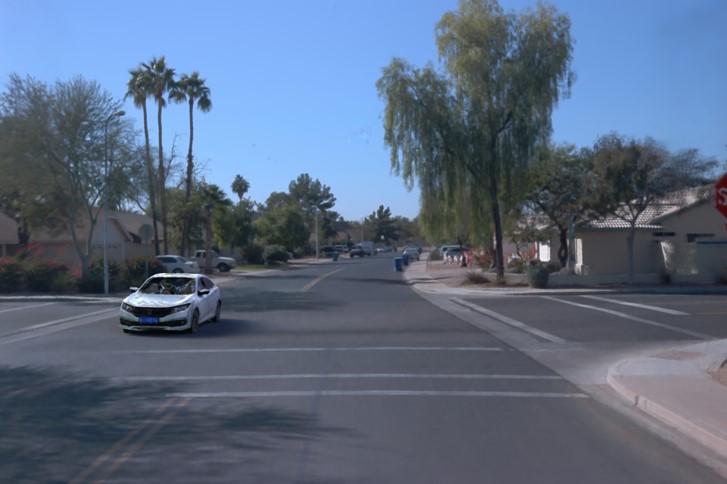}
\end{minipage}%
}%
\hspace{-2mm}
\subfigure{
\begin{minipage}[t]{0.315\linewidth}
\centering
\includegraphics[width=1\linewidth]{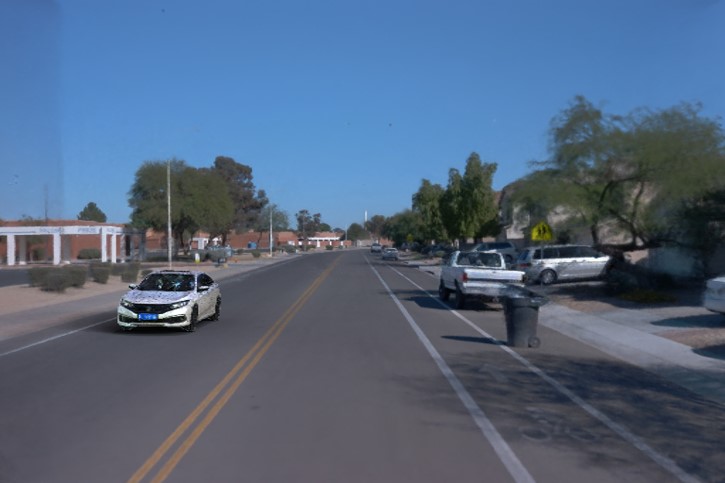}
\end{minipage}%
}%
\caption{\rebuttal{
Our optimization-based refinement can restore accurate lighting conditions, thereby enhancing the realism of the data.}}
\label{fig:pbr}
\end{figure}

\rebuttal{
For the implementation of our optimization-based refinement, we utilize StableDiffusion~\cite{SD} v2.1 within the SDS loss. The number of Gaussian primitives is set to $n=50$ for both $L_c$ and $L_s$, and a cosine learning rate schedule with a peak value of $5e-3$ is applied to all parameters. Training is conducted for 2000 steps using the Adam optimizer. The classifier-free guidance (CFG) is set to 5, with 1024 rays uniformly sampled from the hemisphere for each pixel. Additionally, the noise level $t$ in the SDS loss is uniformly sampled from [0,0.3]. For the foreground material, we adopt the simplest assumption, setting $k_d=1$ and $k_s=0$. 
As shown in Figure~\ref{fig:pbr}, our method is capable of learning the correct ambient lighting.
}

\subsection{Experiment results}

\subsubsection{Renderings of street view in nuScenes and Waymo}
Here we demonstrate the performance of our method by comparing it with the state-of-the-art street view reconstruction method StreetSurf~\cite{streetsurf} and EmerNeRF~\cite{emernerf}.
\begin{figure*}[h]
\centering
 \includegraphics[width=0.96\linewidth]{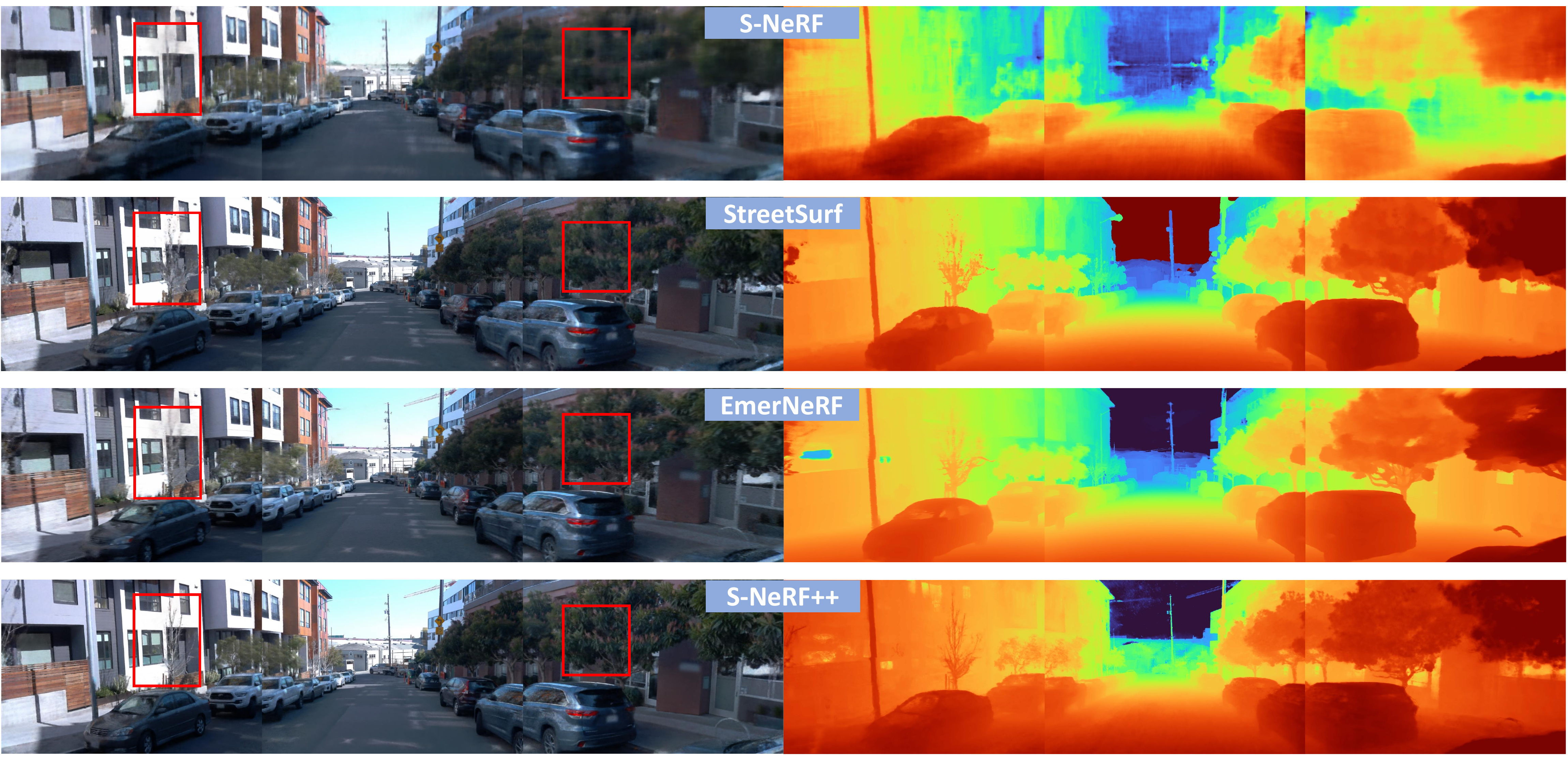}
 \cuthalfcaptionup
 \caption{\color{black} 
Qualitative renderings on Waymo demonstrate that our S-NeRF++ method outperforms others in capturing image details.}
 \label{fig:waymo_seq}
\end{figure*}
\begin{figure*}[h]
\centering
 \includegraphics[width=0.95\linewidth]{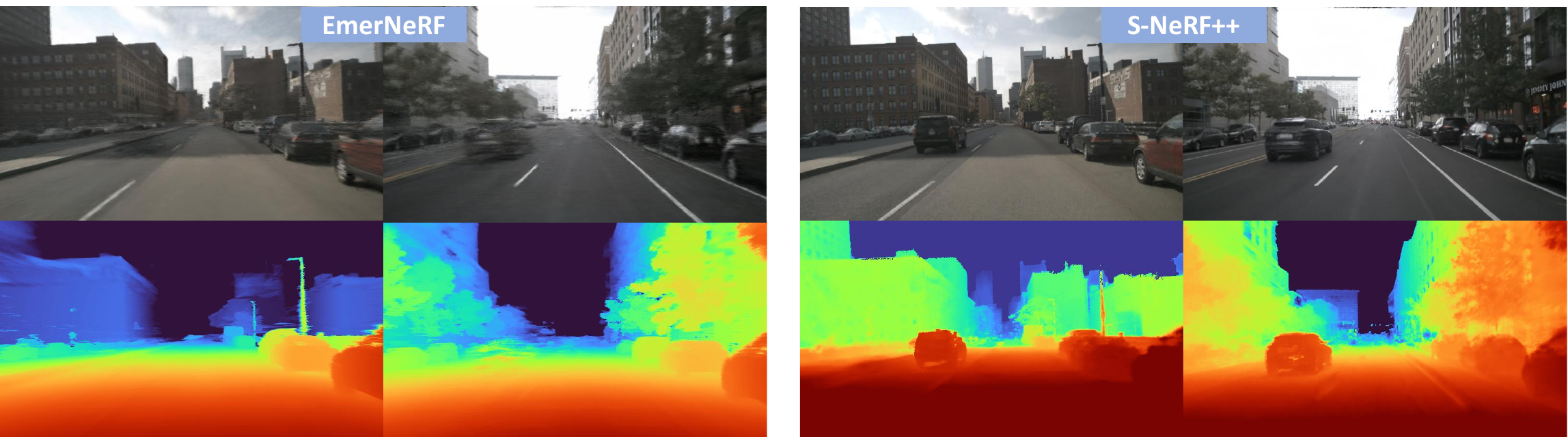}
 \cuthalfcaptionup
 \caption{\color{black} Renderings on nuScenes. Our method delivers superior quality, whereas EmerNeRF often fails to reconstruct high-speed vehicles for the sparsity of LiDAR supervision.}
 \label{fig:nus_mv_seq}
\end{figure*}

\textcolor{black}{We test three nuScenes sequences with the dynamic vehicle and two Waymo sequences with only static background. nuScenes visualizations are in Figure \ref{fig:nus_mv_seq}, and Waymo sequences renderings are shown in Figure \ref{fig:waymo_seq}. We report the evaluation results in Table \ref{tab:scenes} and \ref{tab:waymo_background_eval}.  Extensive experiments show that our methods surpass StreetSurf~\cite{streetsurf} and EmerNeRF~\cite{emernerf} across all three evaluation metrics in terms of static scenes. We observe significant improvements including a 1.64 increase in PSNR, a 2.9\% increase in SSIM, and a 0.191 reduction in LPIPS compared with the EmerNeRF~\cite{emernerf}. Additionally, our method demonstrates strong performance in handling moving vehicles within the scene, achieving superior results in nuScenes dynamic sequences. Our method surpasses Emernerf with a 0.87 increase in PSNR, a 0.93 increase in SSIM in full image metrics, and a 3.06 increase in PSNR for dynamic vehicles.}

\begin{figure}[ht!]
\centering
\hspace{-2.5mm}
\subfigure{
\begin{minipage}[t]{0.32\linewidth}
\centering
\includegraphics[width=1\linewidth]{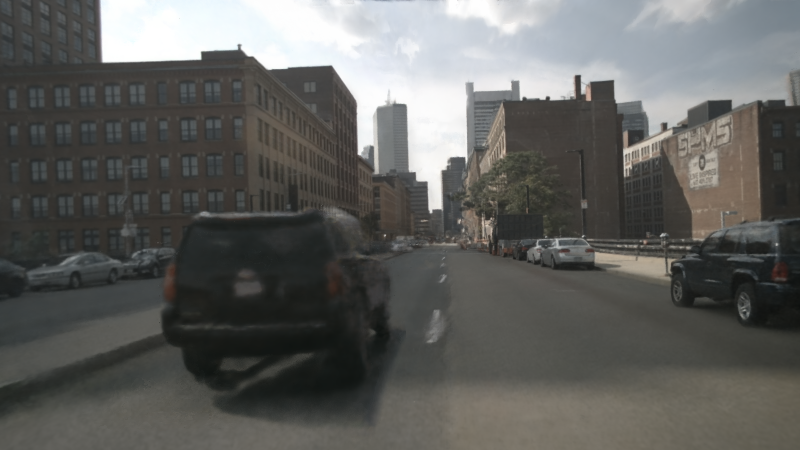}
\end{minipage}%
}%
\hspace{-2mm}
\subfigure{
\begin{minipage}[t]{0.32\linewidth}
\centering
\includegraphics[width=1\linewidth]{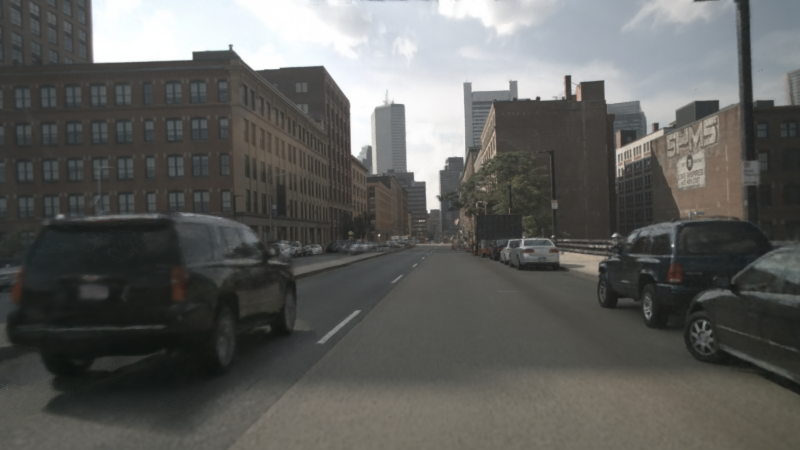}
\end{minipage}%
}%
\hspace{-2mm}
\subfigure{
\begin{minipage}[t]{0.32\linewidth}
\centering
\includegraphics[width=1\linewidth]{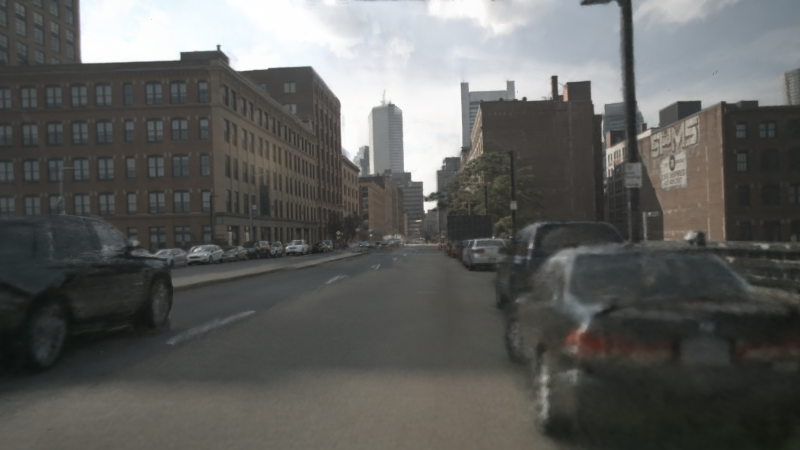}
\end{minipage}%
}%
\caption{\color{black} Free view renderings with shifting camera translation.}
\label{fig:free_view}
\end{figure}

\begin{figure}[ht!]
\centering
\subfigure{
\centering
\includegraphics[width=0.48\linewidth]{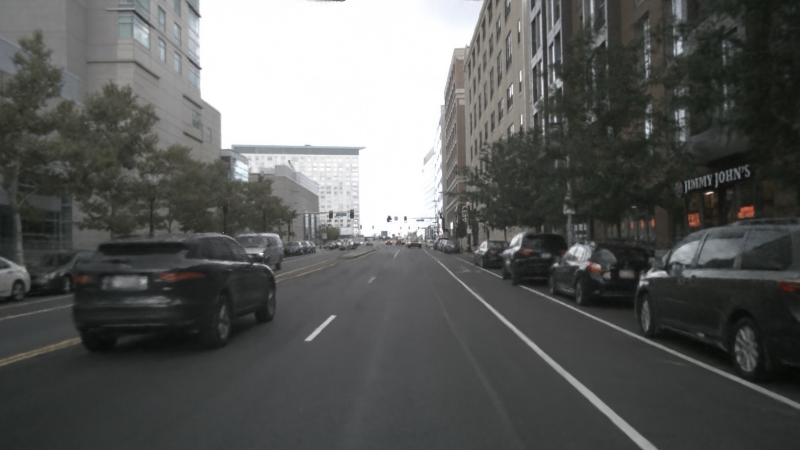}
}%
\hspace{-3mm}
\subfigure{
\centering
\includegraphics[width=0.48\linewidth]{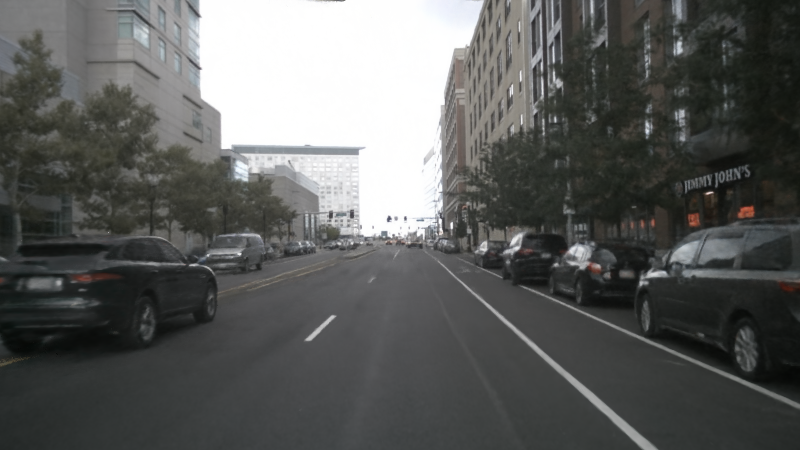}
}
\caption{\color{black} Manipulation of moving vehicles by changing the positions of corresponding bounding boxes.}
\label{fig:new_trace}
\end{figure}

\textcolor{black}{In Figure \ref{fig:free_view} and \ref{fig:new_trace}, we also demonstrate the robustness of our method for synthesizing out-of-distribution viewpoints and its capability in handling moving vehicles.
}

\begin{figure}[htbp]
\subfigure[w/ pose refinement]{
\centering
\includegraphics[width=0.48\linewidth]{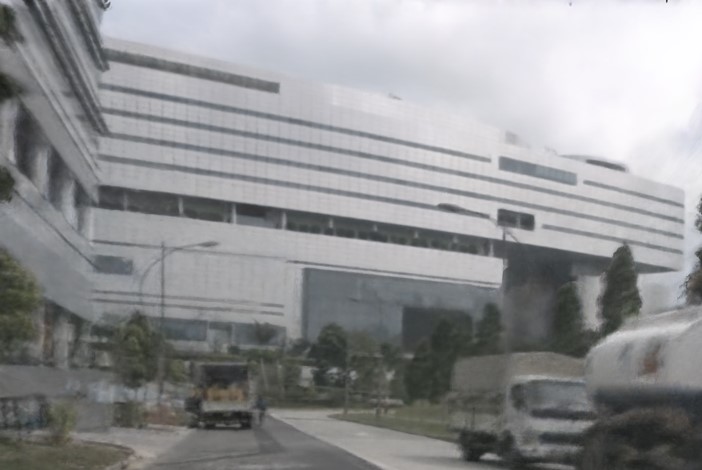}
}%
\hspace{-3mm}
\subfigure[w/o pose refinement]{
\centering
\includegraphics[width=0.48\linewidth]{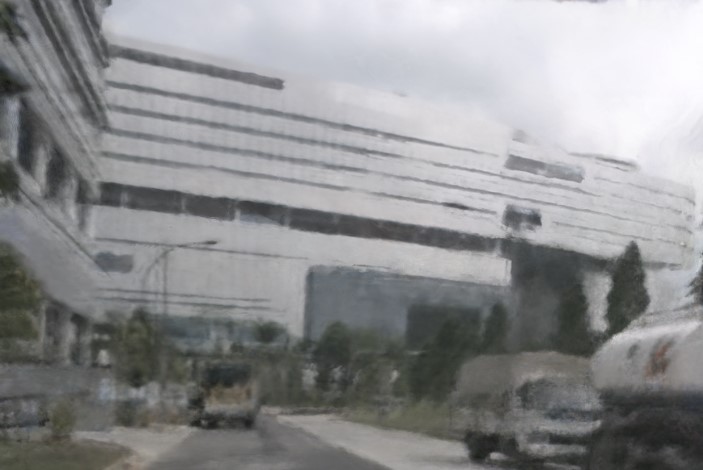}
}
\\
\subfigure[w/ track refinement]{
\centering
\includegraphics[width=0.48\linewidth]{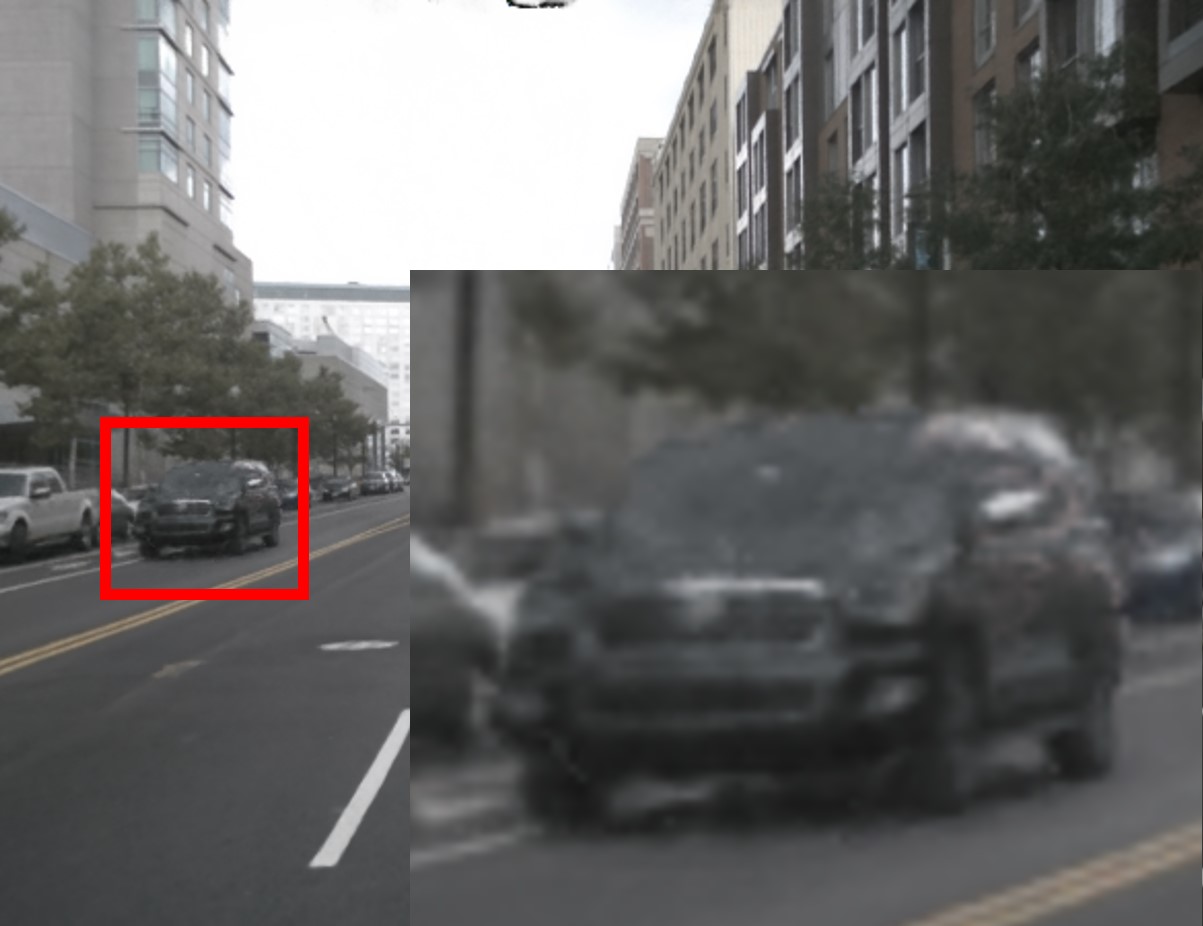}
}%
\hspace{-3mm}
\subfigure[w/o track refinement]{
\centering
\includegraphics[width=0.48\linewidth]{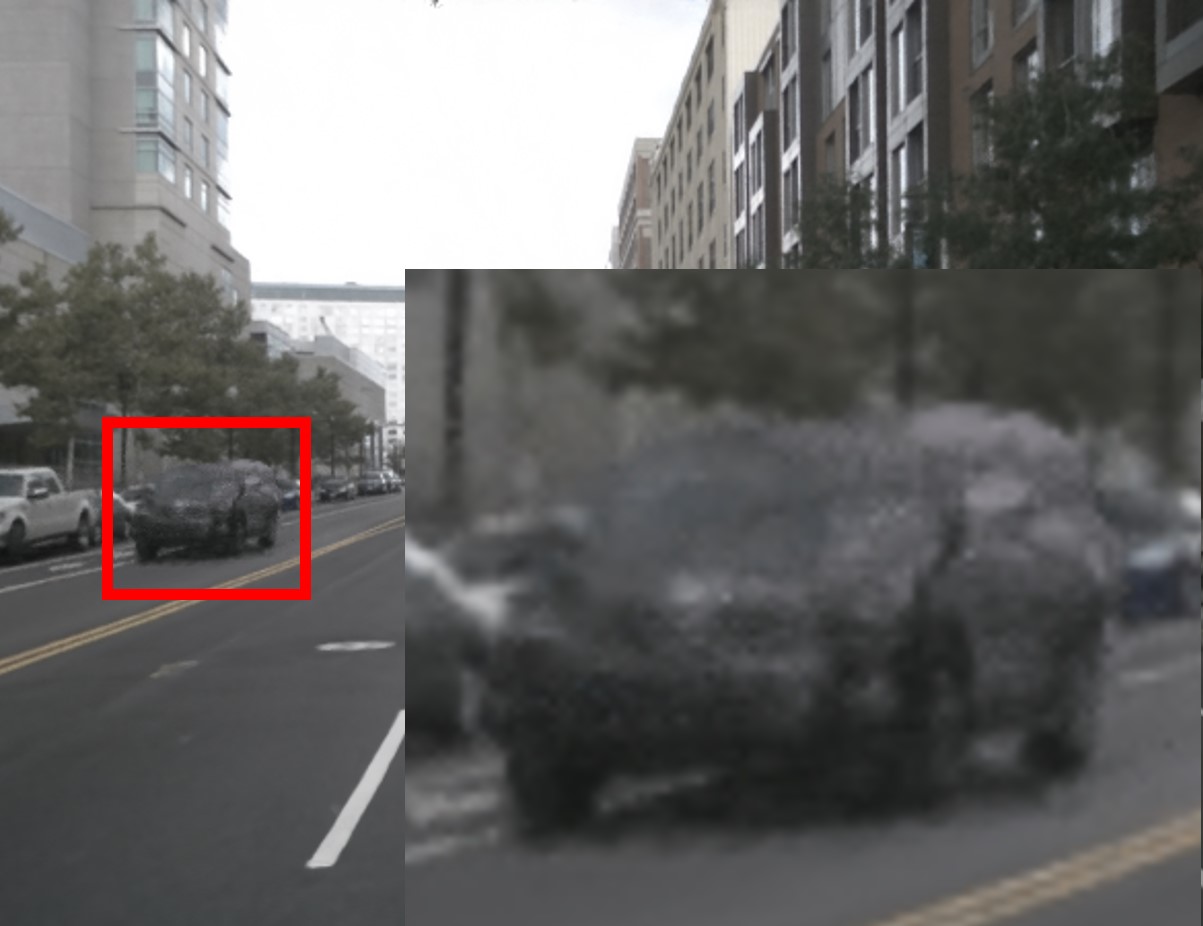}
}
\caption{\rebuttal Refining camera poses and object tracks can alleviate misalignment issues, resulting in clearer rendered images.}
\label{fig:pose refine}
\end{figure}

\begin{table}[t]
\centering
\caption{Monocular 3D vehicle detection evaluation on Waymo}

\begin{tabular}{p{3.5cm}<{}c}
\midrule

\multicolumn{2}{c}{\textbf{PGD~\cite{pgd}  training results on Waymo}} \\
 Data   & Car 3D $\text{AP}_{50}$\\
\midrule
Real 1k & 5.0 \\ 
Real 10k& 11.6 \\
Sim & 6.3\\
\midrule
Sim + Real 1k& \textbf{12.3} \\
\midrule

\end{tabular}
\label{tab:mono 3d detection}

\end{table}

\begin{table}[t]
\centering
\caption{Multi-view 3D vehicle detection evaluation on nuScenes.}

\begin{tabular}{p{3.5cm}<{} c}
\midrule

\multicolumn{2}{c}{\textbf{DETR3D\cite{detr3d} training results on nuScenes}} \\
 Data   & Car 3D $\text{AP}_\text{nus}$\\
\midrule
Real 500 &  21.2\\ 
Real 5k& 43.5 \\
Sim & 33.9\\
\midrule
Sim + Real 500& \textbf{45.7} \\
\midrule

\end{tabular}
\label{tab:mv 3d detection}

\end{table}

\begin{table}[t]
\centering
\caption{19 class semantic segmentation evaluation on Waymo. }
\begin{tabular}{lcc}
\midrule
\multicolumn{3}{c}{\textbf{Semantic Segmentation results on Waymo (mAP)}} \\
 Data   & Segformer~\cite{segformer}  & DeeplabV3~\cite{deeplabv3} \\
\midrule

Real 1.3k &   68.8   &  64.1\\ 
Real 13k& 74.3 & 67.7  \\
Sim & 70.8 & 65.4\\
\midrule

Sim + Real 1.3k& \textbf{76.0} & \textbf{68.5}\\

\midrule
\end{tabular}
\label{tab:image segmentation}

\end{table}

\begin{table}[th]
\centering
\caption{\color{black} Ablation on pose/track refinement influencing the full image and dynamic vehicles.} 

\begin{tabular}{ccc|c}
\midrule
 & \multicolumn{2}{c}{Full image}& Dynamic veh.\\
  Refinement& PSNR$\uparrow$ & SSIM$\uparrow$ & PSNR$\uparrow$\\

\hline
w/& \textbf{29.09 }& \textbf{ 0.860} & \textbf{24.97}\\
w/o& 28.21 & 0.834 & 24.56\\
\midrule
\end{tabular}
\label{tab:poserefine}
\end{table}

\begin{table}[h] 
\caption{\textcolor{black}{Ablations on object insertion refinement module on simulation quality and efficiency}} 
\centering
\begin{tabular}{c|ccc}
\midrule
    &w/o Refinement & Naive & Optimization-based\\ 
\midrule
FID$\downarrow$& 73.8 & 67.1 & \textbf{62.3}  \\ 
Cost time & -- & $\leq$1s & $\approx$20min \\
\midrule
\end{tabular}

\label{tab:fid refinement}
 
\end{table}

\begin{table*}[h]
\centering

\caption{Ablation study on confidence components.}
\begin{tabular}{ccc|cc|cc} 
\midrule

\multicolumn{5}{c|}{Components of confidence} &\multicolumn{2}{c}{\multirow{2}{*}{Metrics}} \\ 
\cline{1-5}
\multicolumn{3}{c|}{Rerojection confidence}& \multicolumn{2}{c|}{Geometry confidence}  &  &                    \\ 
\midrule
rgb                     & ssim      & vgg   & depth         & flow & PSNR           & Depth error rate    \\ 
\midrule
                        &           &       &               &      & 23.15          & 30.70\%             \\ 
            $\checkmark$&           &       &               &      & 23.83          & 26.85\%  \\ 
                        &$\checkmark$ &     &               &      & 23.97          & 26.73\%             \\ 
                        &           &$\checkmark$ &         &      & 23.94          & 26.25\%             \\ 
                        &           &       & $\checkmark$  &       & 23.95         & 25.50\%             \\ 
                        &           &       &               &$\checkmark$& 24.03    & 24.92\%             \\ 
                        &           &       &$\checkmark$   &$\checkmark$       & 23.95         & 24.59\%             \\ 
$\checkmark$            &$\checkmark$&$\checkmark$&         &      & 23.88          & 26.75\%             \\ 
$\checkmark$            &$\checkmark$&$\checkmark$&$\checkmark$&$\checkmark$& \textbf{24.05} & \textbf{24.43\%}    \\
\midrule

\end{tabular}
\label{tab:err_rate}
\end{table*}

\begin{table}[h] 
\caption{Ablations on confidence settings for novel street-view rendering (background).} 
\centering

\begin{tabular}{cc|p{0.6cm}<{\centering}p{0.6cm}<{\centering}p{0.6cm}<{\centering}}
\midrule

  Reporjection & Geometry &  PSNR$\uparrow$ & SSIM$\uparrow$ & LPIPS$\downarrow$    \\ 
\midrule
Yes & No &23.88 & 0.767 &0.385\\
No &$\tau=20\%$& 23.95 &0.770 & 0.385\\
Yes & $\tau=10\%$&23.93 &\textbf{0.771} & 0.385\\
Yes &$\tau=40\%$&23.97 &\textbf{0.771} & \textbf{0.384}\\
\midrule
Yes & $\tau=20\%$ & \textbf{24.05} & \textbf{0.771} & \textbf{0.384} \\
\midrule

\end{tabular}
\label{tab:ablation_conf_background}

\end{table}

\begin{table}[h]
\caption{\color{black} The influence of foreground bank constructed by reconstruction and generative method on downstream tasks.} 
\centering
\begin{tabular}{>{\centering\arraybackslash}p{1.47cm}|>{\centering\arraybackslash}p{1.47cm} >{\centering\arraybackslash}p{1.5cm}|>{\centering\arraybackslash}p{0.7cm}}
\midrule

  \multirow{2}{*}{Task} & \multicolumn{2}{c|}{Foreground Source} & \multirow{2}{*}{Metrics} \\
  \cline{2-3}
  &Generation& Reconstruction & \\
  \midrule
  \multirow{4}{*}{Segmentation} &   &  &56.8\\
   &   & $\checkmark$ &58.2\\
    & $\checkmark$  &  &61.9\\
     & $\checkmark$  & $\checkmark$ &\textbf{62.8}\\
      \midrule
  \multirow{4}{*}{MV-Detection} &   &  &26.5\\
   &   & $\checkmark$ &32.3\\
    & $\checkmark$  &  &28.7\\
     & $\checkmark$  & $\checkmark$ &\textbf{33.9}\\
 \midrule  
\end{tabular}
\label{tab: foreground insertion}

\end{table}

\begin{table}[h]
\centering
\caption{\color{black} Ablation on different simulation strategy influencing the Multi-view 3D vehicle detection performance. The metrics is 3D AP of car. We use the simulation data to pretrain the model and finetuned with 500 real data. }
\begin{tabular}{p{3.5cm}<{} c c}
\midrule
Data  & Pretrain& Finetuned\\
\midrule
Original views &  22.9 & 34.3\\ 
Interpolation & 23.8  & 34.6 \\
Extrapolation w/o insertion & 26.5 & 36.7\\
Extrapolation w/ insertion & \textbf{33.9} & \textbf{45.9} \\
5k real& - &  43.5 \\
\midrule

\end{tabular}
\label{tab:sampling strategy}

\end{table}

\begin{table}[h]
\centering
\caption{\color{black} Ablation on different training strategy influencing the Multi-view 3D vehicle detection performance. }
\begin{tabular}{p{5cm}<{} c}
\midrule
Training strategy  & Car $\text{AP}_\text{nus}$\\
\midrule
Only 10k sim &  33.9\\ 
Only 500 real  & 21.2\\
10k Sim pretraining, 500 Real finetuning & \textbf{45.7} \\
500 real, and 10k sim joint training & 39.7\\
\midrule

\end{tabular}
\label{tab:training strategy}

\end{table}

\begin{figure*}[th!] \centering
\setlength{\abovecaptionskip}{4pt}
\setlength{\belowcaptionskip}{-8pt}
\small
\setlength{\tabcolsep}{1pt}
\begin{tabular}{cccccc}
     BACK-RIGHT & BACK & BACK-LEFT & FRONT-LEFT&  FRONT& FRONT-RIGHT \\
     \includegraphics[width=0.16\textwidth]{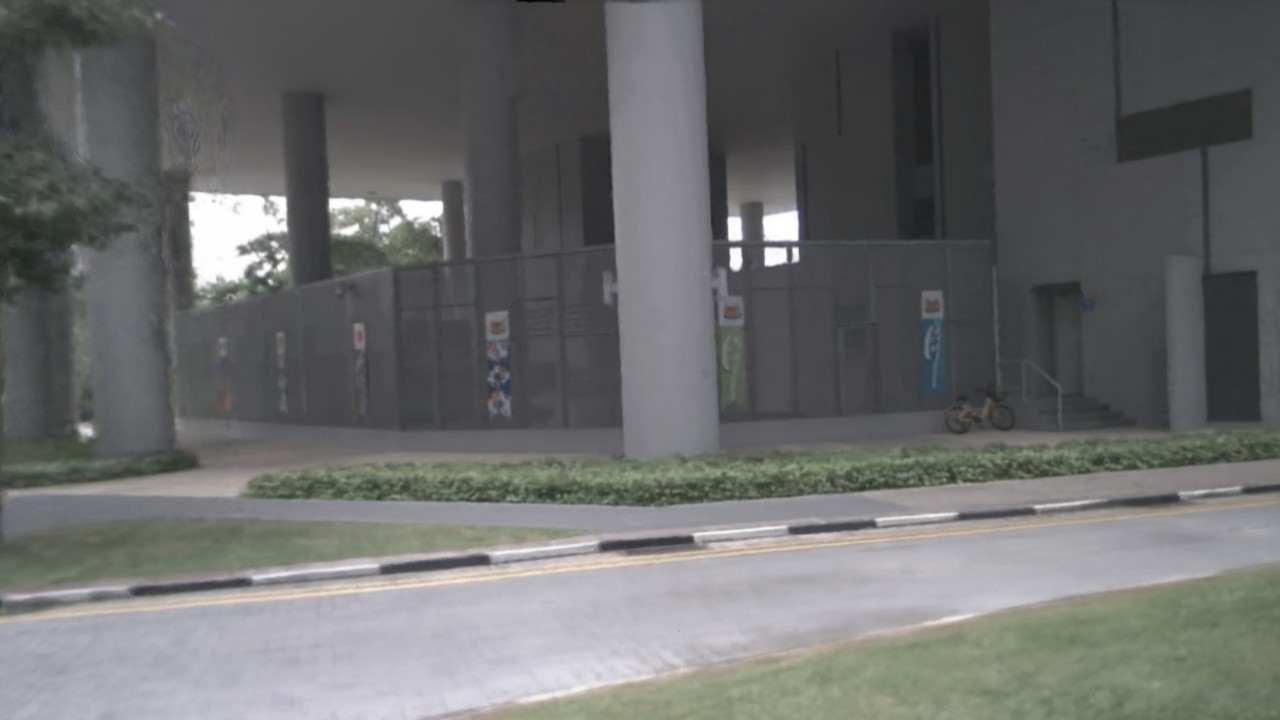} & 
     \includegraphics[width=0.16\textwidth]{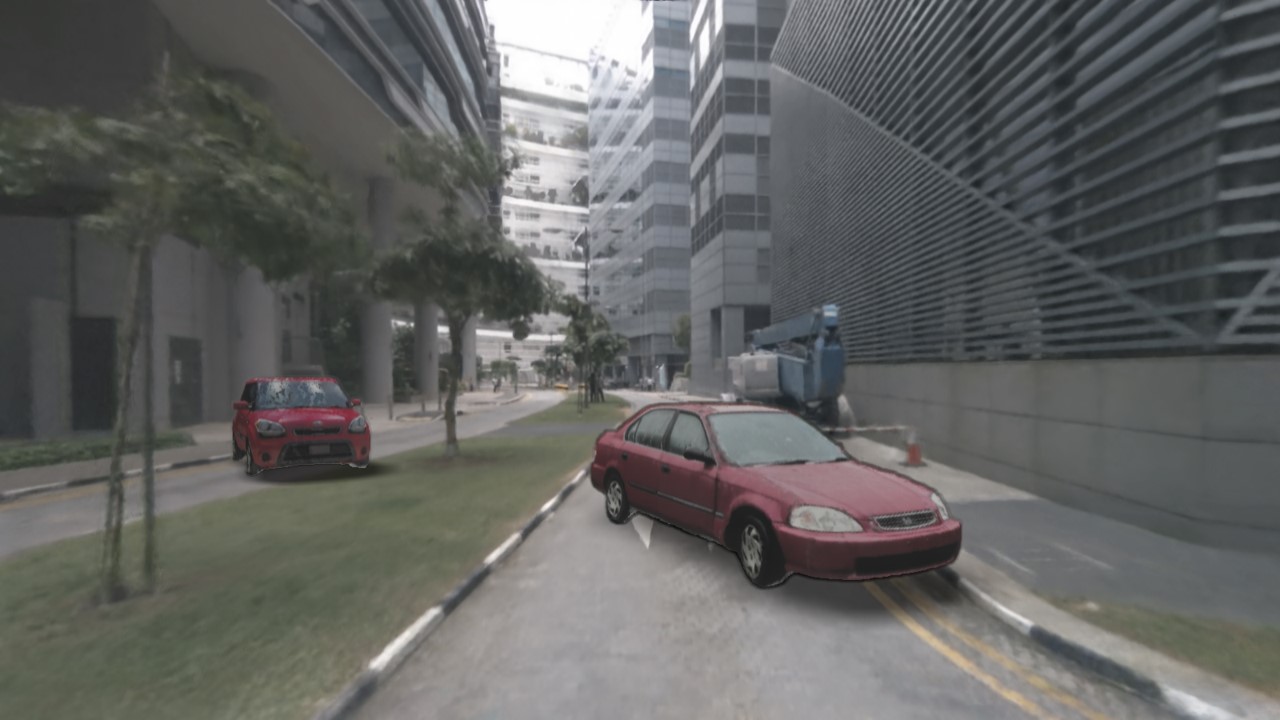} &
     \includegraphics[width=0.16\textwidth]{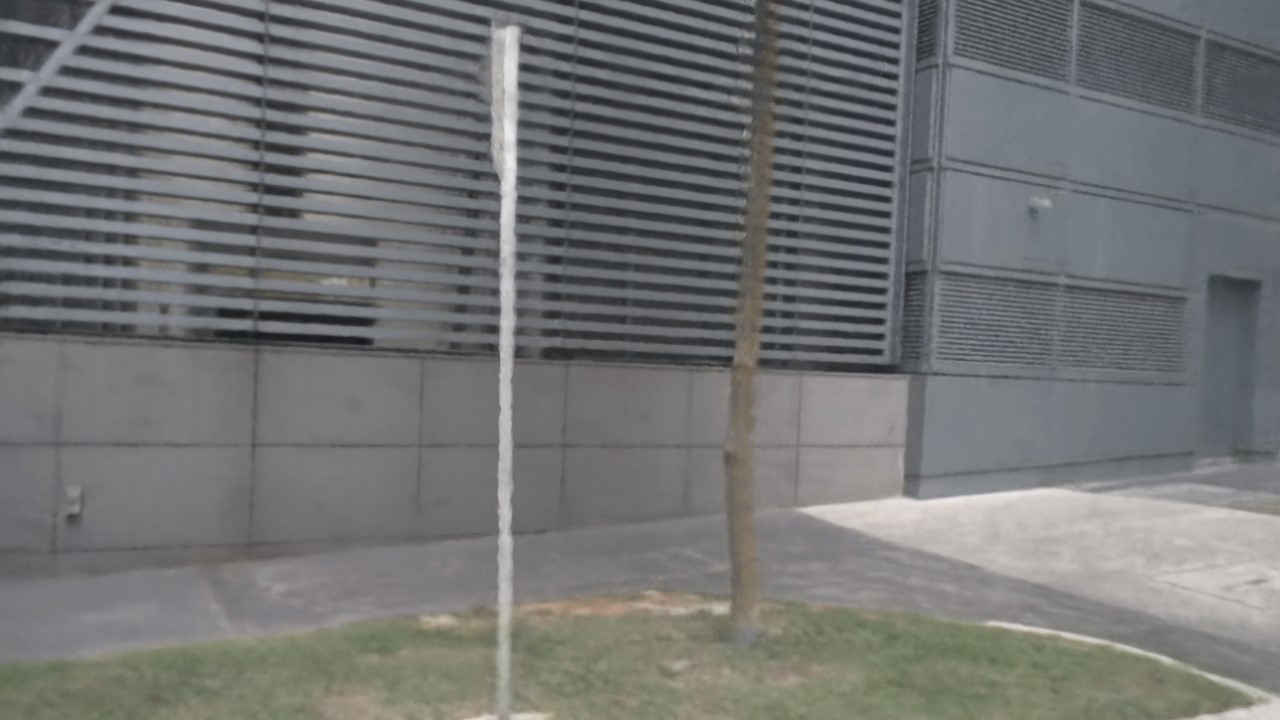}& 
     \includegraphics[width=0.16\textwidth]{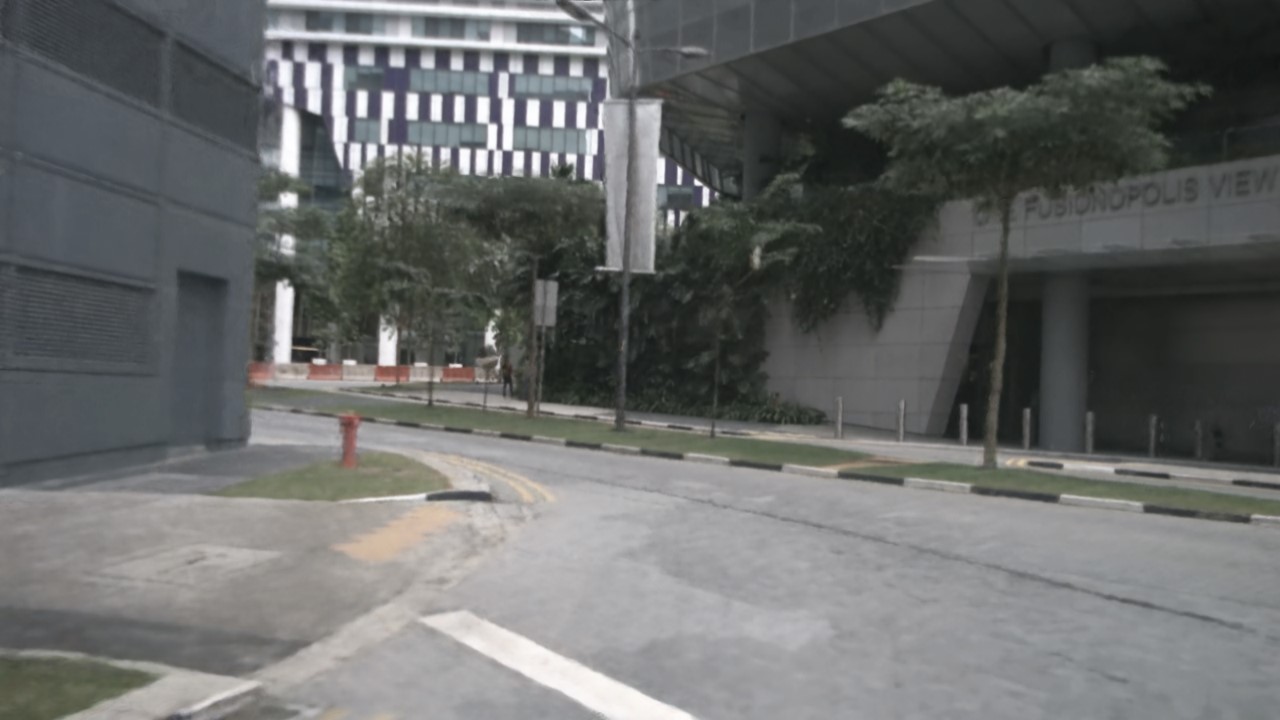} &
     \includegraphics[width=0.16\textwidth]{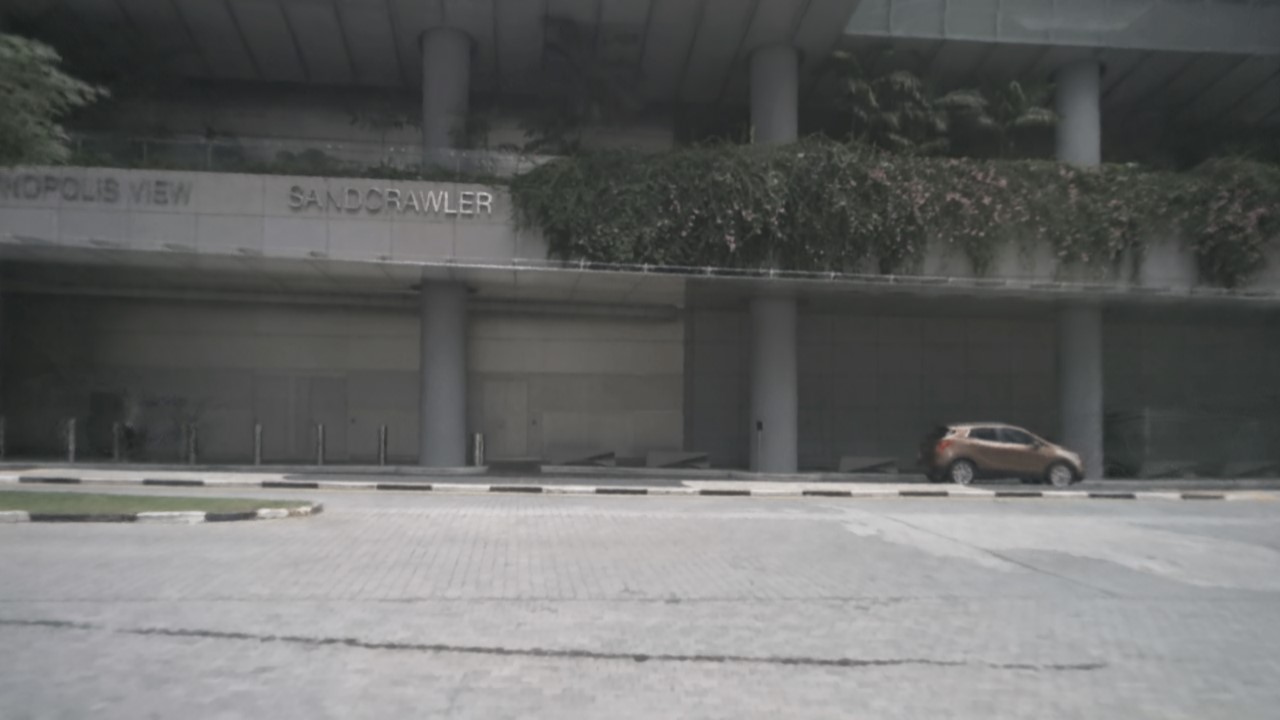}&
     \includegraphics[width=0.16\textwidth]{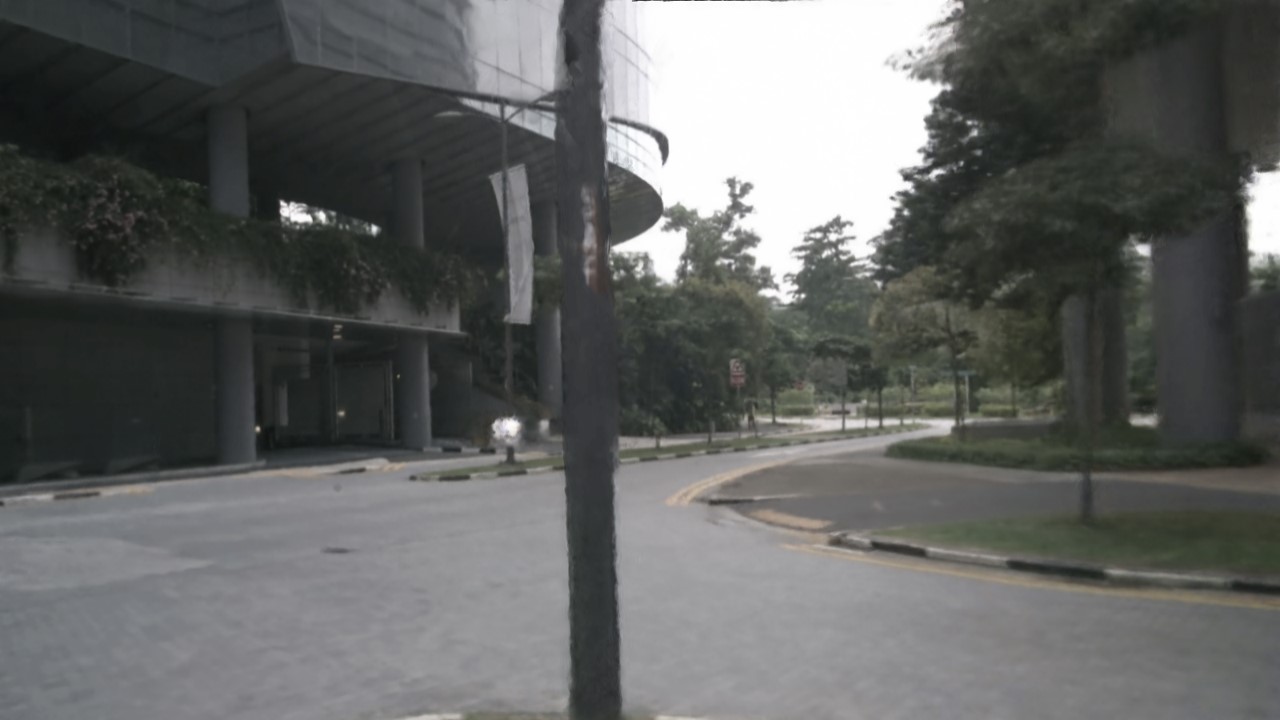}
     \\
    \includegraphics[width=0.16\textwidth]{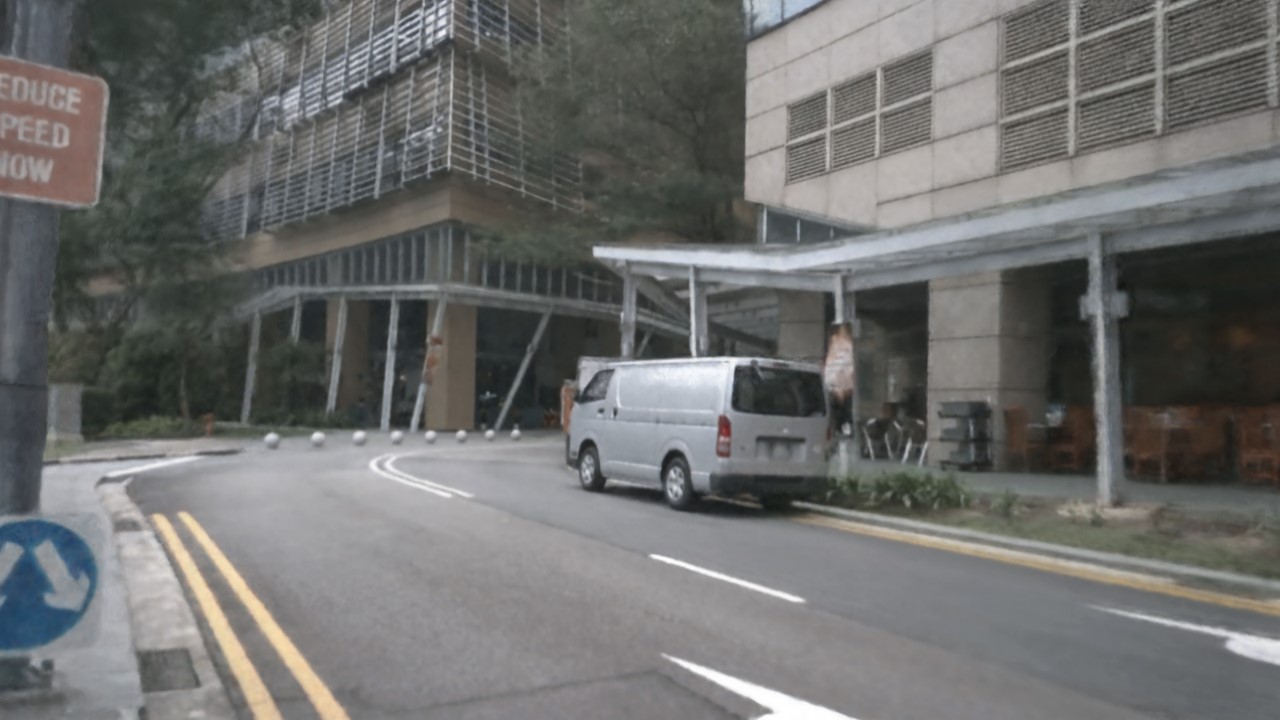} & 
     \includegraphics[width=0.16\textwidth]{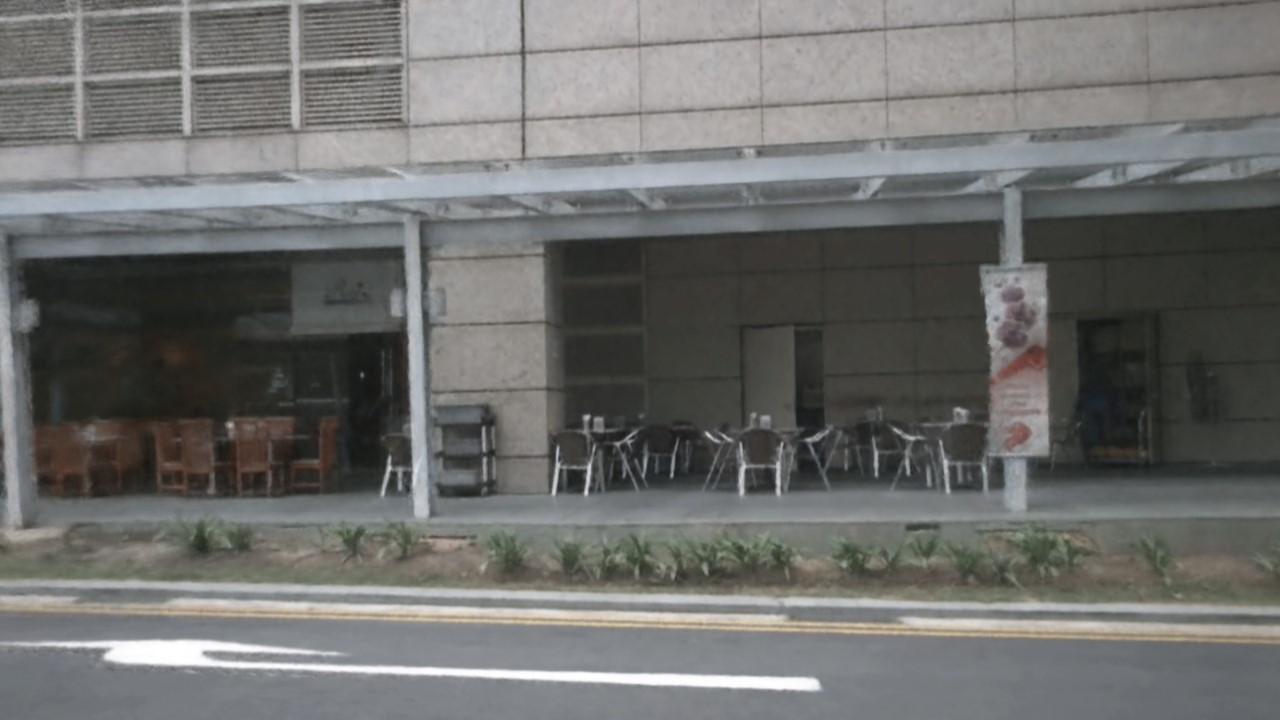} &
     \includegraphics[width=0.16\textwidth]{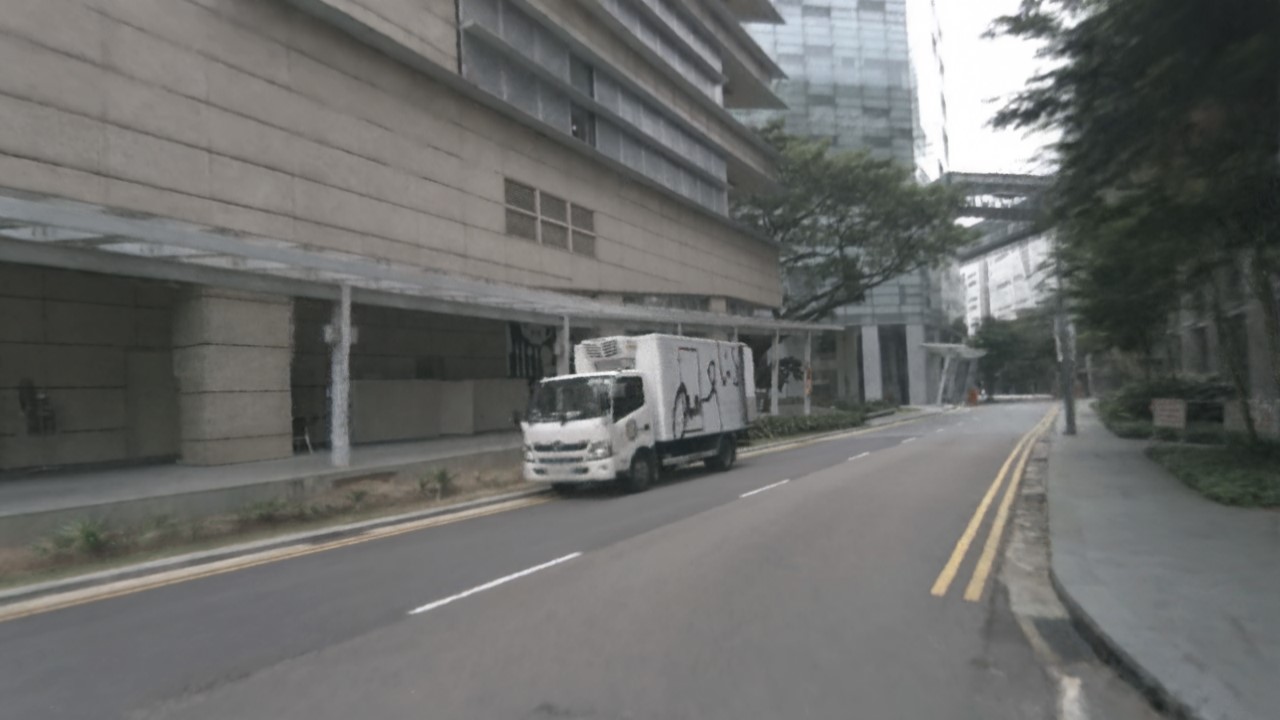}& 
     \includegraphics[width=0.16\textwidth]{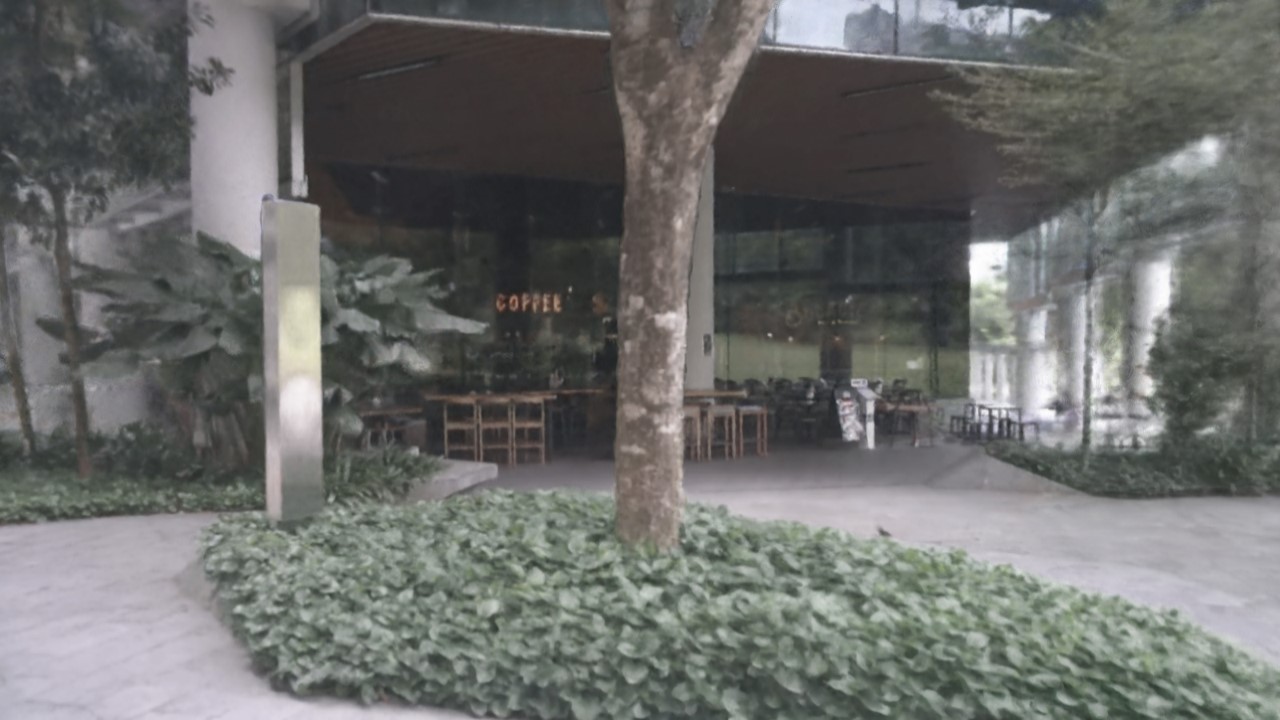} &
     \includegraphics[width=0.16\textwidth]{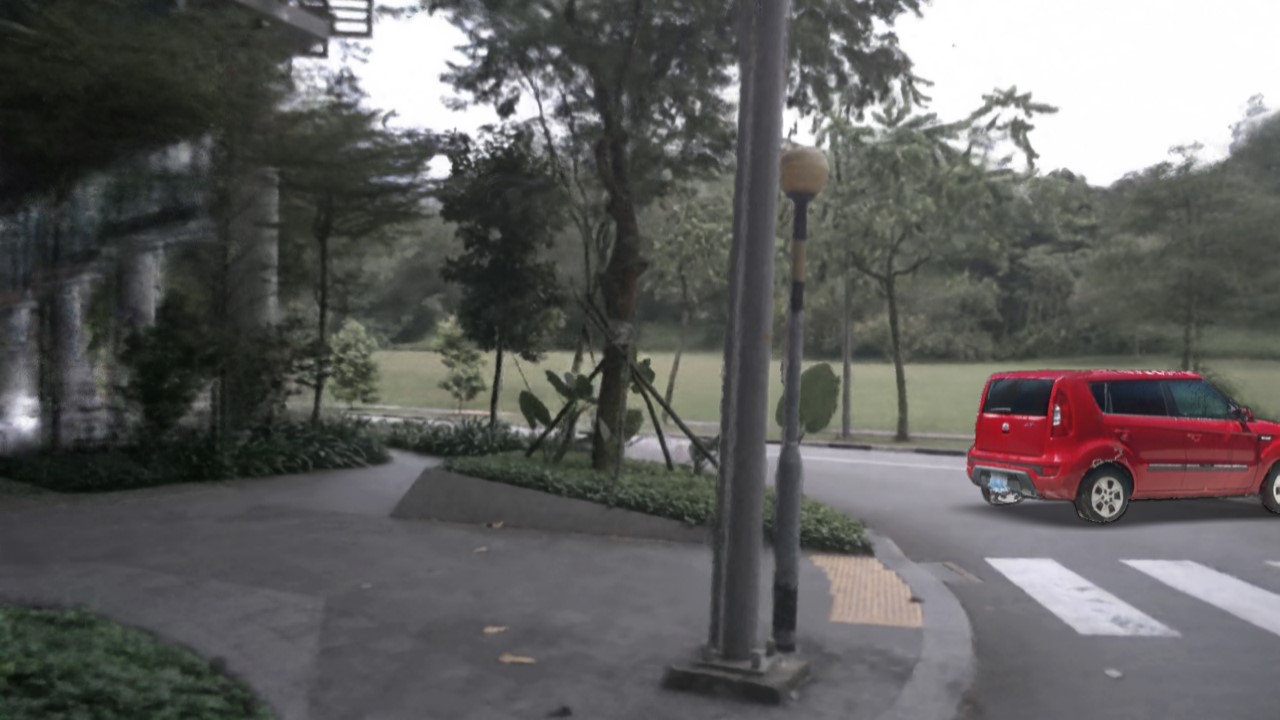}&
     \includegraphics[width=0.16\textwidth]{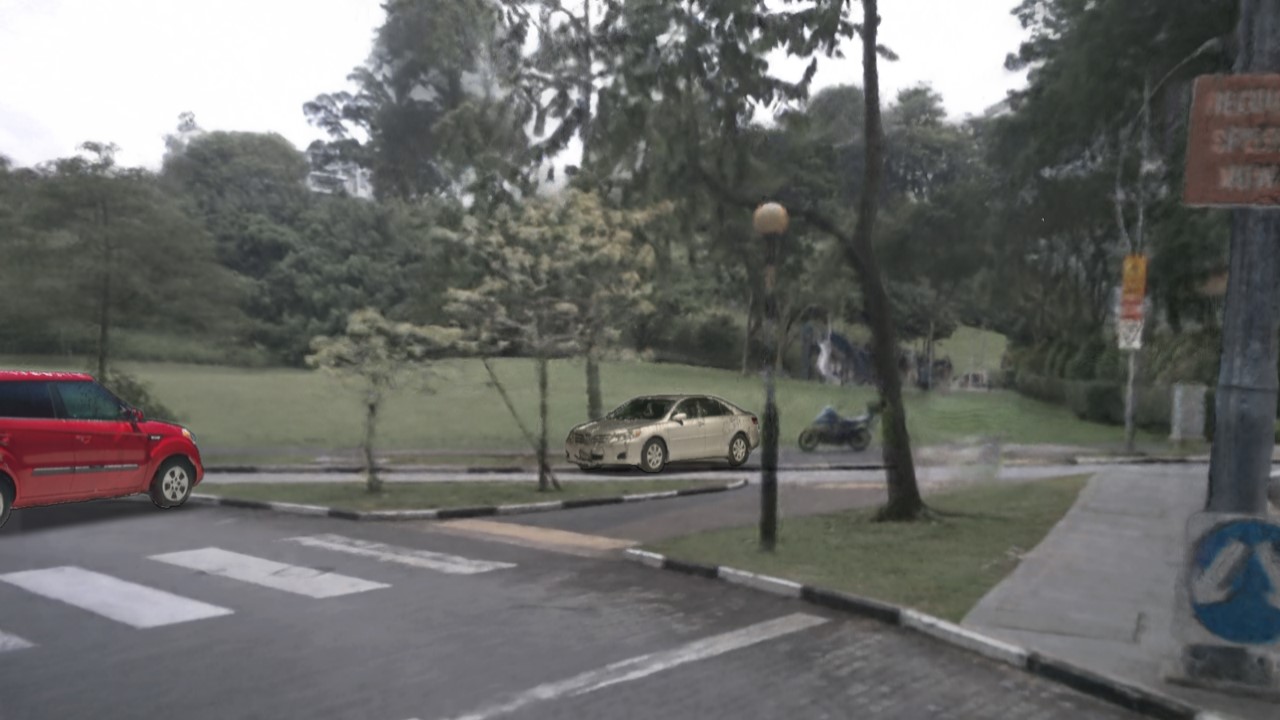}
     \\
\end{tabular}
    \caption{\textcolor{black}{Some NuScenes simulation data, each scene is inserted two cars from foreground bank.} }
    \label{fig:nus_simu_2cars}
\end{figure*} 
\begin{figure*}[th!] \centering
\setlength{\abovecaptionskip}{4pt}
\setlength{\belowcaptionskip}{-8pt}
\small
\setlength{\tabcolsep}{1pt}
\begin{tabular}{cccc}
     Car & Motorcycle & Bicycle & Human \\
     \includegraphics[width=0.24\textwidth]{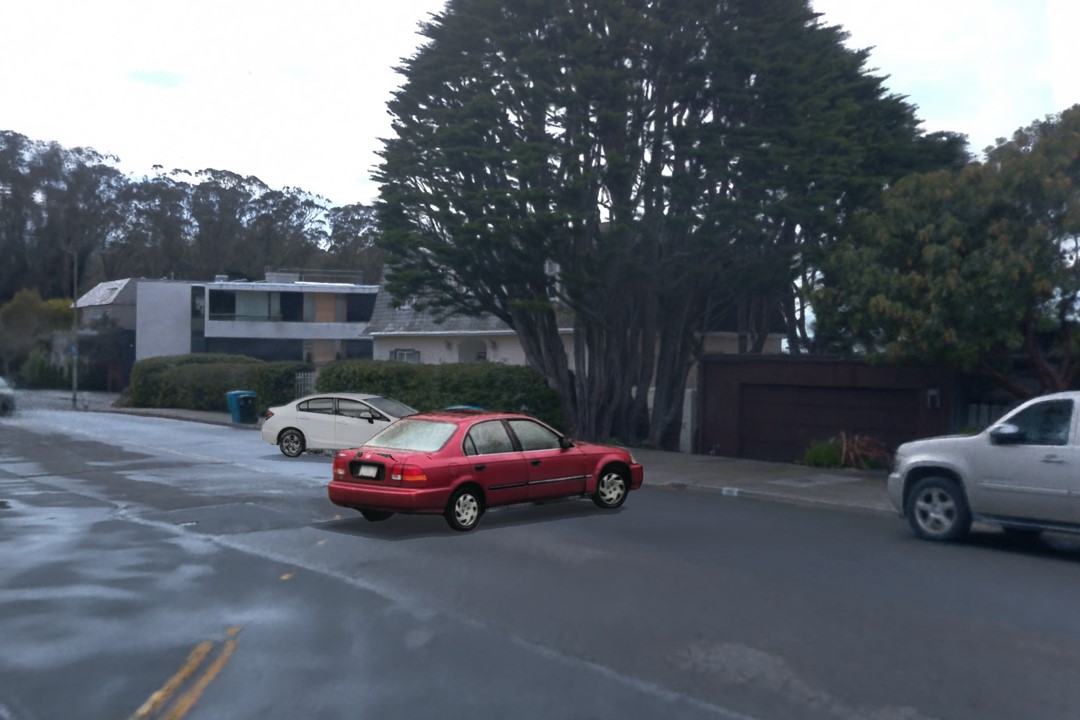} & 
     \includegraphics[width=0.24\textwidth]{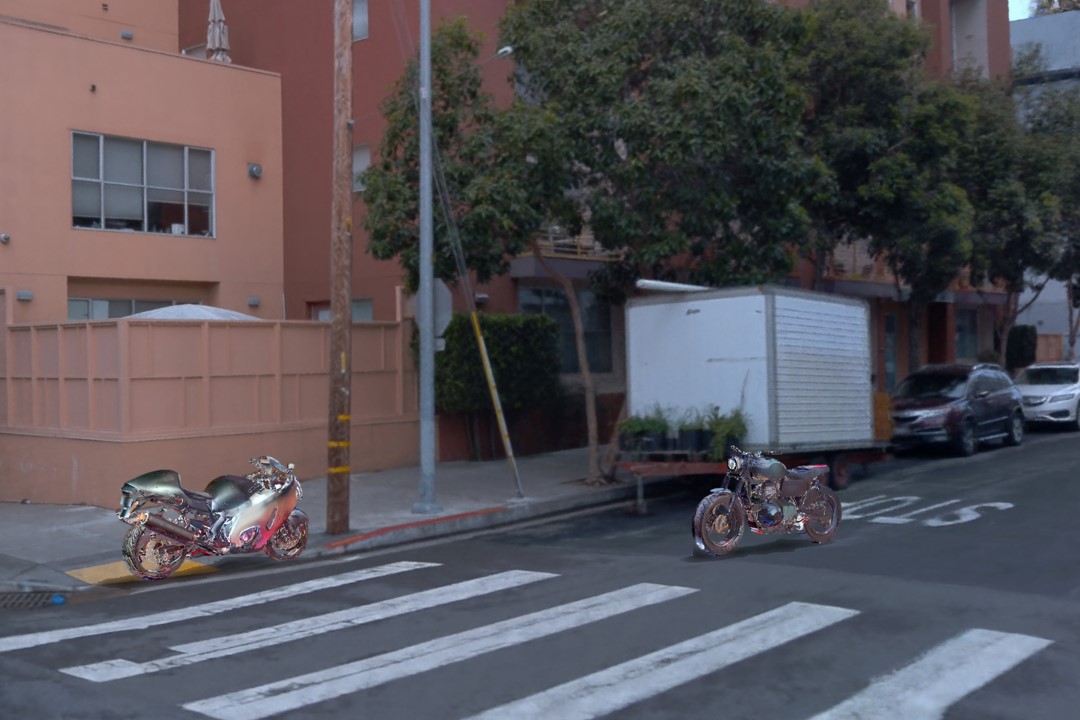} &
     \includegraphics[width=0.24\textwidth]{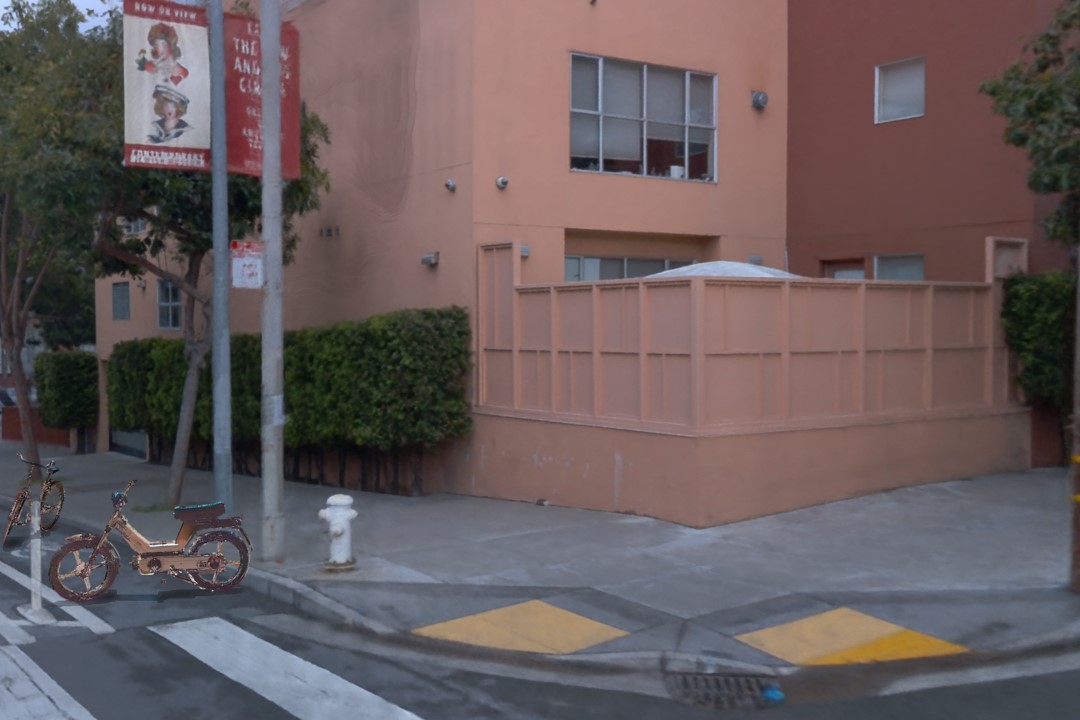}& 
     \includegraphics[width=0.24\textwidth]{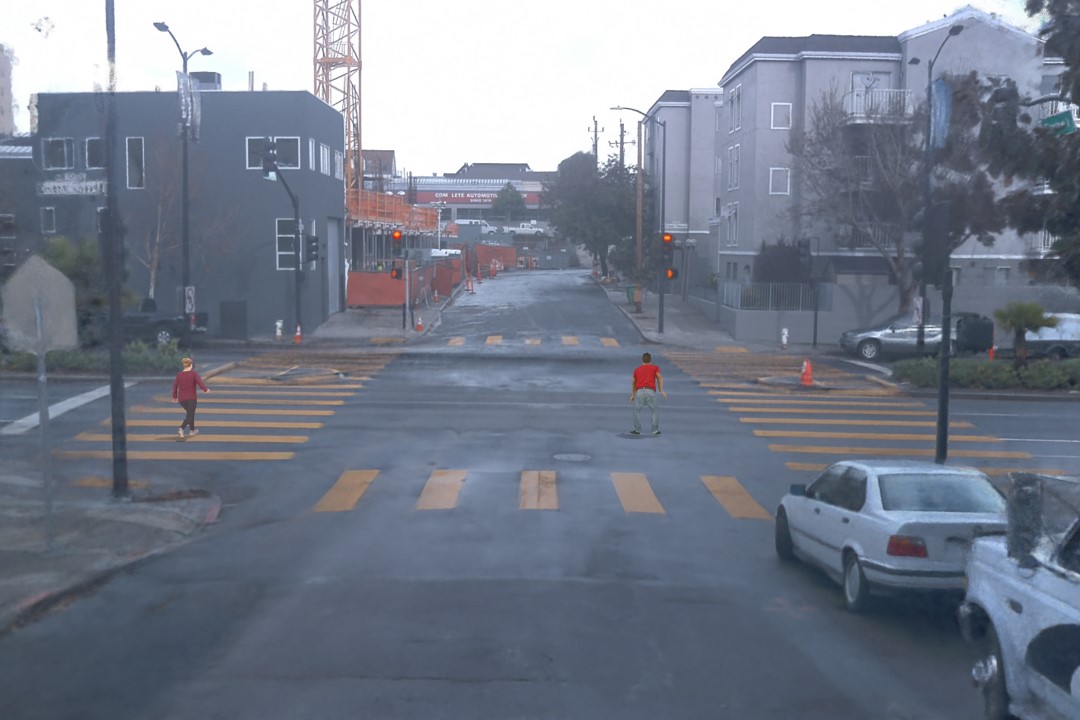}
     \\
     \includegraphics[width=0.24\textwidth]{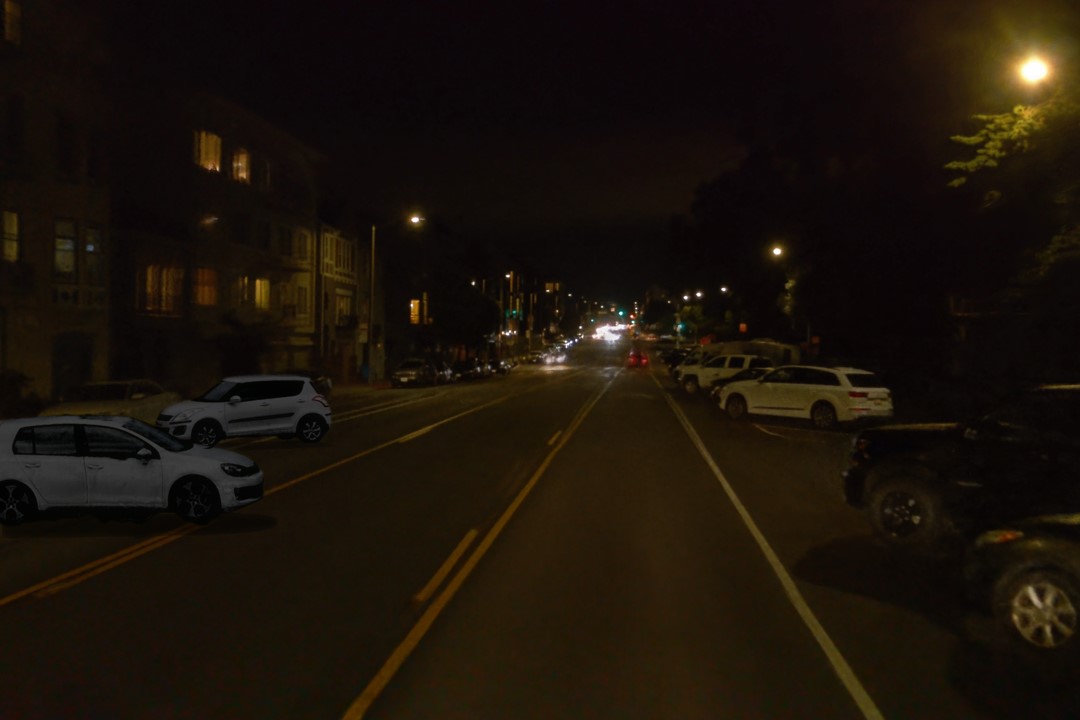} & 
     \includegraphics[width=0.24\textwidth]{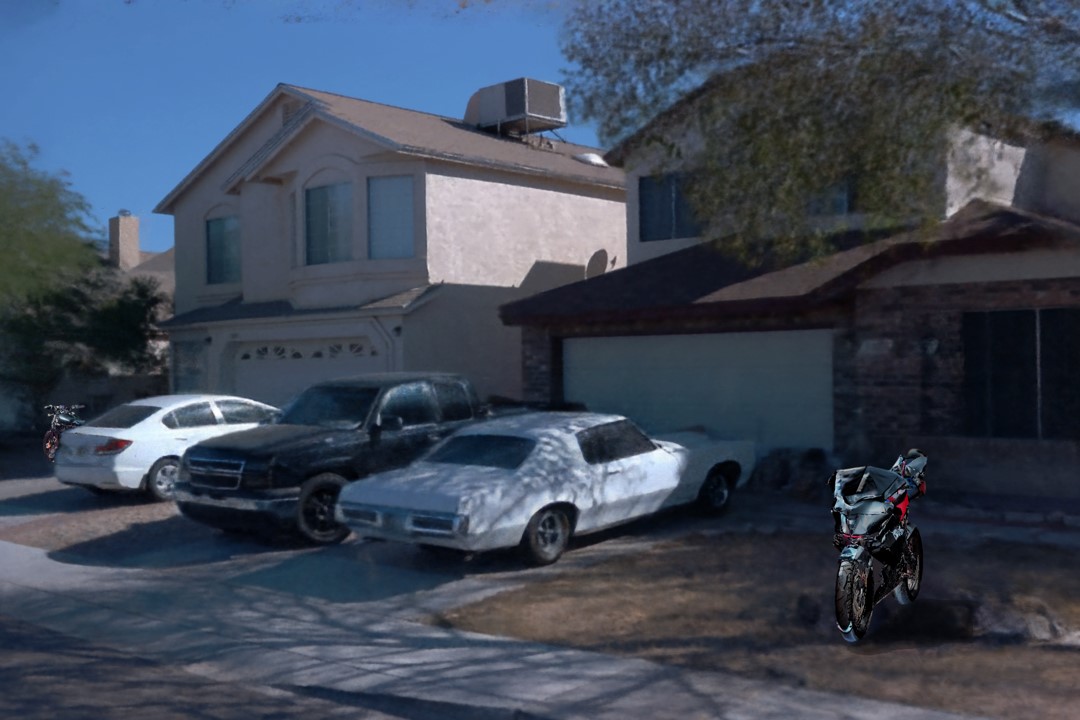} &
     \includegraphics[width=0.24\textwidth]{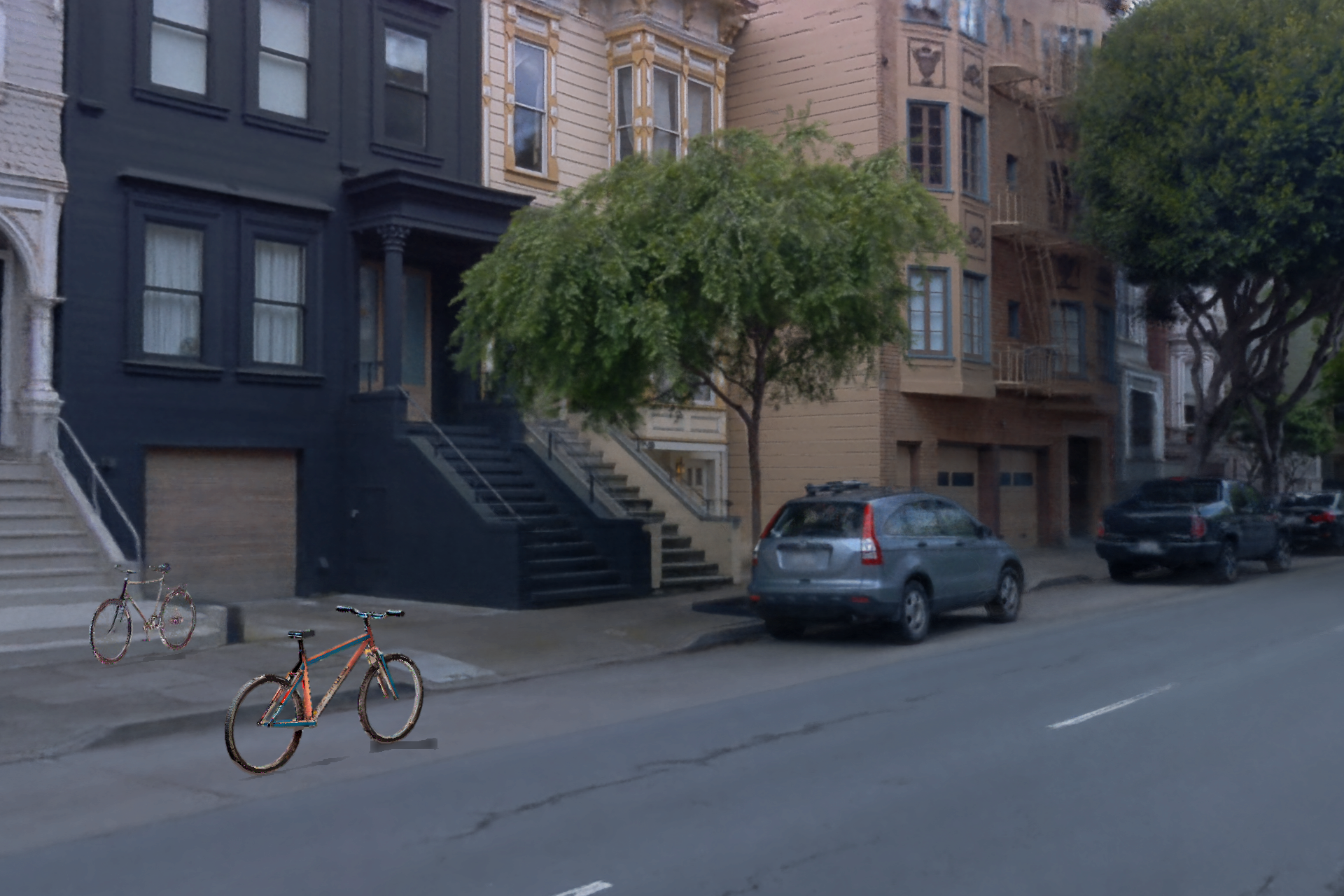}& 
     \includegraphics[width=0.24\textwidth]{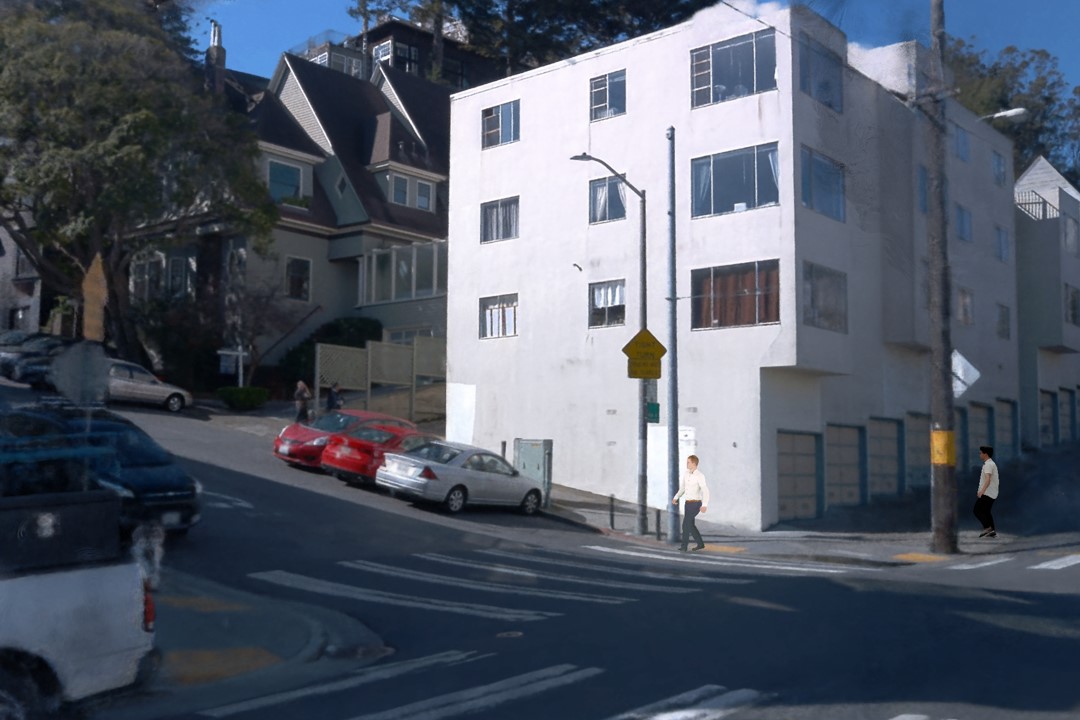}
     \\
    \includegraphics[width=0.24\textwidth]{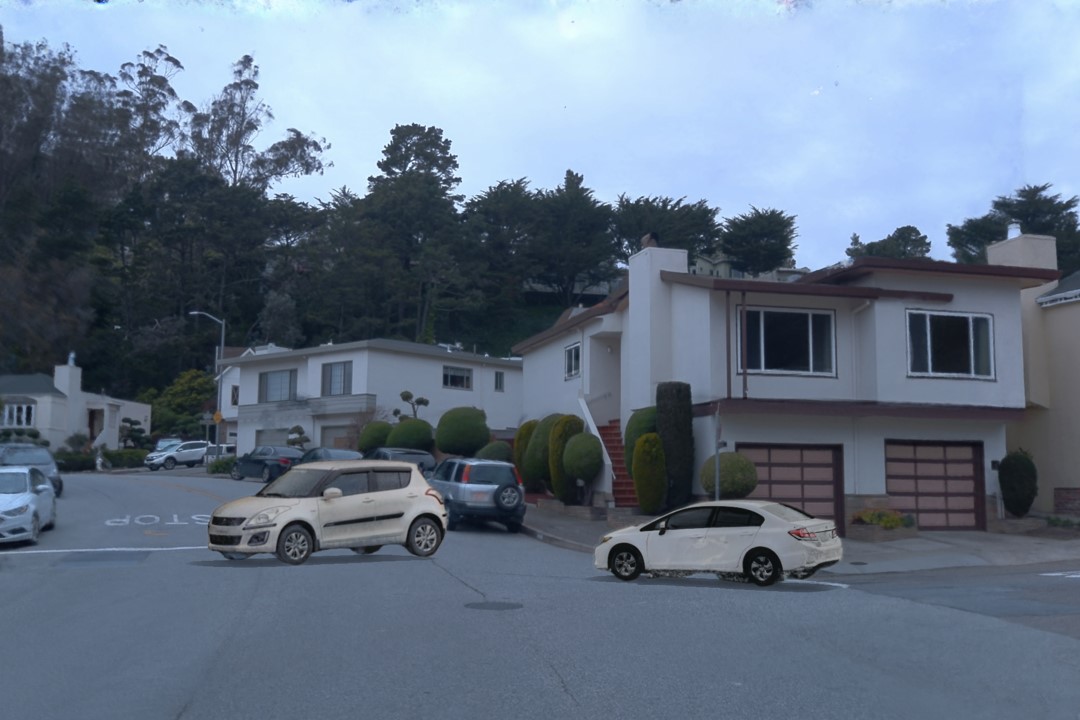} & 
     \includegraphics[width=0.24\textwidth]{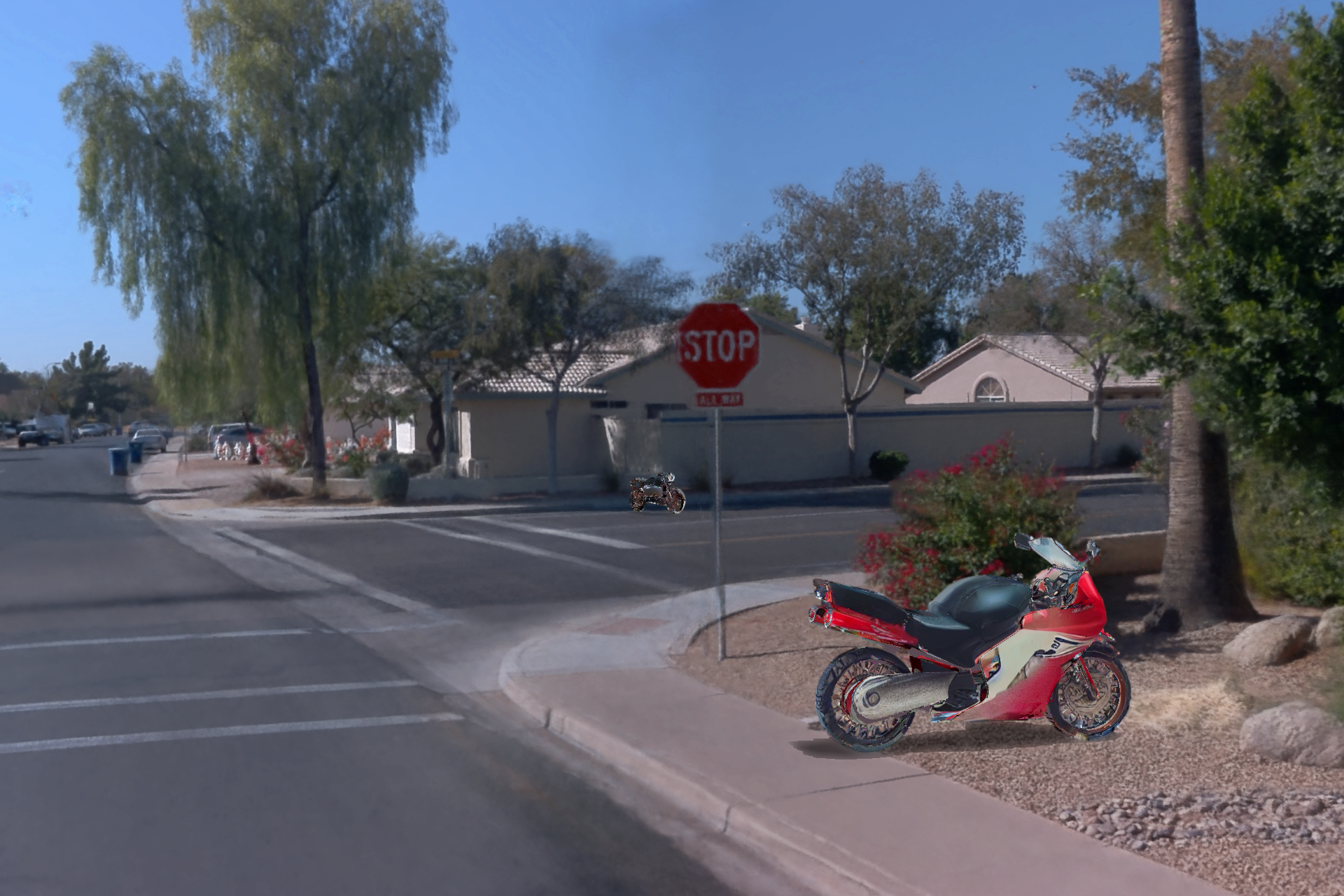} &
     \includegraphics[width=0.24\textwidth]{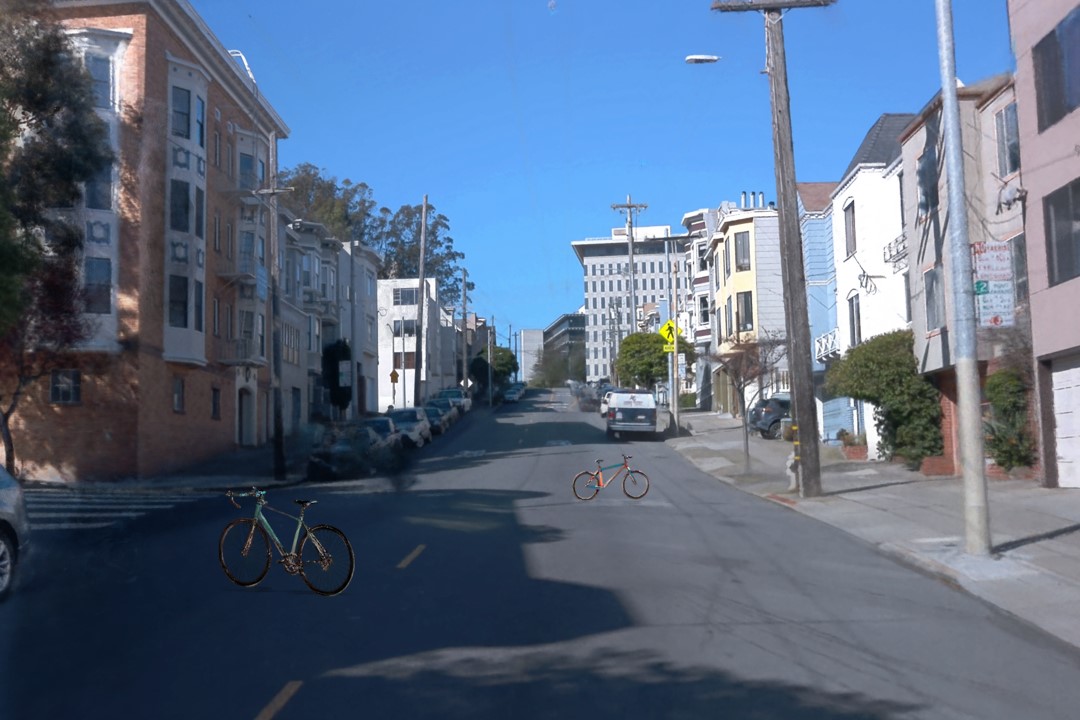}& 
     \includegraphics[width=0.24\textwidth]{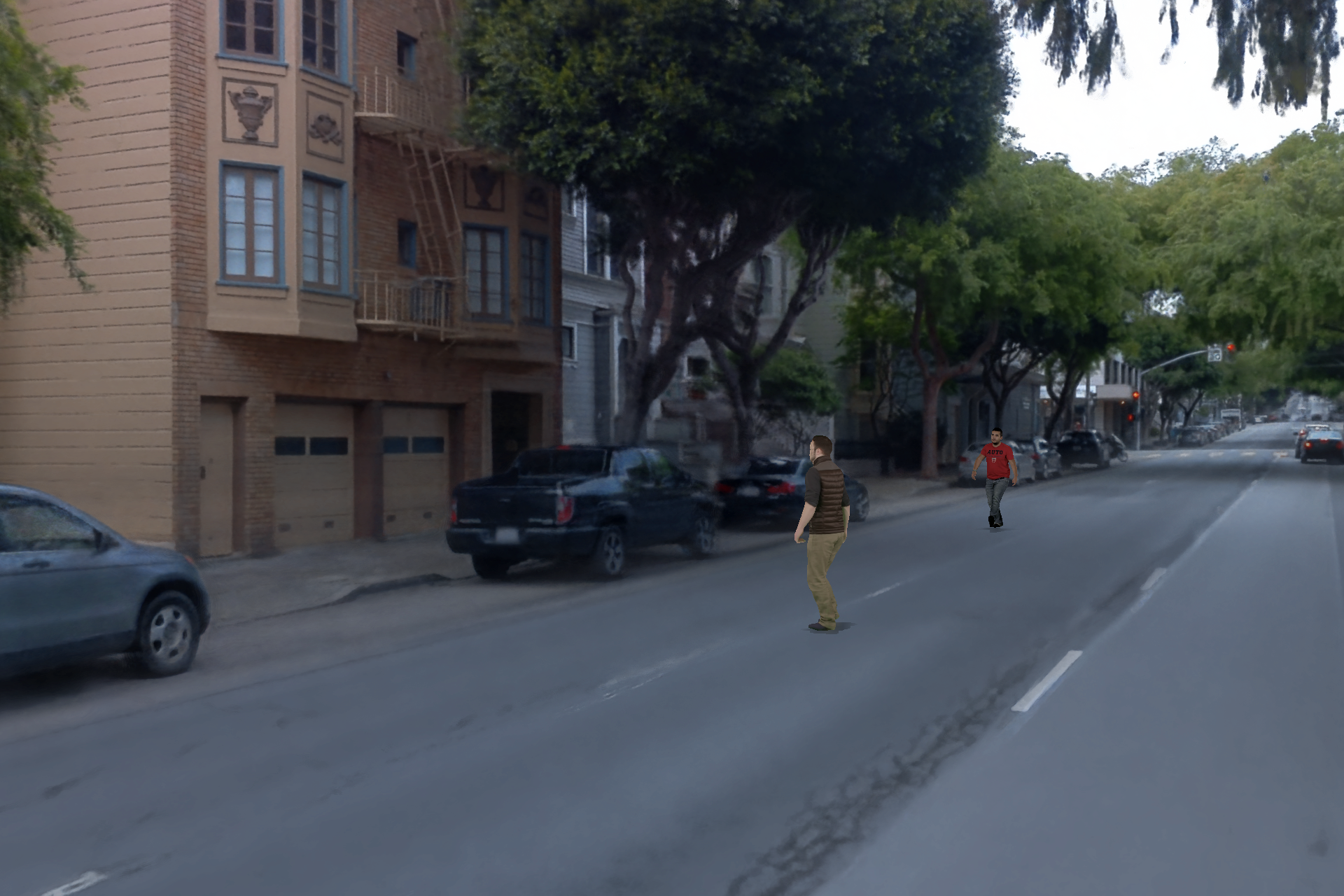} 
     \\
\end{tabular}
    \caption{Some Waymo simulation data, each image is inserted two objects from the foreground bank.} 
    \label{fig:waymo_simu_2obj}
\end{figure*}

\subsection{Simulation evaluation}
In order to verify the reliability of the data generated by our simulation system, we use the simulation data generated by the system to train the downstream task models. We use S-NeRF++ in hash grid based backbone. For most tasks, model trained by the pure simulation data already achieve some good results. Even more exciting, models trained by simulation data and fine-tuned with a small amount of real data can exceed models trained by 10 times real data used in fine-tuning. We show some nuScenes 6-camera simulation data in Figure~\ref{fig:nus_simu_2cars} and front-and-side-facing simulation data on Waymo in Figure~\ref{fig:waymo_simu_2obj}.

To generate the best background renderings, we sample rendering poses near the training cameras' poses. For every sample, we first sample two adjacent frames, and randomly interpolate their camera poses to get a base rendering pose $P_0$. Then we apply a position and rotation disturbance to $P_0$. We uniform sample the position disturbance in $(-\Delta \textbf{s},\Delta \textbf{s})$, the elevation angle disturbance in $(-\Delta \theta,\Delta \theta)$, and the horizontal angle disturbance in $(-\Delta \phi,\Delta \phi)$.

\paragraph{RGB semantic segmentation}
Due to the absence of semantic labels in the Waymo dataset, we use pseudo semantic labels to train our semantic branch. We use a pre-trained segmentation model Segformer~\cite{segformer} to predict the semantic label of each image frame. We sample 400 ground-truth semantic labels in the 100 sequences not seen by the simulation system to serve as our test set while using the remaining 13k ground-truth labels as our baseline training set. Then we randomly sample a tenth of the training set to create a subset for fine-tuning. Around 20k frames of simulation data are used to train segmentation models. Each frame is randomly fused by 2 foreground objects including pedestrians, cars, bicycles, and motorcycles. We set $\Delta \textbf{s} = [3,3,1], \Delta \theta = 5^{\circ}, \Delta \phi = 20^{\circ}$. 
In Table \ref{tab:image segmentation}, we evaluate two segmentation models based on the mIoU metrics across 19 classes. Results show that models trained with our simulation-augmented data outperform those using an amount of real data that is ten times larger. This not only demonstrates the high quality of our simulated data but also highlights its potential as a valuable tool for improving the efficiency and accuracy of model training in various applications, particularly in scenarios where collecting large volumes of real-world data is challenging or impractical.

\paragraph{Monocular 3D Vehicle Detection}
For the monocular 3D detection task, we conduct our experiments on the Waymo dataset. We select 10k real front view image frames from the 100 video sequences, which contain the 100 segments our system used as the baseline training set. From this collection, we randomly chose 1,000 frames to create a fine-tuning set. We then select 400 frames not seen by the simulation system or included in the baseline training set as our test set. We disturb the front camera view of training views and we set $\Delta \textbf{s} = [3,3,1], \Delta \theta = 5^{\circ}, \Delta \phi = 60^{\circ}$. Each simulation frame incorporated two fused cars. For detection, we use PGD~\cite{pgd} as the detector and use 0.5 IoU as the threshold of the positive samples. The evaluation results are shown in Table \ref{tab:mono 3d detection}. 
Our method surpassed the performance of those training on 1k real data and outperformed models training with 10k real data after further fine-tuning.

\paragraph{Multi-view 3D vehicle detection}
We train the multi-view 3D vehicle Detection on the nuScenes dataset. We use the 60 nuScenes segments trained in our simulation platform to generate simulation data. We sample 5k key frames in nuScenes as a baseline real data training set and 500 of them as a fine-tuning set. We select 200 frames from the 60 full nuScenes data sequences but not seen by our simulation platform as the test set. Disturbance is set to $\Delta \textbf{s} = [1,1,0], \Delta \theta = 0^{\circ}, \Delta \phi = 20^{\circ}$. We generate 12k simulated frames. Each frame contains a front, front-left, front-right, back, back-left, and back-right view with the same camera settings as in nuScenes. We insert 2 cars in each frame. Some foregrounds will appear across different views, and our simulation ensures view consistency. The multi-view detection results are shown in Table \ref{tab:mv 3d detection}. We use the method designed for the nuScenes dataset to calculate the AP, and we refer to it as $\text{AP}_\text{nus}$.

\begin{figure}[ht!]
\centering
\hspace{-3mm}
\subfigure[w/o Refinement]{
\begin{minipage}[t]{0.315\linewidth}
\centering
\includegraphics[width=1\linewidth]{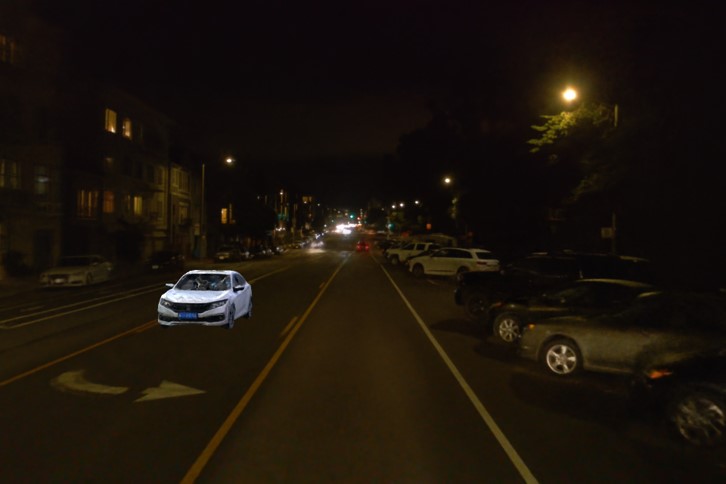}
\end{minipage}%
}%
\hspace{-2mm}
\subfigure[Naive]{
\begin{minipage}[t]{0.315\linewidth}
\centering
\includegraphics[width=1\linewidth]{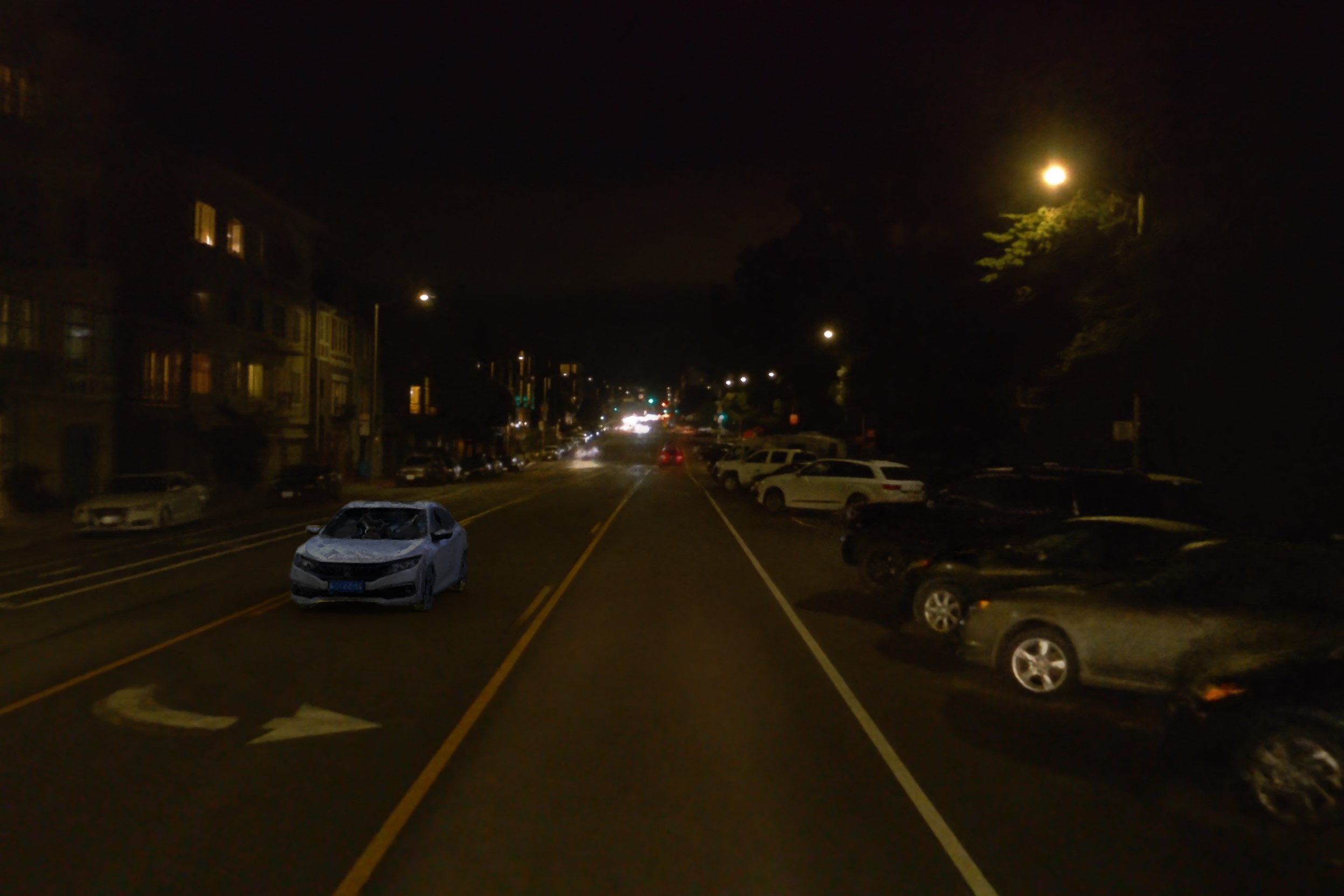}
\end{minipage}%
}%
\hspace{-2mm}
\subfigure[Optimization-based]{
\begin{minipage}[t]{0.315\linewidth}
\centering
\includegraphics[width=1\linewidth]{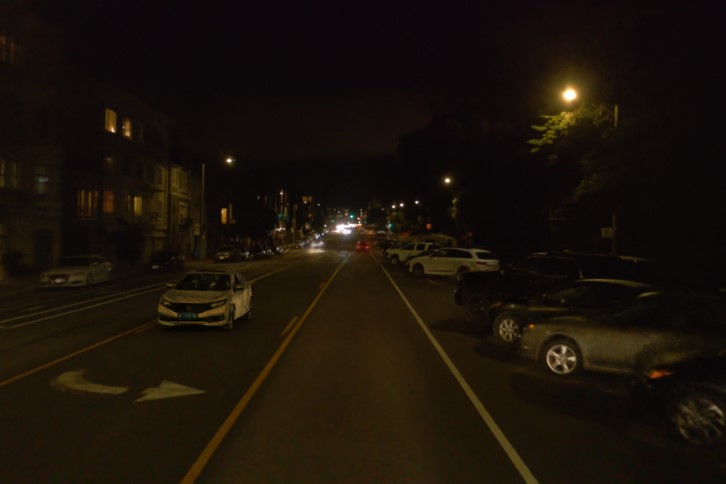}
\end{minipage}%
}%
\caption{\rebuttal{Our object insertion refinement approach improves the realism of the synthesized data; however, the  optimization-based refinement achieves even better results.}}
\label{fig:adaption ablation}
\end{figure}

\subsection{Ablation study}
\subsubsection{In terms of reconstruction components}

In this section, we present a series of ablation studies for our S-NeRF++ model. These studies were conducted using two street-view scenes from the nuScenes dataset. Ablations include using different confidence components, and pose refinement. 

\paragraph{Confidence components}

We also evaluate the effectiveness of S-NeRF++ by incorporating various confidence components (geometry and perception confidence) for learning our final confidence map.
As shown in Table \ref{tab:ablation_conf_background}, when we remove the perception confidence module, PSNR slightly drops by 0.4\%. And when we train S-NeRF++ without geometry confidence, the PSNR and SSIM are about 0.7\% lower.  We also test the effects of the threshold $\tau$ used in the geometry confidence. We find that the geometry confidence is not sensitive to the threshold $\tau$. $[10\%, 40\%]$ is a reasonable range for the threshold $\tau$.

We also report the depth error rate in Table \ref{tab:err_rate}. For the accurate depths, it predicts high confidence to encourage the NeRF geometry to be consistent with the LiDAR signal.

\paragraph{Pose refinement}
\textcolor{black}{We further explore the impact of camera pose refinement and vehicle track refinement on reconstruction quality. As shown in Figure~\ref{fig:pose refine}, we can observe that imprecise poses in the dataset lead to a decline in the quality of distant reconstructions. However, camera pose refinement can partially address this issue. Additionally, optimizing object trajectories can significantly enhance the reconstruction quality of moving vehicles. Table~\ref{tab:poserefine} also shows that refinement will lead to higher reconstruction PSNR.}

\subsubsection{In terms of simulation components}
In our final analysis, we examined the influence of different elements within our simulation system on the quality of the resulting simulation data. This experiment aimed to identify how each element contributes to the overall fidelity and effectiveness of the simulated environment.

\paragraph{Object insertion refinement module}
\rebuttal{Our focus here was to evaluate the enhancement in realism of simulation data achieved by the refinement module, which incorporates inpainting, relighting, and shading. We generated 7k simulation} \rebuttal{samples from Waymo scenes with inserted foreground cars and calculated the Fréchet Inception Distance (FID~\cite{FID}) before and after the refinement on the Waymo dataset. The results, shown in Table \ref{tab:fid refinement} and Figure~\ref{fig:adaption ablation}, demonstrate that images processed by our fusion refinement module achieve greater realism and are closer to the original dataset.
However, while the naive refinement method is less effective compared to the optimization-based refinement approach in terms of realism, it offers substantially higher efficiency, as it eliminates the need for additional optimization time.}

\paragraph{Simulation strategy}
\textcolor{black}{we explored the impact of viewpoint selection and object insertion on downstream task training in Table~\ref{tab:sampling strategy}. In this context, interpolation refers to rendering at linearly interpolated camera poses between adjacent training frames, while extrapolation involves applying perturbations to the training camera poses, such as rotations and translations.
We found that while the original views enhance the perception model, view interpolation provides a slight improvement, and view extrapolation leads to an additional boost. 
Typically, the major performance enhancement in both pretraining and final finetuning comes from the insertion of new vehicles.}

\paragraph{Generation vs. reconstruction}
\textcolor{black}{
We further explore the impact of a foreground asset bank, constructed through reconstruction and generation, on simulation data. We measure the impact using two downstream tasks: semantic segmentation and multi-view 3D vehicle detection. In the semantic segmentation task, we inserted vehicles, people, bicycles, and motorcycles from different source foreground banks to the original scenes to generate the simulation data, and measured the mean IOU for these categories. As shown in Table~\ref{tab: foreground insertion}, generated foreground assets resulted in greater improvements for the semantic segmentation task. This might be because motorcycles and bicycles, which have intricate structures, are difficult to reconstruct and have fewer resources to reconstruct. Generation methods, using existing mesh priors, can effectively enrich these categories. In the detection task, we inserted vehicles, and used car mAP as the metric. Reconstruction methods improve detection performance more than generation methods, due to the higher quality of reconstructed vehicles.}

\paragraph{Pretraining vs. training together}
\textcolor{black}{We investigate the results of using simulation data for pre-training and combining simulation data with all real data for joint training. We find that joint training results in performance degradation in Table~\ref{tab:training strategy}. As the simulation data is much more than the real data, e.g. 10k vs. 500, the training performance will be dominated by the simulated data while there is still a gap between the generated simulation data and the real data.}

\section{Conclusion}

In conclusion, the S-NeRF++ autonomous driving simulation system represents a significant advancement in the field of self-driving data enhancement and traffic scenario simulation. Our system harnesses the power of neural reconstruction to faithfully reconstruct 3D street scenes from autonomous driving datasets, while also rendering visually compelling novel views. \textcolor{black}{S-NeRF++ demonstrates superior performance compared to current state-of-the-art methods in the reconstruction of both static and dynamic scenes.} The integration of sophisticated generation techniques, rooted in the latent diffusion model, has substantially broadened the variety and quality of both the foreground asset bank and the simulation data. Additionally, the implementation of a meticulous foreground-background fusion pipeline addresses the challenges of illumination and shadow processing effectively. The efficacy of S-NeRF++ is further validated through its application in training diverse downstream tasks on nuScenes and Waymo datasets, where it consistently produces high-quality and reliable simulation data. This underscores its potential as a valuable tool in the ongoing development of autonomous driving technologies.

\section*{Acknowledgments}
This work was supported by the National Natural Science Foundation of China (Grant No. 62376060).

\bibliographystyle{IEEEtran}
\bibliography{bibliography.bib}

\begin{IEEEbiography}[{\includegraphics[width=1in,height=1.25in,clip,keepaspectratio]{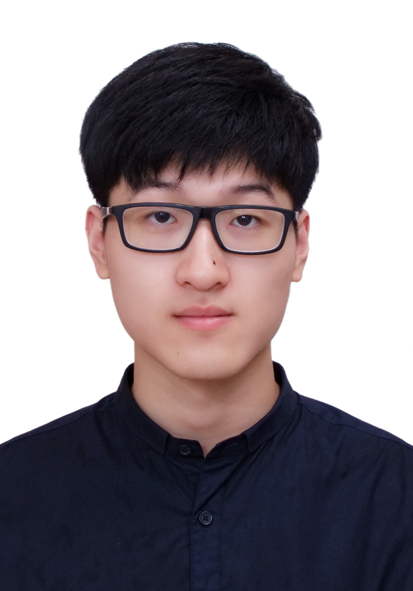}}]{Yurui Chen}
is currently an M.S. student in School of Data Science at Fudan University, supervised by Prof. Li Zhang. Previously, he obtained his Bachelor's degree from the School of Data Science at Fudan University. His research interests include 3D reconstruction, generation and understanding.
\end{IEEEbiography}
\begin{IEEEbiography}
[{\includegraphics[width=1in,height=1.25in,clip,keepaspectratio]{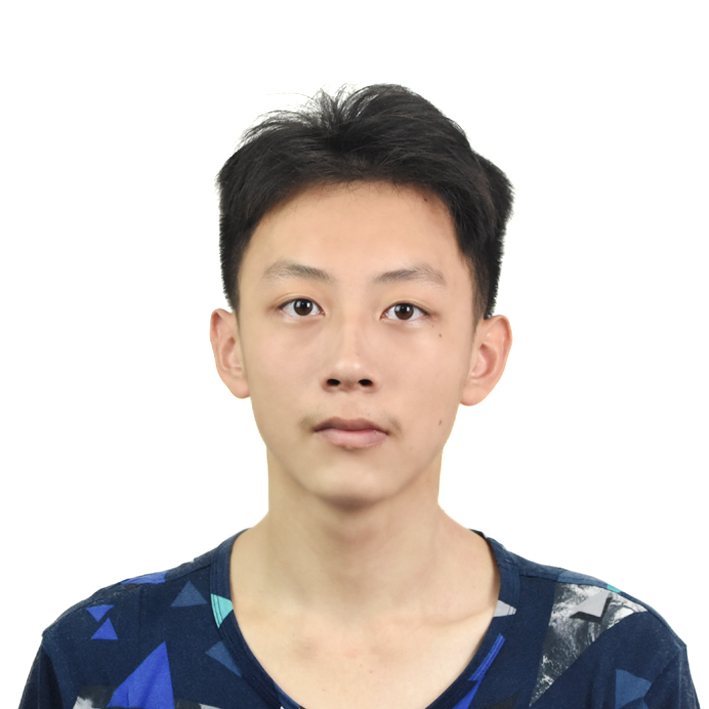}}]
{Junge Zhang}
is currently a Ph.D. student at the University of California, Riverside, supervised by Jiachen Li. Previously, he obtained his Bachelor’s degree and Master’s degree from the School of Mathematical Sciences at Fudan University in 2021 and 2024, respectively. His research interests include robotics, generative models, and 3D vision.
\end{IEEEbiography}
\begin{IEEEbiography}
[{\includegraphics[width=1in,height=1.25in,clip,keepaspectratio]{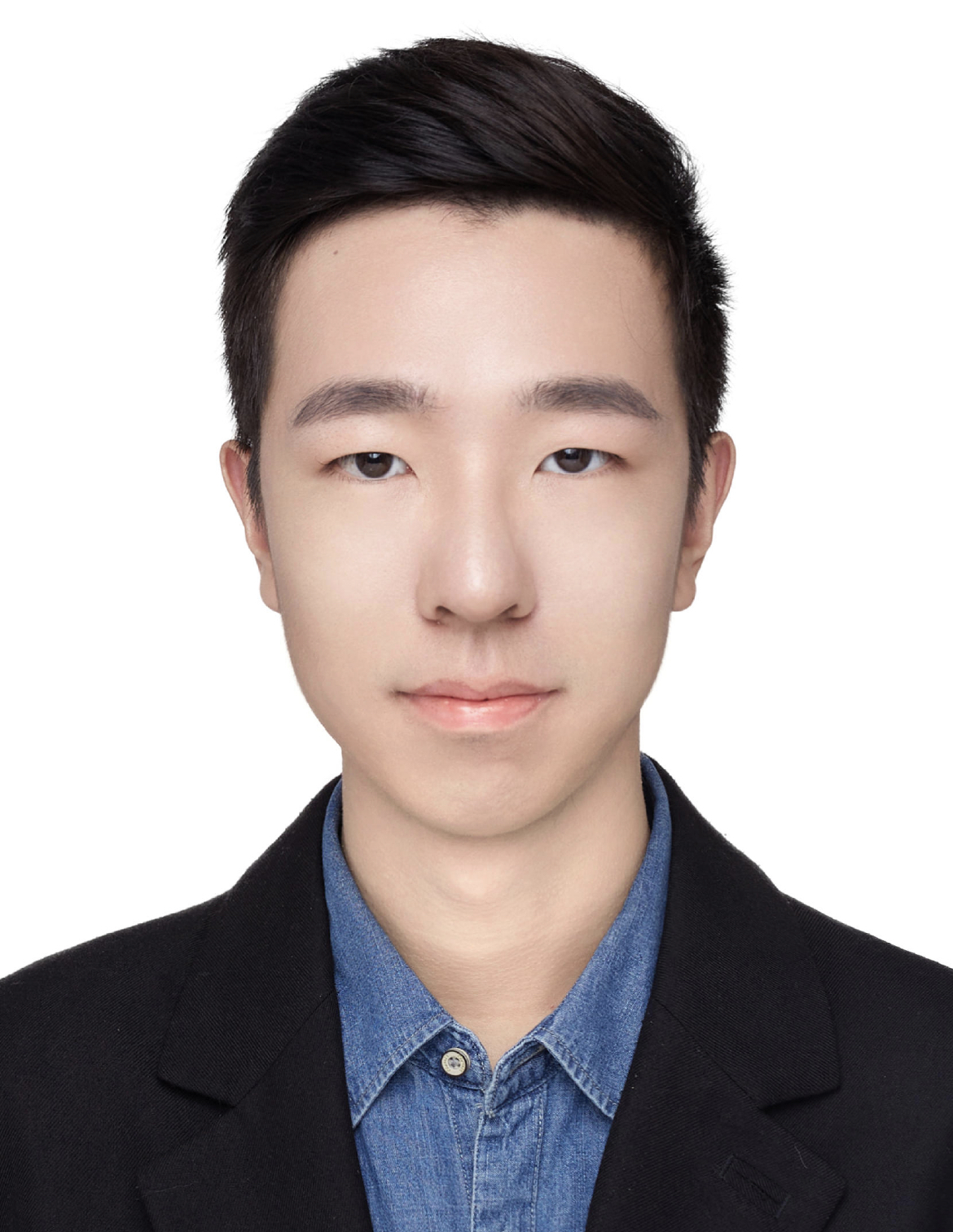}}]
{Ziyang Xie}
is currently an M.S. student in Computer Science at the University of Illinois at Urbana-Champaign (UIUC), supervised by Prof. Shenlong Wang and Prof David Forsyth. Previously, he obtained his Bachelor’s degree from the School of Computer Science at Fudan University and was actively involved in the Zhang Vision Group led by Prof. Li Zhang. His research interests include 3D reconstruction, generation, and simulation.
\end{IEEEbiography}
\begin{IEEEbiography}
[{\includegraphics[width=1in,height=1.25in,clip,keepaspectratio]{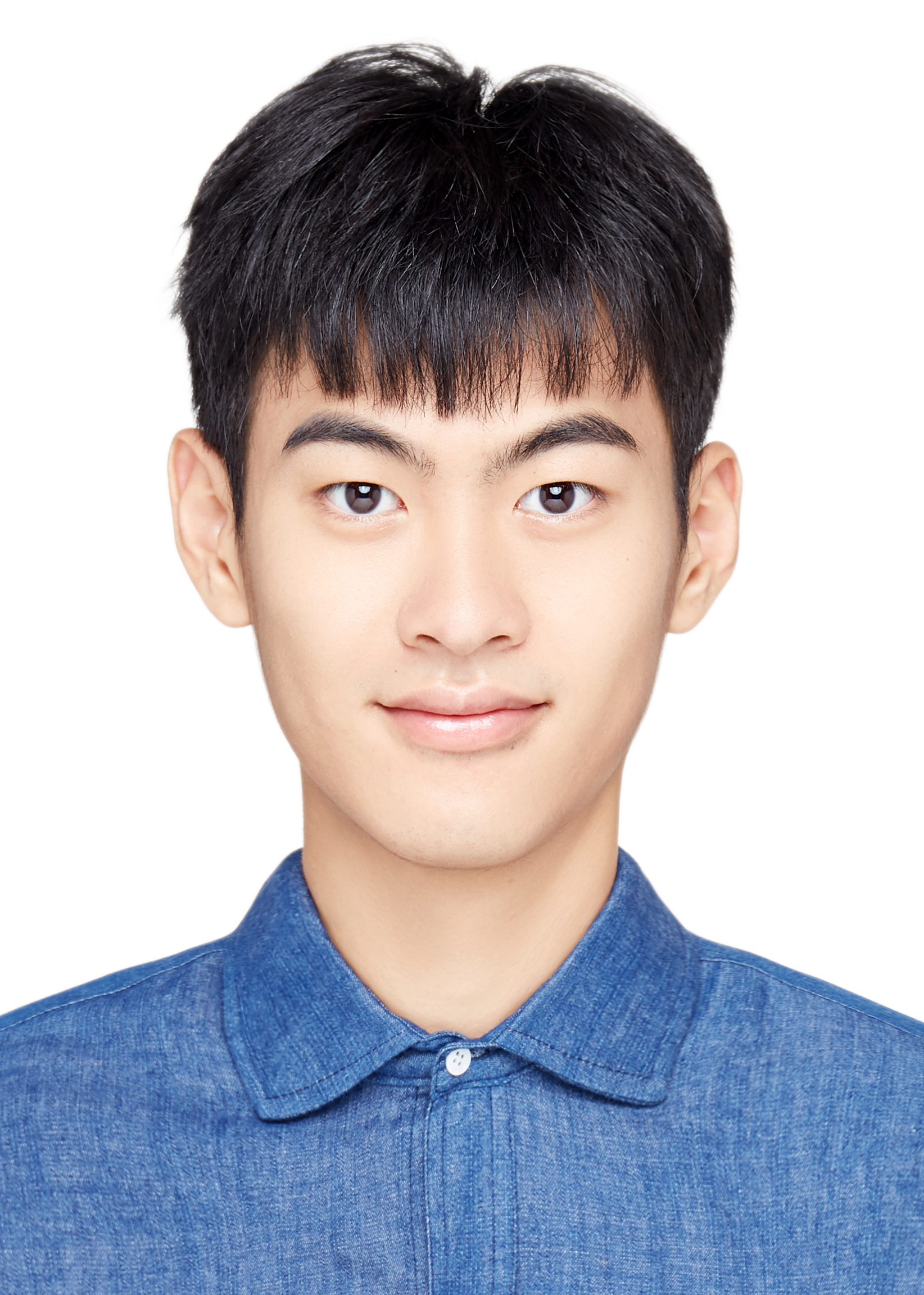}}]
{Wenye Li}
is currently an M.S. student in School of Data Science at Fudan University, supervised by Prof. Ke Wei. Previously, he obtained his Bachelor's degree from the School of Data Science at Fudan University and was involved in the Zhang Vision Group led by Prof. Li Zhang. His research interests include 3D reconstruction and reinforcement learning.
\end{IEEEbiography}
\begin{IEEEbiography}
[{\includegraphics[width=1in,height=1.25in,clip,keepaspectratio]{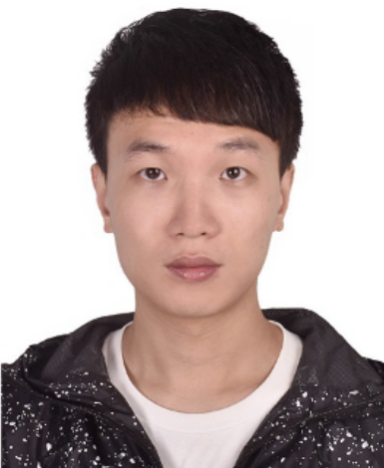}}]
{Feihu Zhang}
received PhD from Department of Engineering Science, University of Oxford, and M.Phil. degree with from The Chinese University of Hong Kong. His research interests are computer vision and 3D generation.
\end{IEEEbiography}
\begin{IEEEbiography}
[{\includegraphics[width=1in,height=1.25in,clip,keepaspectratio]{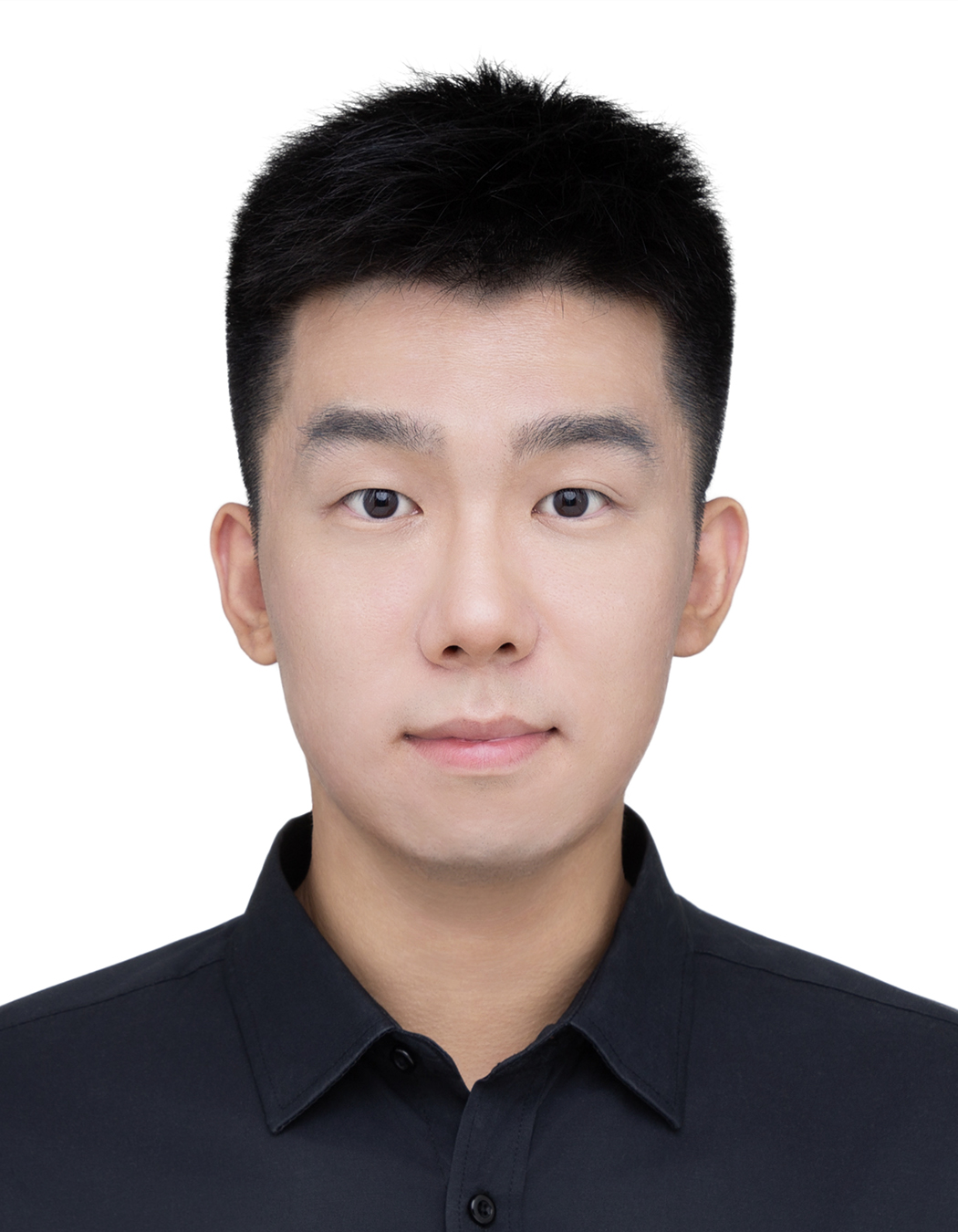}}]
{Jiachen Lu}
received BEng degree in electronic and computer engineering from University of Michigan-Shanghai Jiaotong University (UM-SJTU) Joint Institute, Shanghai, China. He is now a Master student in School of Data Science, Fudan University under the supervision of Prof. Li Zhang. 
His research interests include computer vision, deep learning and autonomous driving.
\end{IEEEbiography}
\begin{IEEEbiography}
[{\includegraphics[width=1in,height=1.25in,clip,keepaspectratio]{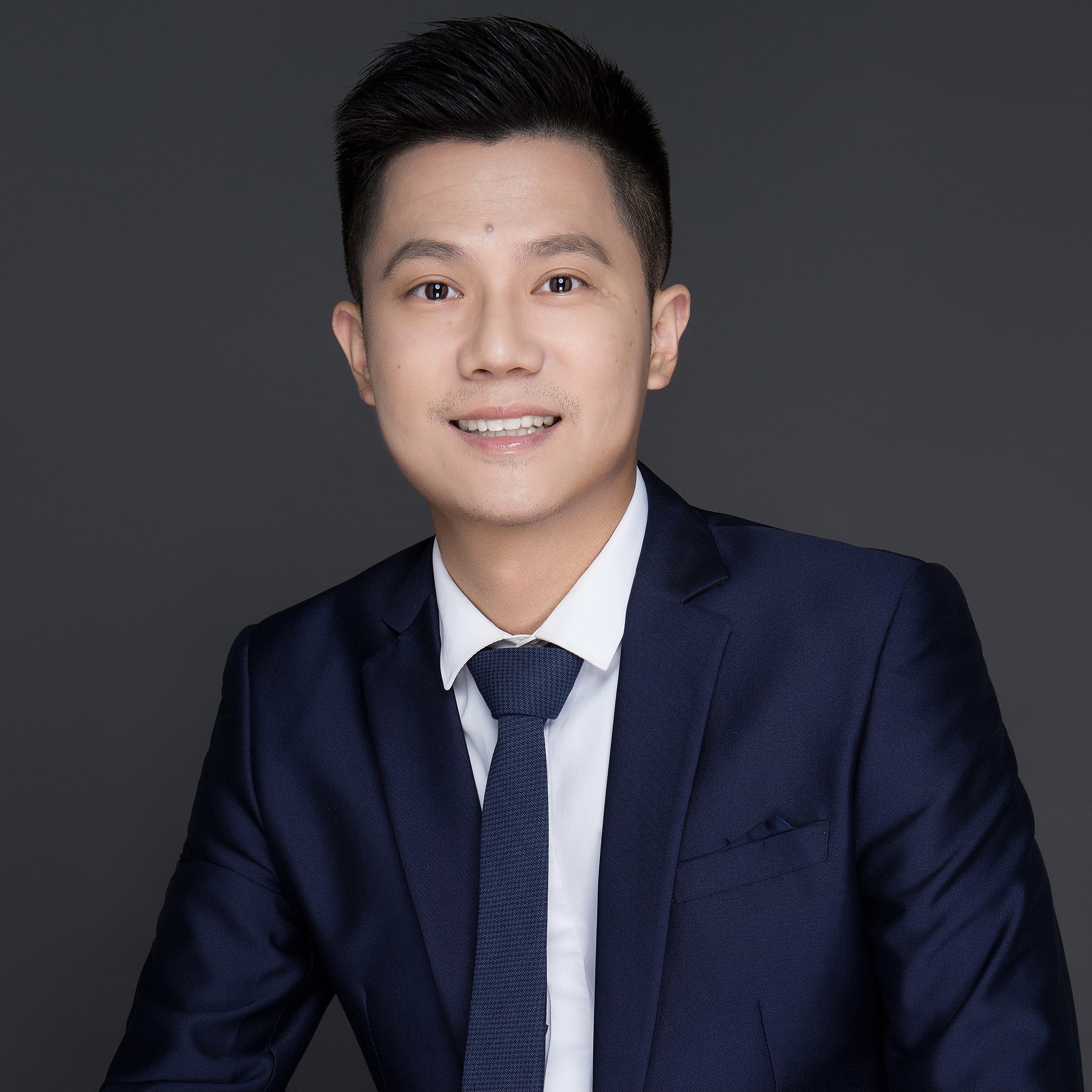}}]
{Li Zhang}
is a Professor at Fudan University. 
Previously, he was a Postdoctoral Research Fellow at the University of Oxford. 
Prior to joining Oxford, he read his PhD in computer science at Queen Mary University of London. 
His research interests include computer vision and deep learning.
\end{IEEEbiography}

\end{document}